%% file: paper.tex
\documentclass[10pt,twocolumn,letterpaper]{article}

\usepackage{cvpr}
\usepackage{times}
\usepackage{epsfig}
\usepackage{graphicx}
\usepackage{amsmath}
\usepackage{amssymb}
\usepackage{multirow}
\usepackage{booktabs}

\usepackage{longtable}
\usepackage{array}
\usepackage{makecell}

\usepackage{cellspace}
\newcommand\cincludegraphics[2][]{\raisebox{-0.2\height}{\includegraphics[#1]{#2}}}

\usepackage[pagebackref=true,breaklinks=true,letterpaper=true,colorlinks,bookmarks=false]{hyperref}

\cvprfinalcopy 

\def\cvprPaperID{4011} 

\newcommand{\PAR}[1]{\vskip4pt \noindent {\bf #1~}}
\newcommand{\PARbegin}[1]{\noindent {\bf #1~}}

\DeclareMathOperator{\height}{h}

\newcommand{\setupname}[1]{\textbf{#1}}
\newcommand{\categoryname}[1]{\textit{#1}}

\ifcvprfinal\fi
\begin{document}

\title{Large-Scale Object Discovery and Detector Adaptation from Unlabeled Video}

\author{Aljo\u{s}a O\u{s}ep\textsuperscript{*}, Paul Voigtlaender\textsuperscript{*}, Jonathon Luiten, Stefan Breuers, and Bastian Leibe\\
Computer Vision Group\\
Visual Computing Institute\\
RWTH Aachen University, Germany \\
{\tt\small \{osep, voigtlaender, breuers, leibe\}@vision.rwth-aachen.de jonathon.luiten@rwth-aachen.de}
}

\maketitle
\thispagestyle{plain}
\pagestyle{plain}

\begin{abstract}
We explore object discovery and detector adaptation based on unlabeled video sequences captured from a mobile platform.
We propose a fully automatic approach for object mining from video which builds upon a generic object tracking approach. By applying this method to three large video datasets from autonomous driving and mobile robotics scenarios, we demonstrate its robustness and generality.
Based on the object mining results, we propose a novel approach for unsupervised object discovery by appearance-based clustering. We show that this approach successfully discovers interesting objects relevant to driving scenarios. In addition, we perform self-supervised detector adaptation in order to improve detection performance on the KITTI dataset for existing categories.
Our approach has direct relevance for enabling large-scale object learning for autonomous driving.
\end{abstract}

\vspace{-3mm}
\section{Introduction}

\begin{figure}[t]
\begin{center}
   \includegraphics[width=1.0\linewidth]{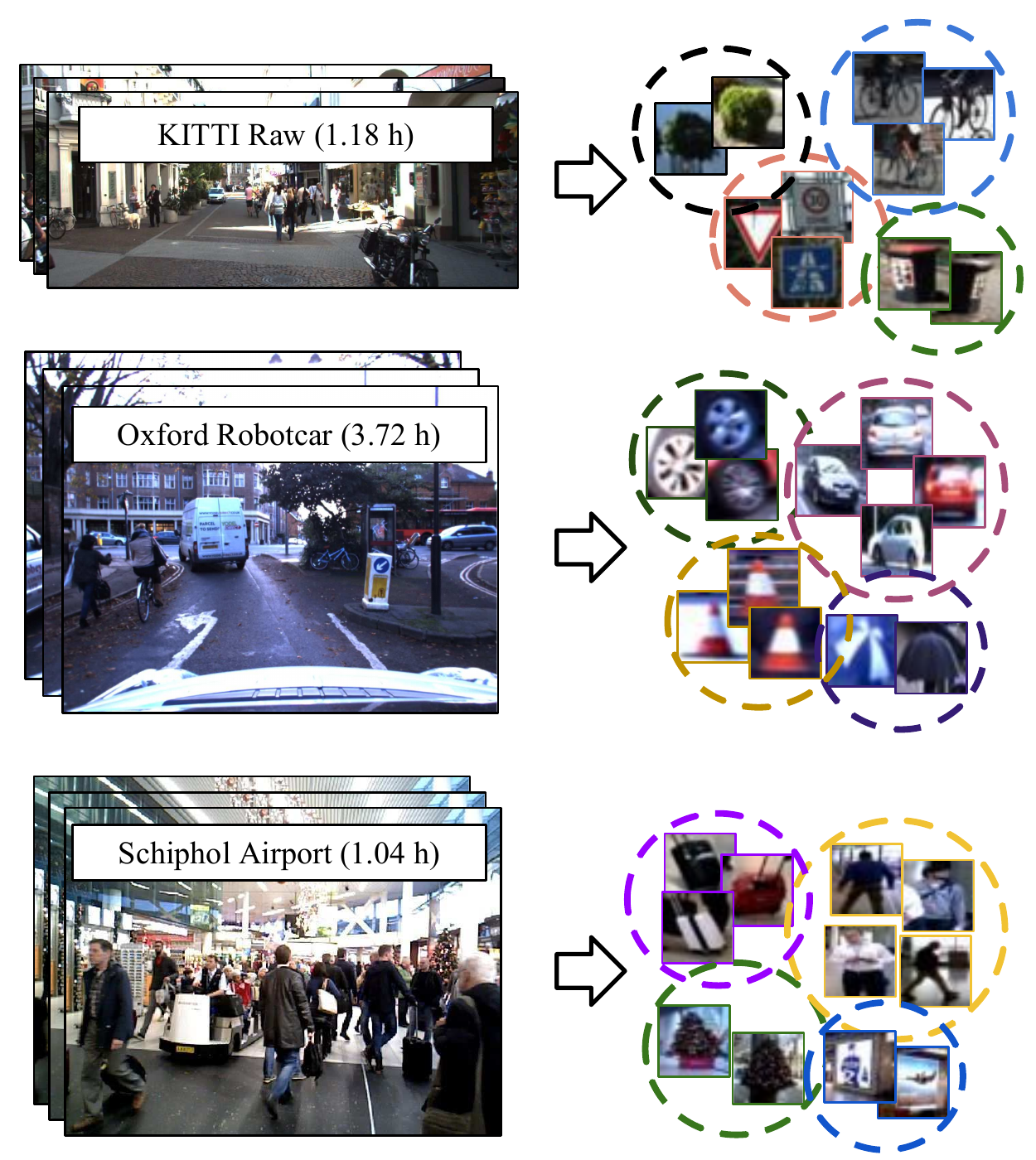}
\end{center}
   \caption{Examples of the object categories automatically discovered by our proposed self-supervised object mining and discovery approach on 3 large datasets.}
\label{fig:long}
\label{fig:onecol}
\end{figure}

\footnotetext{\textsuperscript{*} Equal contribution.}
Deep learning has revolutionized the way research is being performed in computer vision, and the success of this development holds great promise for important applications such as autonomous driving \cite{Janai17ARXIV}.

However, deep learning requires huge quantities of annotated training data, which are very costly to obtain. Consequently, progress has so far been limited to areas where such data is available, and community efforts such as PASCAL VOC \cite{Everingham10IJCV}, ImageNet \cite{Deng09CVPR}, CalTech \cite{Dollar12PAMI}, KITTI \cite{Geiger12CVPR}, Microsoft COCO \cite{Lin14ECCV}, or Cityscapes \cite{Cordts16CVPR} have been instrumental in enabling recent successes. It is largely thanks to those efforts that we nowadays have good object detectors \cite{Ren15NIPS,Redmon16CVPR,Liu16ECCV} at our disposal for a limited number of 20-80 object categories.

When moving from image interpretation tasks to video understanding problems, however, it becomes clear that the current strategy of using exhaustive human annotation will quickly become infeasible. This problem is of particular relevance in autonomous driving, where future intelligent vehicles will have to deal with a large variety of driving scenarios involving challenging imaging conditions, as well as a multitude of relevant object classes, many of which are not captured by today's detectors.
Moreover, differences in sensor setups and imaging conditions imply that even the available detectors often cannot directly be used in autonomous driving scenarios \cite{Rajaram16ITSC}, but they first have to be optimized for the specific application environment in a \emph{domain adaptation} process -- which again involves costly annotations.

In this paper, we investigate how to
significantly reduce the annotation effort for such domain adaptation and object learning tasks. The core idea of our approach is to use large amounts of unlabeled video data (several hours of raw video) in order to mine object tracks for both previously seen and unseen object categories in a fully automatic fashion. Based on the mined object tracks, we then apply \emph{self-supervised training} in order to fine-tune detectors of known object categories and to adapt them to changing application scenarios. In addition, we propose an approach for \emph{unsupervised object discovery} and detector learning of novel, scenario-specific object classes that can complement the existing detectors. 

Our approach builds upon a generic object tracking pipeline that can bootstrap existing region proposal generators \cite{Pinheiro16ECCV}
in order to extract high-quality tracks of both known and unknown object categories from unlabeled video collections in a fully automatic fashion. 

In this paper, we present a large-scale feasibility study\footnote{We will make our code, models, and created annotations publicly available.} for object mining on 3 large datasets (KITTI Raw \cite{Geiger13IJRR}, Oxford Robotcar \cite{Maddern17IJRR}, and Schiphol Airport) for autonomous driving and mobile robotics, comprising together roughly 6 hours of video data consisting of $370,\!000$ frames.
To the best of our knowledge, this is the first time such a large-scale generic object mining effort has been undertaken. 

In summary, this paper makes the following contributions: 
i) We propose a self-supervised approach for object mining and discovery from large unlabeled video collections. We demonstrate the generality of our approach by applying it to 3 different large datasets. 
ii) We demonstrate a use case of our mining pipeline for detector fine-tuning and domain adaptation on the KITTI dataset.
iii) We propose a novel approach for unsupervised object discovery based on a hierarchical clustering of the mined object appearances in a learned embedding space. 
iv) We analyze the effort needed and the yield obtained by object mining in realistic scenarios and we provide a detailed evaluation on the number and type of object categories that can be discovered this way.
v) We provide the KITTI Track Collection (KTC), a set of automatically generated tracks which we manually categorized into 33 categories.

\begin{figure*}[ht!]
\begin{center}
\includegraphics[width=0.9\linewidth]{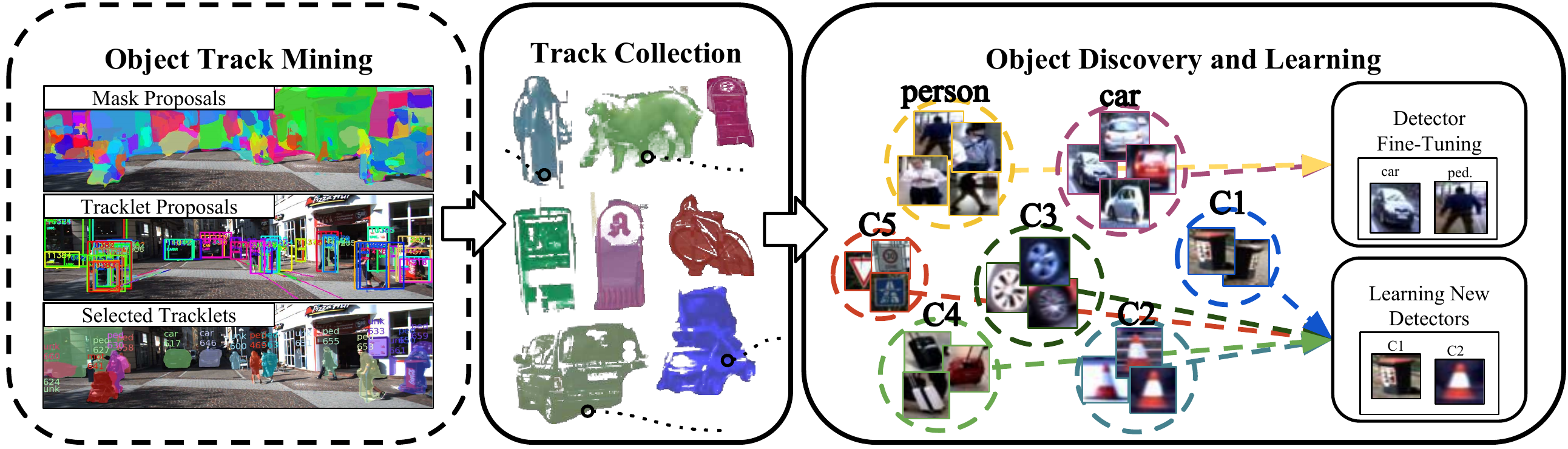}
\end{center}
\caption{We mine unlabeled video sequences using a category-agnostic tracker and create a database of mined tracks. We use these tracks to fine-tune detectors (using known categories), discover novel, previously unknown object categories, and learn detectors for them.}
\label{fig:pipeline}
\end{figure*}

\section{Related Work}

In the following, we provide a brief overview about semi-, weak- and self-supervised learning, clustering, embedding learning, and object discovery.

\PARbegin{Weakly-, semi- and self-supervised Learning.} Semi-supervised learning methods assume datasets that are only sparsely labeled \cite{Zhu02, Joachims99ICML}.
These methods are often utilized in robotics applications, where it is not feasible to densely label data for every deployment scenario. Similar to us, \cite{Teichman2012IJRR} use an object tracker in street scene scenarios to extend sparse labels across LiDAR segments and improve object classification performance.
\cite{Tao15IROS} propose an online semi-supervised learning system that combines online star-clustering with a  label propagation algorithm \cite{Zhu02}. 
All of these methods are based on streams of clean LiDAR data, where even simple methods can be used to segment scans into meaningful regions. Obtaining good object candidates only from image data is a more challenging problem with long research history \cite{Hosang14BMVC, Zitnick14ECCV, Alexe12TPAMI, Chen15NIPS, Osep16ICRA}.
\cite{Misra15CVPR} propose a vision-based approach that uses sparse car annotations in a video stream and mines additional training data using tracking. However, their method uses a static (surveillance) setup for mining, while future cars or robots will need to automatically learn from data obtained in cluttered scenarios, captured from a mobile platform.

Instead of using any form of human supervision, self-supervised learning methods often use other sensory modalities to obtain a supervisory signal, \eg for road and obstacle detection \cite{Dahlkamp06RSS, Lookingbill07IJCV, Wang17ICRA}. Alternatively, an additional supervisory signal can also come in the form of raw pixels \cite{Doersch15ICCV, Doersch17ICCV, Pathak16CVPR}, depth and multiple-views \cite{Sermanet17Arxiv,Zeng16ICRA, Vijayanarasimhan17Arxiv}, video \cite{Wang15ICCV, Misra16ECCV}, and even text \cite{Gomez17CVPR}.
Our method similarly uses video, depth, and coarse geometric information as additional supervisory signals, provided by the multi-object tracker.
 
In context of weakly-supervised learning, 
\cite{Singh16CVPR} propose a method which generates pseudo ground-truth bounding boxes using weakly annotated videos and images.
\cite{Prest12CVPR} propose to train new detectors from videos by localizing the dominant object in each video. 
Both assume that in general there will be one object of a given label 
per video. In our scenarios, there will always be several objects of different categories, often occluding each other, which makes the learning problem significantly more challenging.

\PAR{Clustering and Embedding Learning.}
Clustering is typically used to find patterns in unlabeled data by grouping data points by their similarity. Here the main challenge is defining similarities or distance measures between the data points. Recent methods approach this problem by learning distance metrics \cite{Xie16ICML, Schroff15CVPR, Song17CVPR}. 
A recent approach by \cite{Hsu16ARXIV} presents an end-to-end clustering framework. They use a labeled dataset to learn a Similarity Prediction Network (SPN),
which is then used to produce binary labels for each pair of objects of an unlabeled dataset. These labels are 
used to train ClusterNet, which directly predicts cluster labels.
However, all these methods perform clustering on full images and require pre-processing to obtain clusterings of objects \cite{Song16CVPR, Coates2011AISTATS}, \ie objects of interest need to be manually cropped. In contrast, we perform clustering based on bounding boxes produced by our tracker, which enables us to localize objects and directly tackle object discovery.

\PAR{Object Discovery.} Object discovery denotes the problem of identifying previously unseen object categories without human supervision.
\cite{Russell06CVPR} propose a vision-based method that uses multiple object segmentation in order to group visually similar objects and their segmentations. \cite{Lee10CVPR} demonstrate that recognition in the form of a region classifier helps with the discovery by narrowing down object candidates. 
We similarly utilize multiple object hypotheses in the form of tracklet proposals and utilize a classifier that assigns semantic information  to tracklets.
\cite{Rubinstein13CVPR} explore object discovery using Internet images. They propose to identify potential objects using saliency methods and find reoccurring patterns between images using dense correspondences. For a more detailed overview of existing vision-based methods, datasets and evaluation techniques we refer to \cite{Tuytelaars10IJCV}.

\cite{Kwak15ICCV} propose a method for joint tracking and object discovery in videos. Their method localizes and tracks the dominant object based on motion and saliency cues in each video. A similar idea is applied in \cite{Tsai16ECCV} for semantic co-segmentation in videos. To localize objects, their approach uses semantic segmentation and object proposals.
Both methods demonstrate excellent results on the YouTube-Objects dataset \cite{Prest12CVPR}. However, these video sequences are usually dominated by a single object. 

In the field of mobile robotics, \cite{Endres09RSS, Triebel10RSS, Herbst11IROS} propose methods, in which RGBD scans are segmented into object candidates. These candidates are then grouped using either clustering methods or methods based on probabilistic inference.
However, all of these methods were only applied to simple indoor scenarios, containing well-separated objects such as boxes and chairs or objects placed on tables (mugs, milk, \etc) and it is unclear how they would scale to more realistic scenarios.
\section{Method}

Towards the goal of object discovery and detector adaptation using unlabeled video sequences, we first need to be able to mine these video streams for potential object candidates. To this end, we utilize our multi-object tracker \cite{Osep17ARXIV} (Fig.~\ref{fig:pipeline} \textit{left}), that uses category agnostic object proposals \cite{Pinheiro16ECCV} and depth as input and produces tracklets, marked as either ``known`` (annotated in COCO \cite{Lin14ECCV}) or ``unknown`` object (all the rest).
%
%
%

These object candidates form our track collection (Fig.~\ref{fig:pipeline} \textit{center}).
As a main contribution of this paper, we propose a method that uses this collection to perform object discovery and to find groups of reoccurring objects. In addition, we propose a method for detector fine-tuning using these noisy object tracks of known categories (Fig.~\ref{fig:pipeline} \textit{right}). Furthermore, we show preliminary results on the task of training new detectors using the groups of discovered objects.

\subsection{Object Track Mining}
\label{subsec:obj-track-mining}
For video-based data mining it is important to narrow down the search space for potential objects. State-of-the-art object proposal methods such as \cite{Pinheiro16ECCV} need to produce 100-1000 proposals per frame to achieve a high recall, which would result in a very large set of object candidates on the level of an entire video.
Instead, we resort to our tracking-before-detection system \cite{Osep17ARXIV}, specifically designed for the object mining task.
Note that state-of-the-art tracking-by-detection systems would not be suitable for the task, as these can only track objects for which detectors are available.

In a nutshell, object candidates are obtained using our tracker as follows.
Input to the tracker are frame-level mask proposals \cite{Pinheiro16ECCV}.
The tracker uses these mask proposals in order to create a set of category-agnostic tracklet proposals, \ie no semantic information is used at this stage. 
Afterwards, we use the classifier component of a Faster R-CNN \cite{Ren15NIPS} based detector to classify the tracklet proposals on an image crop level. 
This way we transfer semantic information from a pre-trained detector about proposal tracklets that the pre-trained detector can recognize. 
The remaining tracklet proposals are labeled as unknown potential objects.
Finally, for each frame, a mutually consistent subset of tracks is picked by performing near-online MAP inference using a conditional random field (CRF) model (for details, see \cite{Osep17ARXIV}).
This way, we obtain a reduced set of tracklets, \ie the selected tracklets. 

\PAR{Towards Tracking-based Video Mining.} 
Since model selection is performed on a per-frame basis, one object may be split into several (often overlapping) tracklets. 
As an example, it may occur that an object is represented by a poorly-localized track while it is far away from the camera. As objects get closer to the camera, the tracker switches to a better localized tracklet, effectively refining its explanation of the observed cues as more information is available. 

To compensate for this behavior, we perform a final postprocessing step. We progressively merge short selected tracklets into tracks. In each frame, either i) the existing tracklet $h_i$ is re-selected and trivially continues the existing track $H_k$, or ii) tracklet $h_i$ is not re-selected and its track is continued by another selected tracklet $h_j$ if they have a sufficient overlap. If $h_i$ and $h_j$ do not have sufficient overlap, $H_k$ is terminated and $h_j$ starts a new track. In order to measure the amount of overlap, we look at the masks provided by SharpMask, which are associated to each track. Two masks are considered to be a match when the mask intersection-over-union (IoU) is higher than a threshold $\gamma$ in frame $t$. 
As an overlap criterion, we use the quotient between the number of matching masks and the length of the shorter tracklet:
\begin{equation*}
\lambda\! \left ( h_i, h_j \right ) = \frac{\left | \left \{  t  |  IoU \left ( h_i^t, h_j^t \right )  > \gamma \right \} \right |}{\text{min} \left ( \left | h_i \right |, \left | h_j \right |  \right )}.
\end{equation*}
%
%
In this case, the $\left | \cdot \right |$ operator denotes the tracklet length. Intuitively, this simple criterion measures how much of the shorter tracklet is covered by the longer one. This way, we obtain the final tracks that constitute our track collection.


\subsection{Object Discovery}
\label{subsec:category-discovery}

After running the tracker on several datasets, we obtain a reduced set of sequence-level object candidates, \ie tracks. Each object track is either classified as one of the categories defined in the COCO dataset \cite{Lin14ECCV} or marked as unknown. 
Next, we aim to find patterns in the track dataset by clustering the mined tracks. Such a clustering will result in i) groups dominated by already known objects (\eg pedestrians or cars), and ii) groups of unknown objects. We aim to utilize the ``known clusters`` for detector fine-tuning and the ``unknown clusters`` for learning detectors for new, previously unknown objects. This is a challenging problem: the first question is how to define a similarity measure between the object tracks. Second, object tracks will always contain outliers and occasionally imprecise localization of objects. Typically, partitional clustering methods will simply assign outlier tracks to their nearest clusters, which is undesirable. In the following, we address both issues.

\PAR{Track Similarity Measure.} The central challenge in clustering is defining a distance measure between data points (in our case, object tracks). Rather than hand-crafting a similarity measure, we aim for a data-driven approach and make use of the recent advances in the area of feature embedding learning \cite{Weinberger09JMLR, Schroff15CVPR, Song17CVPR, hermans2017defense}. We propose to cluster the tracked objects by their appearance using a learned feature embedding network for defining a distance measure. 
For clustering, we utilize the hierarchical density-based clustering algorithm HDBSCAN \cite{Campello15TKDD} due to its scalability to large datasets and its inherent ability to deal with outliers in the data. We show in Sec.~\ref{subsec:clustering-eval} that this approach outperforms simpler alternatives. As a distance measure between tracked objects, we use the Euclidean distance in the learned embedding space.

The feature embedding network we propose is based on a wide ResNet variant with 38 hidden layers \cite{wu2016wider} pre-trained on ImageNet \cite{Deng09CVPR}. We train it on the COCO detection dataset \cite{Lin14ECCV} using cropped detection bounding boxes which we resized bilinearly to $128\times128$ pixels as inputs. We apply a triplet loss \cite{Weinberger09JMLR} to learn an embedding space in which crops of different classes are separated and crops of the same class are grouped together. To this end, we adopt the batch-hard triplet mining and the soft-plus margin formulation of \cite{hermans2017defense} (for more details see supplementary material). We trained the network to discriminate between the 80 object classes in the COCO dataset, however, our experiments show that the learned embedding can generalize beyond the 80 categories.

When applying the embedding network on tracks, we first extract a representative embedding vector for each track. We take the embedding vector of the crop that is closest to the mean of the embedding vectors of the tracklet`s crops. This proved to be more robust than directly taking the mean. We then cluster these representative embedding vectors using the HDBSCAN algorithm and transfer the resulting cluster label to the whole tracklet.

\subsection{Detector Fine-tuning}
\label{subsec:det-fine-tuning}
The goal of detector fine-tuning is to use mined object tracks of known categories in order to adapt a pre-trained detector to a new domain.
To this end, we can utilize additional training samples that the tracker provides via reasoning on the sequence level. However, the mined tracks will also contain outliers and occasionally bad localization of the tracked objects, especially at the farther distances. In the following, we describe a novel approach for robust detector fine-tuning using such noisy tracks. For this task, we start with a detector using the same categories as used in Sec.~\ref{subsec:obj-track-mining}
and use the bounding boxes and class labels provided by the mined tracks.

\PAR{Detector Architecture.}
We fine-tune a state-of-the-art Faster R-CNN \cite{Ren15NIPS} based detector with an Inception-Resnet-v2 \cite{Szegedy17AAAI} backbone with weights pre-trained \cite{Huang17CVPR} on COCO.
Note that this is the same network which we used to classify the proposals in Sec.~\ref{subsec:obj-track-mining}.

Faster R-CNN consists of two stages. The first stage is a region proposal network (RPN), which uses a large number of anchor boxes which are refined and scored to produce category-agnostic object bounding box proposals. In the second stage, these proposals are cropped, resized, and classified.

\PAR{Subselection of Anchors.}
\begin{figure}[t]
\begin{center}
   \includegraphics[width=1.0\linewidth]{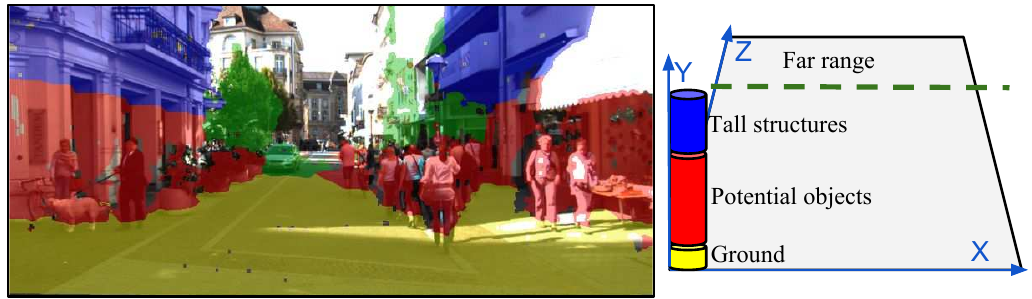}
\end{center}
   \caption{We use geometric information derived from stereo and a ground-plane estimate. The yellow and blue masks contain ground pixels and tall structures above 2.5 m height. We utilize both of these regions to sample negative training examples, since they should not contain any object. Objects in regions which are far away from the camera (green) or which do not have reliable depth measurements cannot be robustly tracked. We exclude these regions from training. The red region may contain objects and is used to sample tracked objects in this region as positive examples.}
\label{fig:geometric_info}
\end{figure}
The RPN uses a large number of anchor boxes which cover the whole image. For each training step, a random subset of the anchors is selected as training examples using for each anchor either a positive label as target if the anchor box overlaps enough with a ground truth object or a negative label otherwise. When working with manually annotated data, using all anchor boxes as potential training examples is perfectly reasonable, but when obtaining annotations automatically by tracking, this strategy can introduce errors since objects can be missed by the tracker or be assigned an unknown class label.

To circumvent this problem, we propose to only select anchors for which we are sure about their correct label and to ignore other anchors during training. In particular, we retain anchor boxes which have an IoU of more than $50\%$ with a bounding box of an object which has been recognized as ``known``. We ignore objects which have been tracked but classified as unknown, by not using them to select anchor boxes, as we do not know their true label.

In order to get a sufficient number of negative examples, we exploit simple geometric knowledge (see Fig.~\ref{fig:geometric_info}). The tracker provides a ground-plane estimate, which we use to create a mask which contains the ground-pixels and the area that spans more than 2.5m above the ground.
We use this mask to find regions in which we assume no objects to be located. We exploit this information by retaining anchor boxes for which at least $90\%$ of their area is contained in the mask. These anchor boxes will then be used as negatives.

One particular problem is that our tracker does not reliably track objects which are far away from the camera (beyond 30m), such that these objects would incorrectly be used as negative training examples. To circumvent this problem, we use the stereo depth information to exclude anchor boxes from training if at least $50\%$ of their pixels are more than a threshold (\eg 30 meters, depending on the dataset) away from the camera. 
%

\section{Experiments}

To evaluate our method we consider three different datasets recorded from mobile platforms.

\subsection{Datasets}
\label{eval:datasets}
The KITTI \cite{Geiger12CVPR} and KITTI Raw \cite{Geiger13IJRR} datasets have been recorded in street scenes from a moving vehicle in Karlsruhe, Germany. For our experiments we only use the stereo cameras and a subset of 1.18 h of video data ($42,\!407$ frames, for details see supplementary material).

The Oxford Robotcar dataset \cite{Maddern17IJRR} has a similar setup as KITTI and it has been collected from a mobile vehicle in street scenes, mainly in the inner city of Oxford, England. In our experiments, we only used the stereo setup.
In total 1,000 km have been recorded over 1 year, from which we use a small but representative subset of 3.72 h of video ($214,\!024$ frames).

The third dataset was recorded at the Amsterdam-Schiphol airport and is part of a large private dataset that was collected from a moving robot platform during the EU project SPENCER \cite{Triebel15FSR} and has partly been evaluated in \cite{Linder16ICRA}.
We use a subset 
of the data recorded from an Asus Xtion Pro Live RGBD sensor lasting 1.04 h ($112,\!723$ frames) in total.
The indoor-setting of the airport reaches from narrow hallways to wide open areas.
It is visited by around 150,000 passengers each day which leads to challenging crowded situations with many unknown objects. 


\subsection{KITTI Track Collection}
\label{subsec:track-dataset}
In this subsection, we describe and analyze the tracks we mined from the KITTI Raw dataset.
%
Our tracker takes 100 mask proposals per frame, of which $\sim$85 pass the geometric consistency checks in a typical inner-city sequence. The tracker internally maintains $\sim$97 tracklet proposals per frame, of which $\sim$13 are selected. Table \ref{tab:tracks-dataset} displays a short summary of the track mining for all three datasets. Even state-of-the-art object proposal approaches require an extremely large number of object candidates to achieve high recall for such sequences, rendering direct object discovery from proposals infeasible. Using tracking, we are able to reduce the number of object hypotheses to a manageable level and achieve a significant compression factor per image (\ie. from 100 object candidates per image to $\sim$13), and an even greater compression factor on the sequence level. With this procedure, we obtain a manageable number of object tracks.

For the purpose of a detailed analysis of tracks and clustering evaluation, we manually annotate tracks mined on the KITTI Raw dataset and obtain the KITTI Track Collection (KTC), which we will make publicly available. These annotations have only been used for evaluation of the clusterings. We have
not trained anything or optimized any hyper-parameters using these labels. We label each track as one of 33 categories which we manually identified in the tracks. Tracks that diverge from the tracked object are marked as a tracking error.
When the tracked object was recognized as a valid object but does not fit into any of the 33 classes, it was labeled as a valid unknown object.
Both the erroneous tracks and unknown tracks are excluded for the object discovery evaluation.

The largest annotated categories are \categoryname{car}, \categoryname{greenery}, \categoryname{window}, and \categoryname{person} with $2,\!405$, $1,\!124$, $370$, and $272$ instances, respectively. A tracking error only occurred in 745 tracks ($9.3\%$) which demonstrates the robustness of our tracker. Some of the categories which are not annotated in COCO are \categoryname{van}, \categoryname{trailer}, and \categoryname{rubbish bin}, for which we obtained 142, 45, and 100 instances, respectively. This demonstrates that our tracker can deliver tracks for interesting previously unseen categories, but the amount of data from KITTI Raw (1.18 h) is not yet sufficient to train robust detectors. For a detailed overview of annotated categories and additional statistics see the supplementary material.
%
%
%
\begin{table}[h!]
\footnotesize
  \begin{center}
    \begin{tabular}{l|r|r|r}
       & \textbf{KITTI Raw} & \textbf{Oxford} & \textbf{Schiphol}\\ 
		\hline
		Frames & 42,407 & 214,024 & 112,723\\
		Duration (h) & 1.18 & 3.72 & 1.04\\
		Per-Frame Proposals & 4,240,700 & 21,402,400 & 11,272,300\\
		All Tracklets & 173,822 & 788,877 & 265,173\\
		Selected Tracklets & 16,141 & 98,808 & 33,556\\
		Tracks (total) & 8,005 & 58,284 & 13,883\\
		Tracks (labeled) & 6,070 & - & - \\
		Tracks (unknown) & 1,190 & - & - \\
		Tracking Errors & 745 & - & - \\
    	\end{tabular}
    \vspace{5pt}
    \caption{Statistics of track mining from unlabeled videos. As evident, we achieve significant reduction using our tracking approach in comparison to per-frame object candidates, making object discovery from unlabeled video streams feasible.}
	\label{tab:tracks-dataset}
  \end{center}
\end{table}

\subsection{Object Discovery}
\label{subsec:clustering-eval}

\begin{figure}
\begin{center}
   \includegraphics[width=1.0\linewidth]{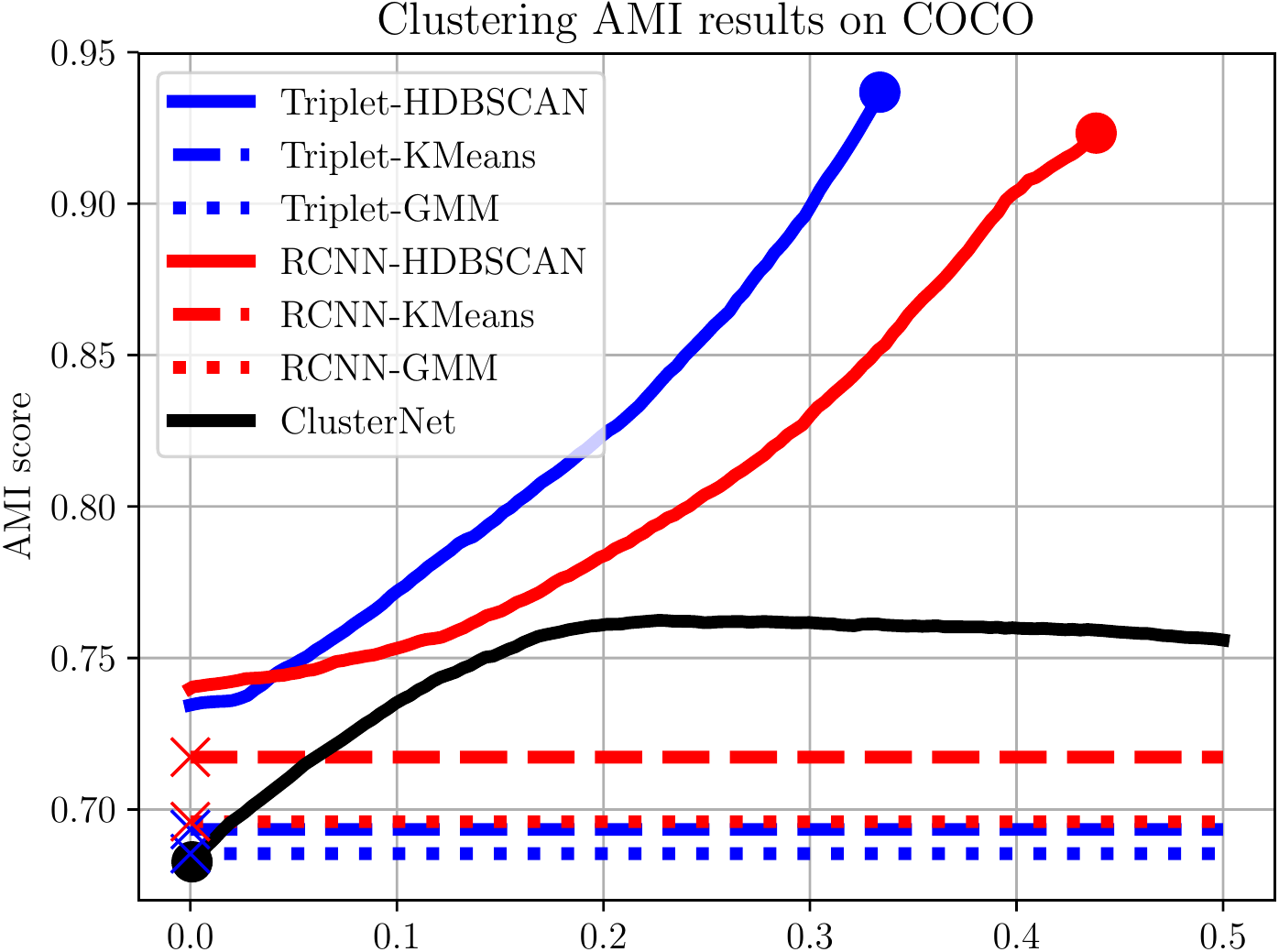}\\%
   \vspace{3pt}
   \includegraphics[width=1.0\linewidth]{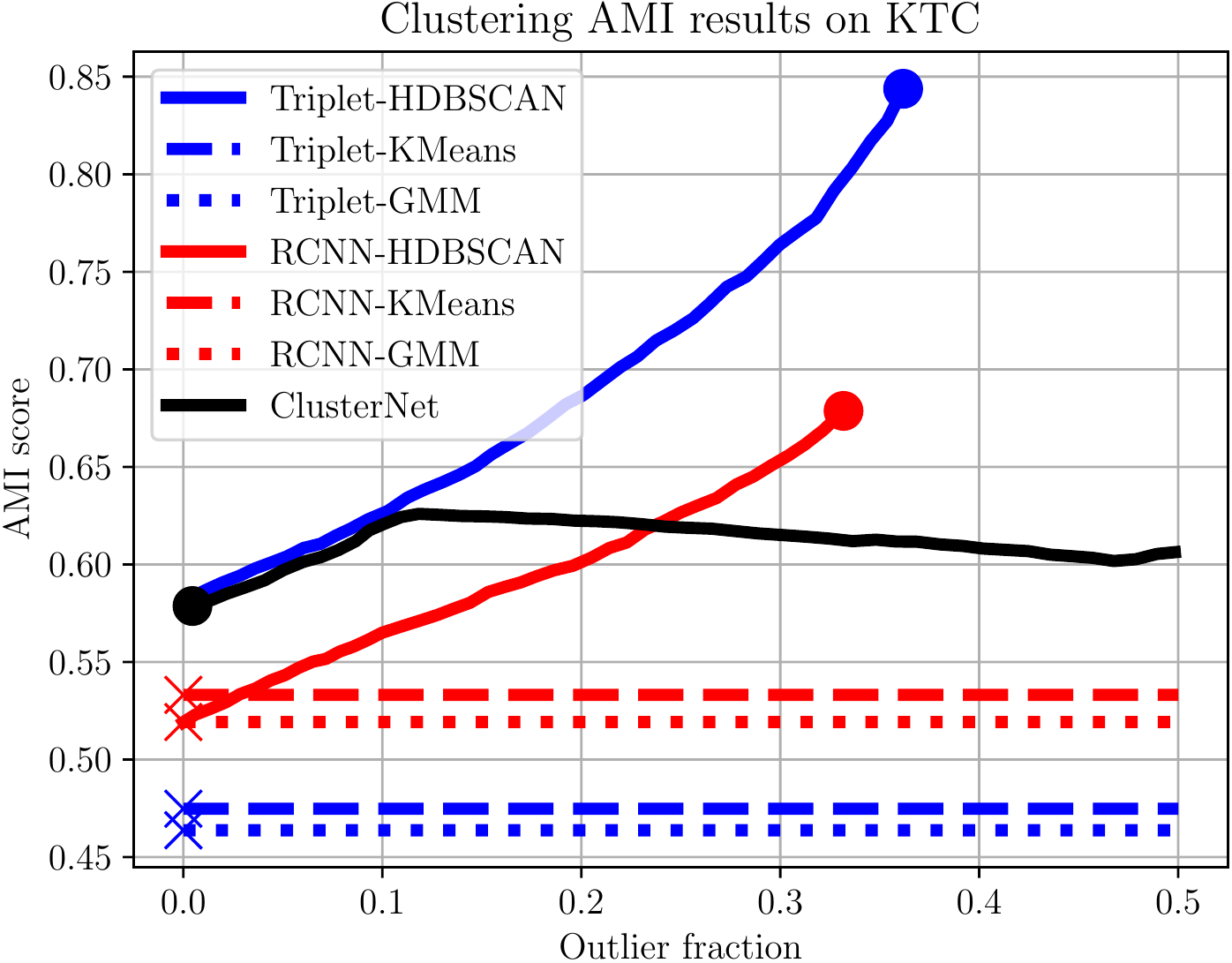}%
\end{center}
   \caption{Clustering results on COCO and KTC measured by AMI. Circle and Cross markers represent the clustering algorithms default setting. The methods shown with a circle have been evaluated with different outlier fractions to ensure a valid comparison between methods. Methods with a cross can not be extended for different outlier fractions, but are shown as a line for clarity.}
   \label{fig:clustering-eval}
\end{figure}
We evaluate the clustering quality using the adjusted mutual information (AMI) criterion which measures how well the clustering fits the ground truth classes. For additional evaluation measures and quantitative clustering results, see the supplementary material. The  results are shown in Fig.~\ref{fig:clustering-eval}.
Since HDBSCAN assigns an outlier label to a part of the data, we measure the performance as a function of the allowed fraction of outliers, which will be excluded from the evaluation. For algorithms which do not assign outlier labels or confidence values, we only show one data point without outliers in the graphs.

As the results show, HDBSCAN performs better for this task compared to other clustering approaches, including K-means \cite{lloyd1982least}, and a Gaussian Mixture Model (GMM) \cite{dempster1977maximum}. 
In addition, we demonstrate that the learned embedding (``Triplet``) outperforms a simple baseline which uses the L2-normalized activations of the last layer before the classification layer (``RCNN``) of our Faster R-CNN based detector (\cf Sec.~\ref{subsec:det-fine-tuning}).

As an additional baseline, we assess the performance of our reimplementation of ClusterNet \cite{Hsu16ARXIV} on our data. Here we trained the Similarity Prediction Network (SPN) \cite{Zagoruyko15CVPR} as a Siamese network with a two-class softmax using the same wide ResNet variant as the proposed feature embedding network. We trained the SPN on COCO to predict whether two input crops belong to the same class. We then used the output of the COCO-trained SPN to train a ClusterNet on the KITTI Raw tracklets to directly predict cluster labels.

The graphs show that on both datasets i) the learned embedding features (``Triplet``) are better suited for clustering than the features from Faster R-CNN (``RCNN``) and our proposed combination of the learned embedding and HDBSCAN outperforms both classical clustering algorithms which use the same embedding and the strong baseline provided by ClusterNet. 

\begin{figure}[t]
\begin{center}
\includegraphics[width=0.87\linewidth]{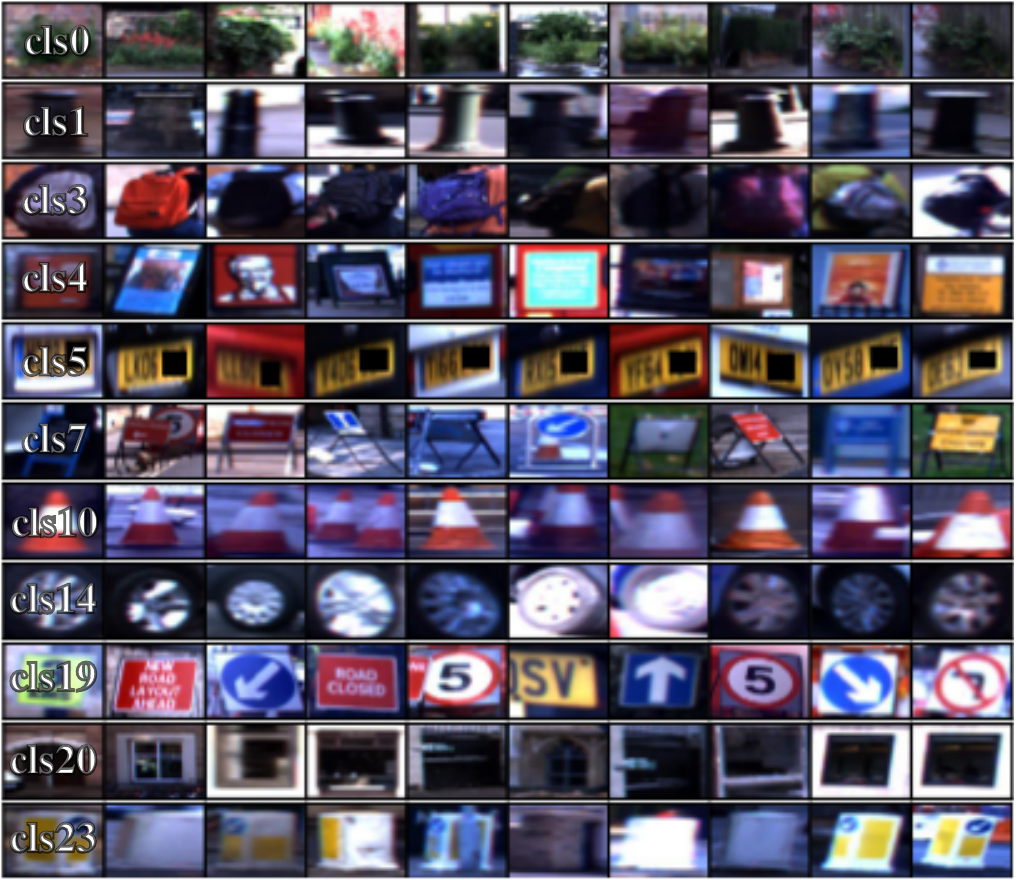} \\%
\vspace{3pt}
\includegraphics[width=0.87\linewidth]{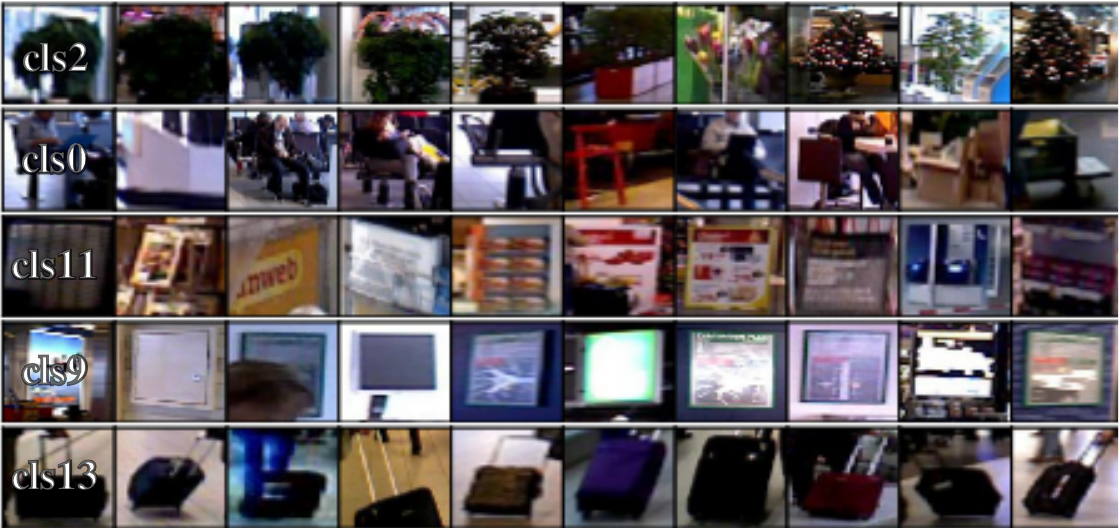}\\%
\end{center}
\caption{Visualization of some of the discovered objects in street scenes (Oxford Robotcar, \textit{top}) and Schiphol Airport \textit{(bottom)}.
Each row shows example crops for one cluster.}
\label{fig:oxford-clusters}
\end{figure}

We show qualitative results of a subset of obtained clusterings in Fig.~\ref{fig:oxford-clusters}. As evident, we obtain clusters for several object types, that are not present in the COCO dataset: \categoryname{bush}, \categoryname{pole}, \categoryname{advertisement}, \categoryname{license plate}, \categoryname{car wheel}, \categoryname{traffic cone}, \categoryname{christmas tree}, \categoryname{luggage} \etc. We show additional results in the supplementary material.

\subsection{Detector Adaptation and Learning}
\begin{figure*}[t!]
\begin{center}
\includegraphics[width=1.0\linewidth]{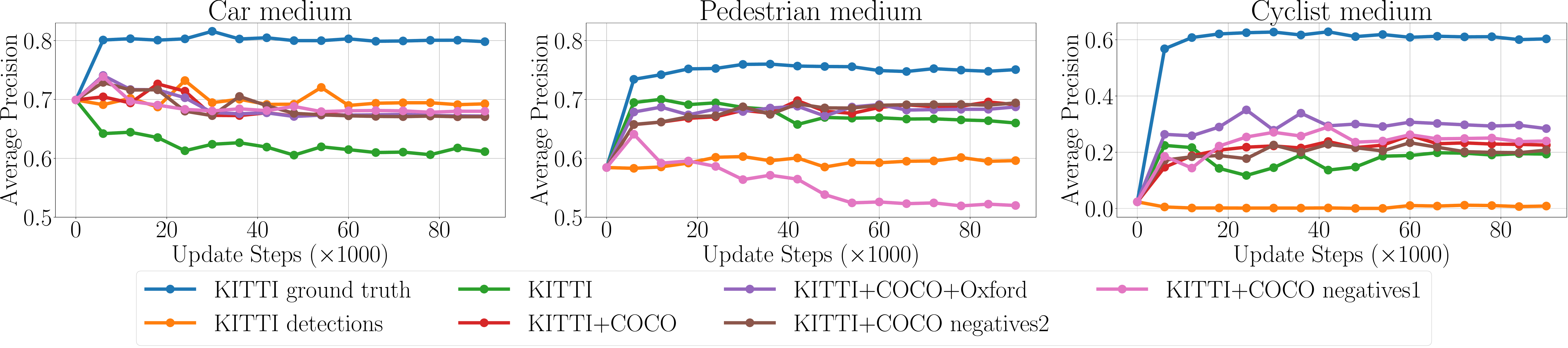}
\end{center}
\caption{Detection performance on the KITTI Detection validation set during fine-tuning with different training setups for the medium difficulty level. Each update step uses a single image. The setups are explained in detail in Sec.~\ref{subsec:clustering-eval}. Note the different axis range for cyclists. Best viewed in color.}
\label{fig:fine-tuning}
\end{figure*}
\PAR{Detector Fine-tuning for Known Categories.}
In the following, we show that using the automatically mined tracks and our anchor subselection method, we can adapt the detector to the automotive scenario of KITTI and bridge the gap between COCO's \categoryname{person} and KITTI's \categoryname{pedestrian} categories. 
To this end, we set up several experiments, where we start from the COCO pre-trained Faster R-CNN detector in all cases. We reuse the pre-trained output for COCO's \categoryname{person} category for KITTI's \categoryname{pedestrian} class, COCO's \categoryname{bike} class for initializing KITTI's \categoryname{cyclist} class, and keep the output for the \categoryname{car} class which is present in both datasets. 
To evaluate the detection performance, we used a validation set consisting of roughly half of the 7,481 images of the KITTI Detection training set which we split according to Chen \etal \cite{Chen15NIPS}. For fine-tuning we made sure not to use sequences of KITTI Raw in which images of the detection validation set are contained. 

\PAR{Baselines.}
Fig.~\ref{fig:fine-tuning} shows the results of all our detector fine-tuning experiments. Before delving into adapting our detector using automatically generated tracks, we consider several baselines to put our results into perspective. The first thing to note is that all setups start from the COCO pre-trained model and hence all curves start at the same point which corresponds to the performance of the pre-trained detector. The setup \setupname{KITTI ground truth} (blue curve) was fine-tuned on the ground truth annotations of the KITTI Detection training set. It provides a strong baseline which is very hard to beat since it was trained on manually annotated data, while we aim at improving the detector without manual annotations. We additionally create the baseline  \setupname{KITTI detections} (orange curve) for which we evaluate the COCO pre-trained detector on KITTI Raw and fine-tune it on its own predictions. To this end, we use all detections which are predicted with a confidence of more than a tuned threshold of $0.3$ as positive examples and do not apply any subselection on the anchor boxes. As can be seen in the plot, fine-tuning the detector in this way does not significantly change the detector performance. Hence, we will now investigate whether the tracking step can improve the quality of the training examples such that they are useful for improving the detector.

\PAR{Full Method.}
We will now evaluate the performance fine-tuning the detector on automatically generated tracks while using our novel anchor subselection method. When fine-tuning only on KITTI Raw (\setupname{KITTI}, green curve), the performance for \categoryname{cars} degrades, since the COCO pre-trained detector was already very strong on cars and we fine-tune with automatically obtained data which can contain errors. For \categoryname{pedestrians}, however, we can significantly improve over the COCO baseline.
For \categoryname{cyclists}, we created training data with the following simple rule: we merged tracklets, recognized as \categoryname{bicycle} and \categoryname{person} when their spatial distance was smaller than one meter, and marked it as \categoryname{cyclist}. In order to prevent the degradation of performance for \categoryname{cars}, we propose to jointly train on the tracks of KITTI and on the ground truth annotations of images of COCO which contain \categoryname{cars}. When training on multiple datasets jointly, for each update step, an image from either of the considered datasets is sampled with equal probability, and for KITTI and Oxford we apply our anchor subselection method while on COCO all anchors are retained. When adding in COCO for joint training (\setupname{KITTI+COCO}, red curve), we can mostly avoid the degradation of performance of the \categoryname{car} category and obtain further improvements for pedestrians and cyclists. This experiment clearly demonstrates that we are indeed able to significantly improve the detector performance for multiple object categories without using any manual annotations.
When additionally adding Oxford data for joint training (\setupname{KITTI+COCO+Oxford}, purple curve), has no significant effect on the performance for \categoryname{cars} and \categoryname{pedestrians}, but further boosts the performance for \categoryname{cyclists} which are more common in the Oxford dataset.

\PAR{Subselection of Anchors.}
Next, we investigate the effectiveness of our proposed anchor subselection, which was used for the former setups. We provide two variants of \setupname{KITTI+COCO} with alternative ways to subselect the anchor boxes.
For \setupname{KITTI+COCO negatives1} (pink curve), we do not only retain anchors which cover pixels on the ground or tall structures as before, but also in the potential objects region (\cf Fig.~\ref{fig:geometric_info}). This means that objects in the potential objects region which were not picked up by the tracker can be used as negative examples.
As can be seen in the figure, the detection performance for pedestrians degrades significantly. For the \categoryname{car} category, however, this is not the case, since the pre-trained detector already provided very accurate classifications, so that selecting examples in a simpler way does not introduce many errors. The performance for cyclists is less sensitive to the choice of anchors. This is probably because our automatic way of merging persons and bicycles into cyclist provides very unclean data with low recall, so that the details of the fine-tuning procedure only have a limited effect. The result for \categoryname{pedestrians} clearly shows the effectiveness of our anchor box subselection scheme based on geometric information.

The setup \setupname{KITTI+COCO} \setupname{negatives2} (brown curve) differs from \setupname{KITTI+COCO} by that we do not exclude anchors containing tracks classified as \categoryname{unknown} anymore. These tracks are then potentially used as negatives, although they might contain valid but unrecognized objects of the categories we consider. The results show that this change of the anchor selection only has a very limited effect. This means that using unknown objects as negatives conceptually sounds dangerous, but does not introduce many errors in practice.
%
%
The above experiments demonstrated the usefulness of the mined tracklet data and the effectiveness of the proposed anchor subselection method for improving the detector performance for known categories.
In the following, we will show that it is also possible to learn a detector for the automatically discovered categories without any form of manual labeling.

\begin{figure}[t]
\begin{center}
\includegraphics[width=0.87\linewidth,trim=0 255 0 135,clip]{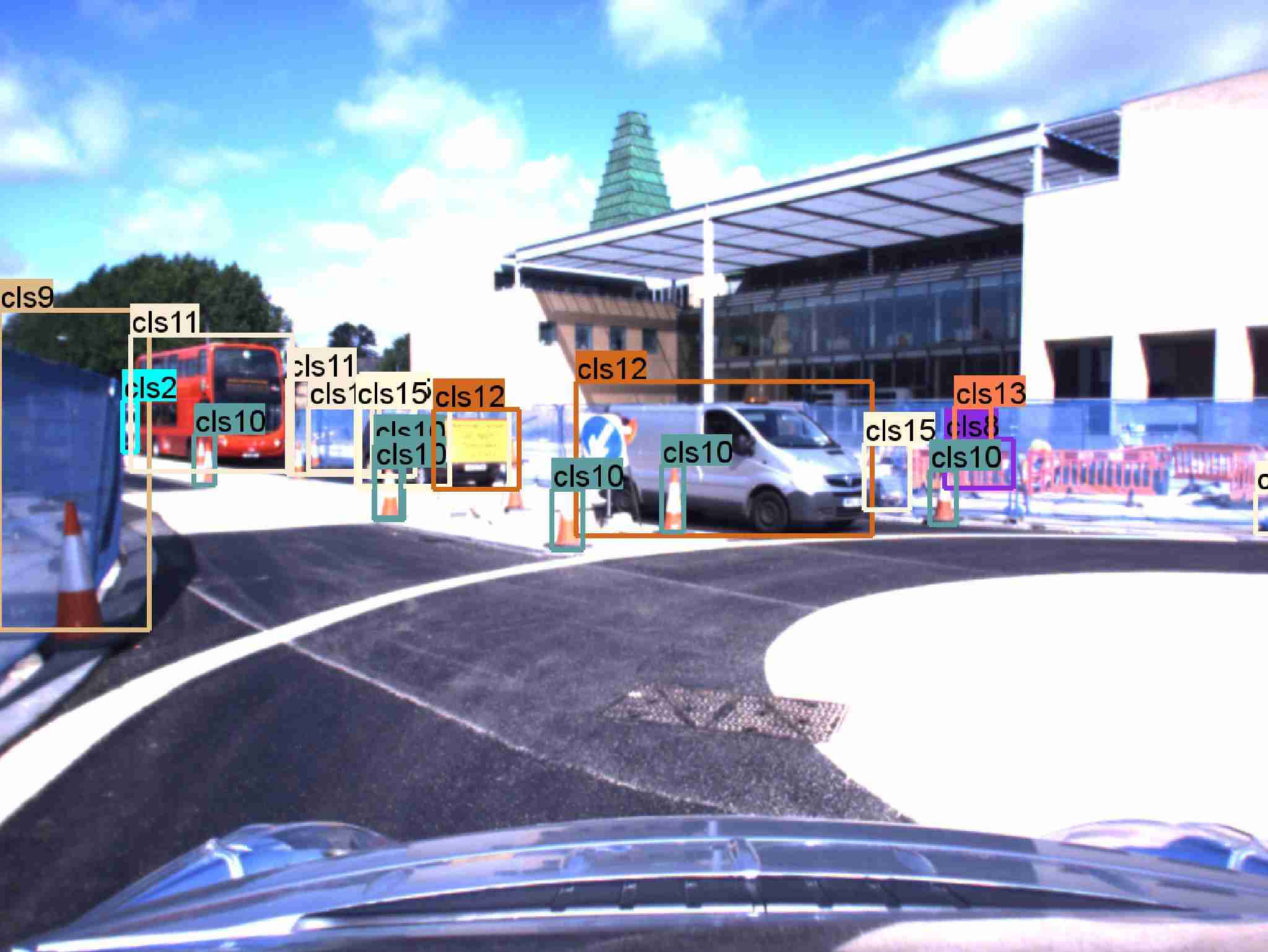}\\%
\includegraphics[width=0.87\linewidth,trim=0 255 0 135,clip]{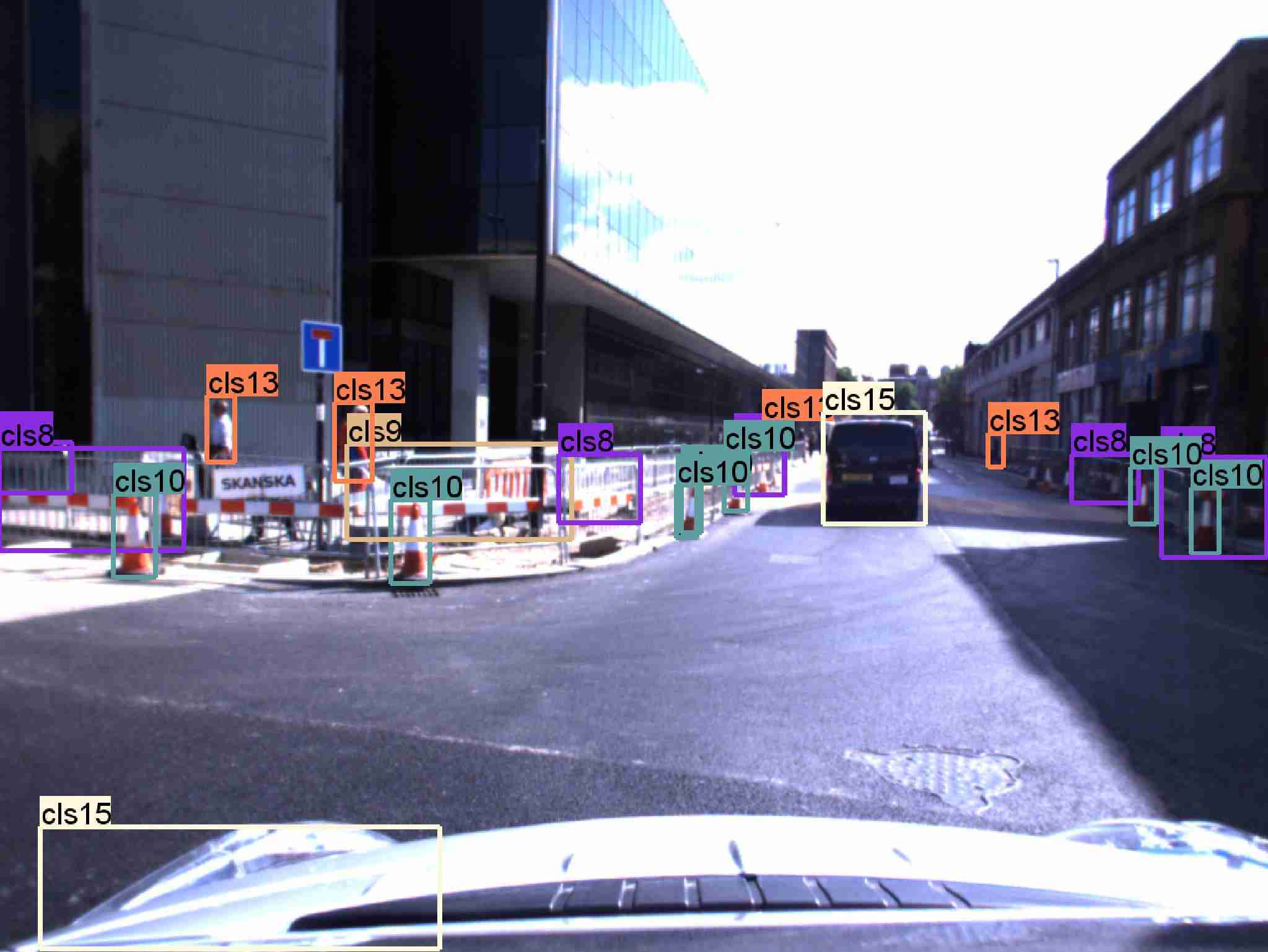}%
\end{center}
\caption{Qualitative results of learning a detector using the automatically discovered categories on Oxford. The shown labels (\eg cls12) denote indices of clusters (see Fig.~\ref{fig:oxford-clusters} for a visualization). The images have been slightly cropped.}
\label{fig:oxford-qualitative}
\end{figure}

\PAR{Detector Learning for Automatically Discovered Categories.}
By using the clustering results instead of the Faster R-CNN classification, we are able to learn a detector for previously unseen categories. 
To this end, we randomly initialize a new classification output layer with as many output units as the number of clusters and keep the COCO initialization for all other layers. Here we again use our anchor subselection method.
Fig.~\ref{fig:oxford-qualitative} shows qualitative results for learning a detector on the automatically discovered categories on Oxford. As can be seen, the detections for the newly learned categories are not yet at the same robustness level as the detections for established categories such as cars or pedestrians, but they show considerable promise for detecting interesting scenario-specific objects, such as traffic cones (cls10) or construction site fences (cls8).
%

\section{Conclusion}
This work is an initial study about learning from unlabeled data by automatically extracting generic object tracks. We showed that it is indeed possible to automatically adapt a pre-trained detector to a new scenario and that we can automatically discover previously unseen categories through clustering. Furthermore, we show initial qualitative results for training a detector for the automatically discovered categories.
We believe that this work is a starting point and there is still a large potential for further exploiting unlabeled data. In future work we will scale the amount of mined data up by at least one order of magnitude and investigate further application scenarios. In particular this huge amount of data will provide an opportunity to learn trajectory prediction just on the basis of appearance. Additionally, we expect that the quality and usefulness of an automatically created clustering can be significantly improved by manually merging clusters and removing outliers with minimal effort. 

\footnotesize {\PAR{Acknowledgments:} This work was funded by ERC
Starting Grant project CV-SUPER (ERC-2012-StG-307432). We would like to thank
Alexander Hermans, Wolfgang Mehner and Istv\'an S\'ar\'andi for helpful discussions}

{\small
\bibliographystyle{ieee}
\bibliography{abbrev_short,paper}
}

\clearpage
\appendix
\normalsize
\input{supplemental_raw.tex}
\end{document}

%% file: supplemental_raw.tex
\part*{Supplemental Material}
\noindent In this supplemental material, we provide:
\begin{itemize}
	\item Detailed statistics on the KITTI Track Collection 
	\item Implementation details of all stages of the system
	\item Additional experimental evaluations for the clustering
	\item Additional qualitative results
\end{itemize}

\section{KITTI Track Collection Statistics}
We created a KITTI Track Collection (KTC) by manually labeling the tracks mined from the KITTI Raw dataset \cite{Geiger12CVPR} using our category-agnostic multi-object tracker \cite{Osep17ARXIV}. We excluded sequences that overlap with the KITTI Detection dataset, calibration sequences and very short sequences from further processing, resulting in 42,407 frames and 1.18h of video. 

Tab.~\ref{Tab:allClasses} gives an overview of the object classes, labeled in the KTC. These 33 classes have been chosen after visually inspecting the data and identifying the most frequently appearing objects. Note that 18 of these classes are not present in the COCO dataset \cite{Lin14ECCV} and three only have similar, but more specific correspondences (greenery - potted plant, road sign - stop sign, animal - dog). These annotations have only been used for evaluation of the clusterings. We have not trained anything or optimized any hyper-parameters using these labels.

Beside object categories, we also labeled tracks as \categoryname{tracking error} and \categoryname{unknown object}. These have been excluded from the evaluation.
Tracks were labeled as \categoryname{tracking error} when an object was not tracked consistently, \ie more than 10\% of the crops diverged from the object (roughly, intersection-over-union was less  than 0.5).
Tracks were labeled as \categoryname{unknown object} when they represented valid objects, that did not fit into any of the 33 categories, \eg cigarette vending machines, phone booths, mobile toilettes, \etc.
\input{KTC_stats.tex}

\section{Implementation Details}

In the following, we provide implementation details for i) object detection fine-tunning, ii) feature embedding learning, iii) clustering, and iv) ClusterNet \cite{Hsu16ARXIV}.

\subsection{Detector}
\label{subsec:detector}
In the proposed method we utilize an object detector, pre-trained on the COCO dataset \cite{Lin14ECCV}. This detector is used for i) tracklet proposals classification (using the classification component of the detector only) and ii) as a base detector that we use for the detector fine-tuning experiments.

For the detector, we adopt the code and pre-trained weights provided by the TensorFlow detection API \cite{Huang17CVPR}. In particular, we use a state-of-the-art Faster R-CNN \cite{Ren15NIPS} based network with an Inception-ResNet-v2 \cite{Szegedy17AAAI} backbone with atrous (dilated) convolutions pre-trained on COCO (\textit{faster\_rcnn\_inception\_resnet\_v2\_atrous\_coco}), which achieves a mean average precision (mAP) of $37\%$ on COCO. Note that the proposed detector fine-tuning and automatic learning of newly discovered categories can also be done with different detector architectures.

In all experiments, we use a batch size of 1 image which was resized without changing the aspect ratio such that the smaller image dimension is 600 pixels, if the resulting larger image dimension is no more than 1024. Otherwise it is resized such that the larger image dimension is 1024 pixels.
The runtime depends on the image aspect ratio, but for typical images one update step takes around 0.7 seconds with a GTX 1080 Ti GPU. For optimization, we use Adam \cite{Kingma15ICLR} with an initial learning rate of $10^{-6}$ which is reduced to $10^{-7}$ after $60,000$ update steps. For each setup, we train for $90,000$ steps. For data augmentation, we use horizontal flipping and gamma augmentations \cite{Pohlen2017CVPR}.

\subsection{Feature Embedding Network}
%
For the object discovery task, we i) extract features from the cropped bounding boxes from each track and ii)
use these features to define a distance measure for clustering to group objects of same or similar categories and thus discover new categories that are present in the data.
For this task, we propose to use a trained feature embedding network to extract discriminative features for the cropped bounding boxes from each track. 

The feature embedding network we propose is based on a ResNet variant of \cite{wu2016wider} pre-trained on ImageNet \cite{Deng09CVPR}. This architecture uses only 38 hidden layers but with more units per layer than in the original ResNets \cite{He16CVPR}. It has roughly 124 million trainable parameters and achieves outstanding results on multiple datasets, which motivated us to adopt this architecture.
We replace the last layer with a 128 dimensional fully connected layer, for which the outputs are our embedding vectors. We train this network on the COCO detection dataset \cite{Lin14ECCV} using cropped detection bounding boxes which we resize bi-linearly to $128\times128$ pixels as inputs. 

We use a triplet loss \cite{Weinberger09JMLR} to learn an embedding space in which crops of different classes are separated and crops of the same class are grouped together. To this end, we adopt the batch-hard triplet mining and the soft-plus margin formulation of \cite{hermans2017defense}. We train this feature embedding network with a batch size of 64 images.
%
%
The employed ``batch hard formulation`` means that for each image in the batch we take this to be the anchor image, and we find the image in the same class for which its embedding vector is the furthest away (measured by Euclidean distance). This largest distance is then used as the anchor-positive distance in the triplet loss. 

Afterwards, we find the image from all the images of a different class in the batch that has the smallest Euclidean distance from the anchor image embedding vector, and use this as the anchor-negative term in the triplet loss. The loss we use is then simply the soft-plus of the difference between this anchor-positive distance and the anchor-negative distance. There are 64 contributions to this loss for each batch, as each image in the batch is used as an anchor image and these are averaged to get the final loss. The loss function including batch-hard triplet mining and the soft-plus formation is given by
\begin{equation}
\begin{split}
L(\theta,X) = \sum_{i=1}^P \sum_{a=1}^K g \big(\max_{p=1 \dots K} D \left(f_{\theta}(x_a^i), f_{\theta}(x_p^i) \right) \\
- \min_{\substack{j=1..P\\ n=1..K\\ j \neq i}} D \left( f_{\theta}(x_a^i), f_{\theta}(x_n^j) \right) \big),
\end{split}
\end{equation}
where $g$ is the soft-plus function $g(x)=\ln(1+\exp(x))$, $f_{\theta}$ is the learned embedding function with parameters $\theta$, $D$ is the Euclidean distance, $x_j^i$ is the $j$-th image for the $i$-th class, the index $a$ is over the $K$ images within each class and $i$ is over the $P$ classes within the batch. The maximization and minimization implement the batch-hard mining. For more details see \cite{hermans2017defense}.

Within each batch we sample randomly from 16 different categories, 4 images from each to make up the batch.
Our experiments show that this ratio of anchor-positive to anchor-negative pairs gives stable training results with different ratios and batch sizes causing the norm of the embedding space to either collapse into 0, or to expand without bound.
These findings also agree with what was posited in an earlier exploration of the triplet loss using a batch hard formulation and a soft-plus margin \cite{hermans2017defense}.

We trained this triplet based feature embedding network using Adam \cite{Kingma15ICLR} with an initial learning rate of $10^{-5}$ which was reduced to $10^{-6}$ after $1,500,000$ training samples (crops) and further reduced to $10^{-6}$ after $2,500,000$ training samples. In total, we trained this network for $5,000,000$ training samples.

\subsection{Clustering}

%
For our clustering experiments, we use two variants to create features which define the distance measures:
i) the proposed $128$ dimensional output of the triplet loss based network and ii) the $1,536$ dimensional L2-normalized activation vector of the last layer (before the softmax layer) of the COCO pre-trained Faster R-CNN \cite{Ren15NIPS} detector (\cf Section \ref{subsec:detector}). 
We evaluate clustering on both the hand-labeled KITTI Track Collection and on the COCO 'minival' set of 5000 images which is commonly used for validating detection performance.

For the evaluation on the COCO dataset, we use the annotated bounding boxes to extract object image crops.
When applying the embedding network on KTC, we first extract a representative embedding vector for each track. We take the embedding vector of the crop that is closest to the mean of the embedding vectors of the track`s crops. This proved to be more robust than simply taking the mean. We then cluster these representative embedding vectors and transfer the resulting cluster label to the whole track.

For the k-means and Gaussian Mixture Model (GMM) baselines, we use the scikit-learn \cite{Pedregosa11JMLR} Python implementation. The only parameter to set is the number of cluster centers. For these two baseline experiments we use the ground truth number of cluster centers, \ie $80$ for COCO, and $33$ for KITTI Raw. For the GMM we use the full covariance matrix variant.
For the HDBSCAN clustering algorithm we use the Python implementation of \cite{McInnes17OSS}. There are two parameters to be set, the minimum size of a cluster (\textit{minsize}) and the minimum number of samples in the neighborhood of a point for it to be considered a core point (\textit{minsamples}). As recommended by \cite{McInnes17OSS}, we set \textit{minsamples} to be equal to \textit{minsize}. The higher the \textit{minsize} parameter, the fewer clusters we obtain. We adjust this parameter by visual inspection of the resulting clusters. For COCO we set \textit{minsize} equal to $3$. For the KITTI Raw, Schiphol Airport and Oxford Robocar datasets we set \textit{minsize} to $14$. We obtain 13 clusters on the KITTI Raw dataset, 18 on the Schiphol Airport dataset, and 24 for the Oxford Robotcar dataset.

\subsection{ClusterNet}
\PARbegin{Architecture.}
ClusterNet \cite{Hsu16ARXIV} is a neural network architecture which can be trained using unlabeled data to directly predict cluster labels for each input image. In order to train it, we only need pairwise binary constraints, \ie for each pair of input crops we need a binary label which indicates whether both crops should be assigned to the same cluster. To obtain the binary labels, \cite{Hsu16ARXIV} propose to learn a  similarity prediction network (SPN) on a different dataset which is labeled. The SPN is a Siamese convolutional network with two branches with shared parameters. After the shared convolutional layers, the features of each of the two input crops are collapsed into a vector. Afterwards, the two collapsed vectors are concatenated and fed into a fully-connected layer. Finally, a two-class softmax is used to predict the probability of the two crops belonging to the same class. When training ClusterNet, the SPN is evaluated on the fly for each pair of image crops in the mini-batch and we obtain the binary labels by taking the argmax of the softmax output. ClusterNet uses a softmax output layer, where each output unit provides the probability that the input crop belongs to the associated cluster. For each pair of input crops, the KL-divergence between the two resulting output distributions is either maximized or minimized depending on the binary label. If the binary label indicates that both crops should be assigned to the same cluster, the KL-divergence is minimized, otherwise it is maximized using a margin. For more details see \cite{Hsu16ARXIV}.
The ClusterNet as well as the SPN are both trained utilizing every possible combination of pairs in the mini-batch to contribute toward the clustering loss, as was recommended in \cite{Hsu16ARXIV}.
\PAR{Specific Implementation.}
In our implementation, we adopt the same ResNet variant \cite{wu2016wider} with 38 hidden layers which we used for the feature embedding network for both the SPN and the ClusterNet. The SPN is trained on the ground truth annotations of COCO \cite{Lin14ECCV}, while ClusterNet is trained on the bounding boxes of automatically generated tracks on the unlabeled target dataset (\eg KITTI or Oxford). In both cases we resize the cropped bounding boxes to a fixed size of $128\times128$ pixels while changing the aspect ratio. Both networks are trained with a mini-batch size of 64 crops using Adam \cite{Kingma15ICLR} with an initial learning rate of $10^{-5}$ which is reduced to $10^{-6}$ after $1,500,000$ training samples (crops) and further reduced to $10^{-6}$ after $2,500,000$ training samples. For each network, we train for $5,000,000$ training samples in total. For data augmentation we use horizontal flipping and gamma augmentations \cite{Pohlen2017CVPR}. 
On the COCO dataset \cite{Lin14ECCV}, we use the ground truth number of cluster labels ($80$) in the output layer for training. On KITTI Raw \cite{Geiger12CVPR} we use $50$ cluster labels.


\section{Additional Experimental Evaluation}
\subsection{Object Discovery (Clustering)}

\begin{figure*}[t]
\begin{center}
   \includegraphics[width=0.45\linewidth]{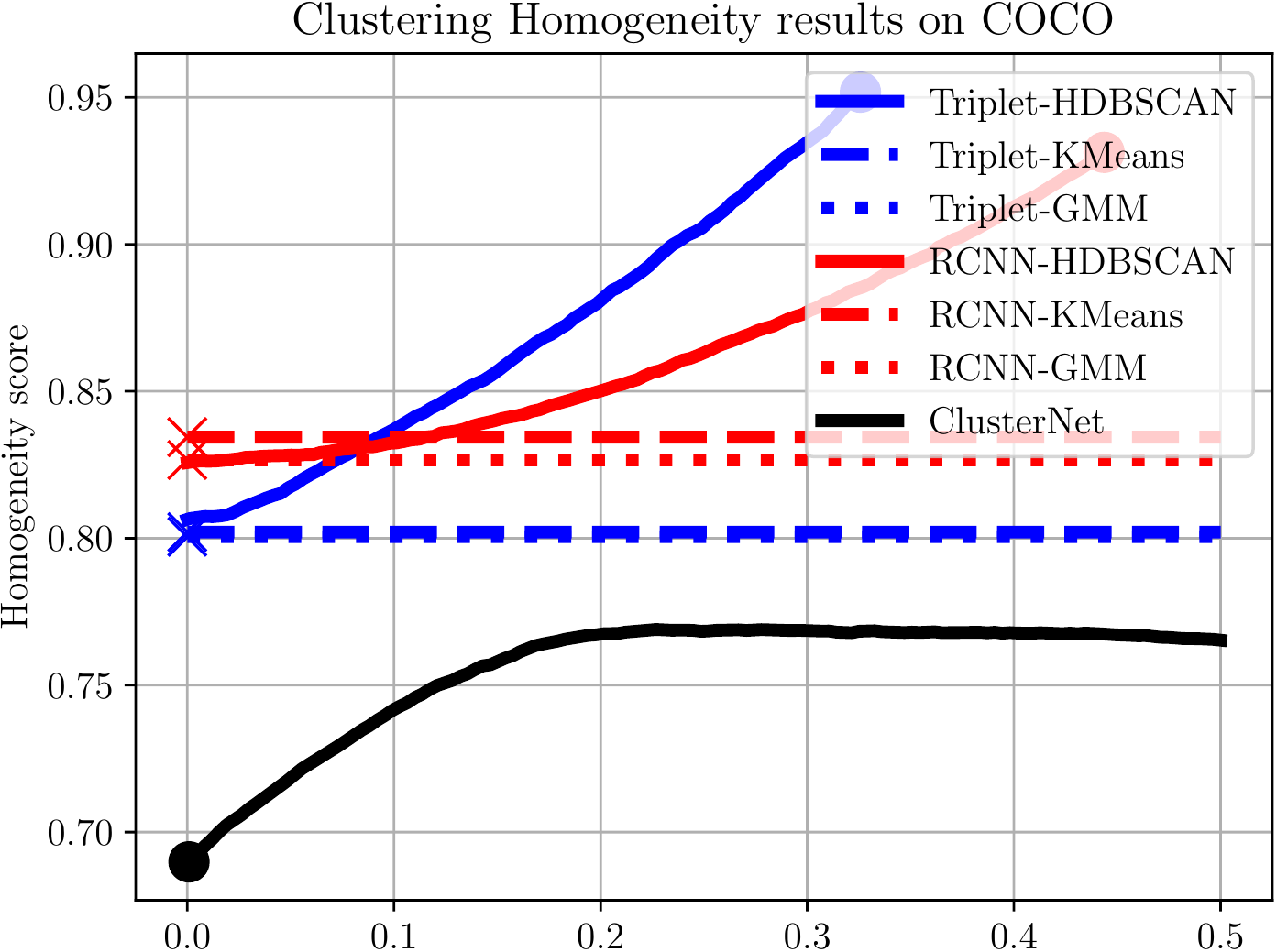}%
   \hspace{10pt}
   \includegraphics[width=0.45\linewidth]{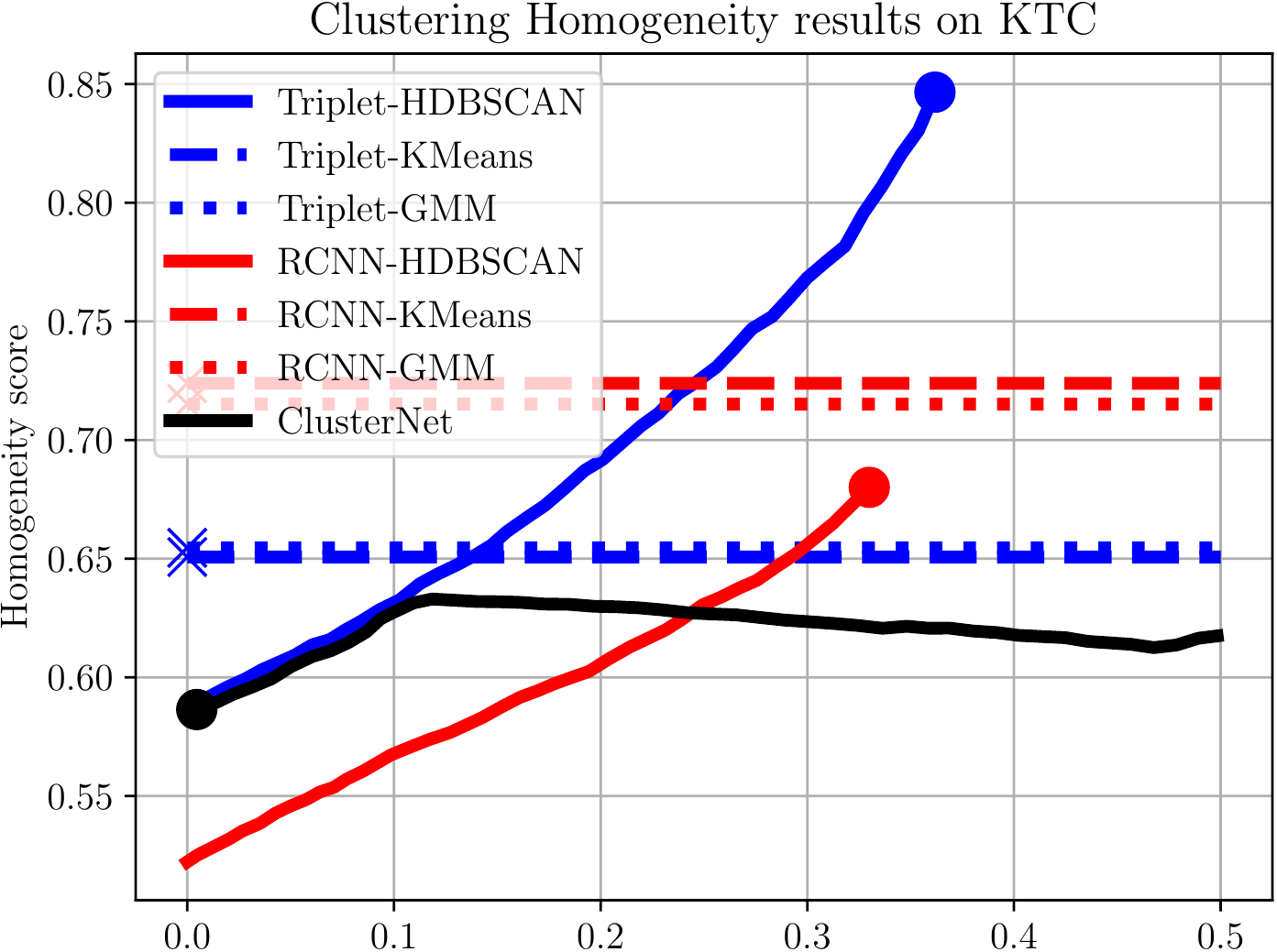}\\%
   \includegraphics[width=0.45\linewidth]{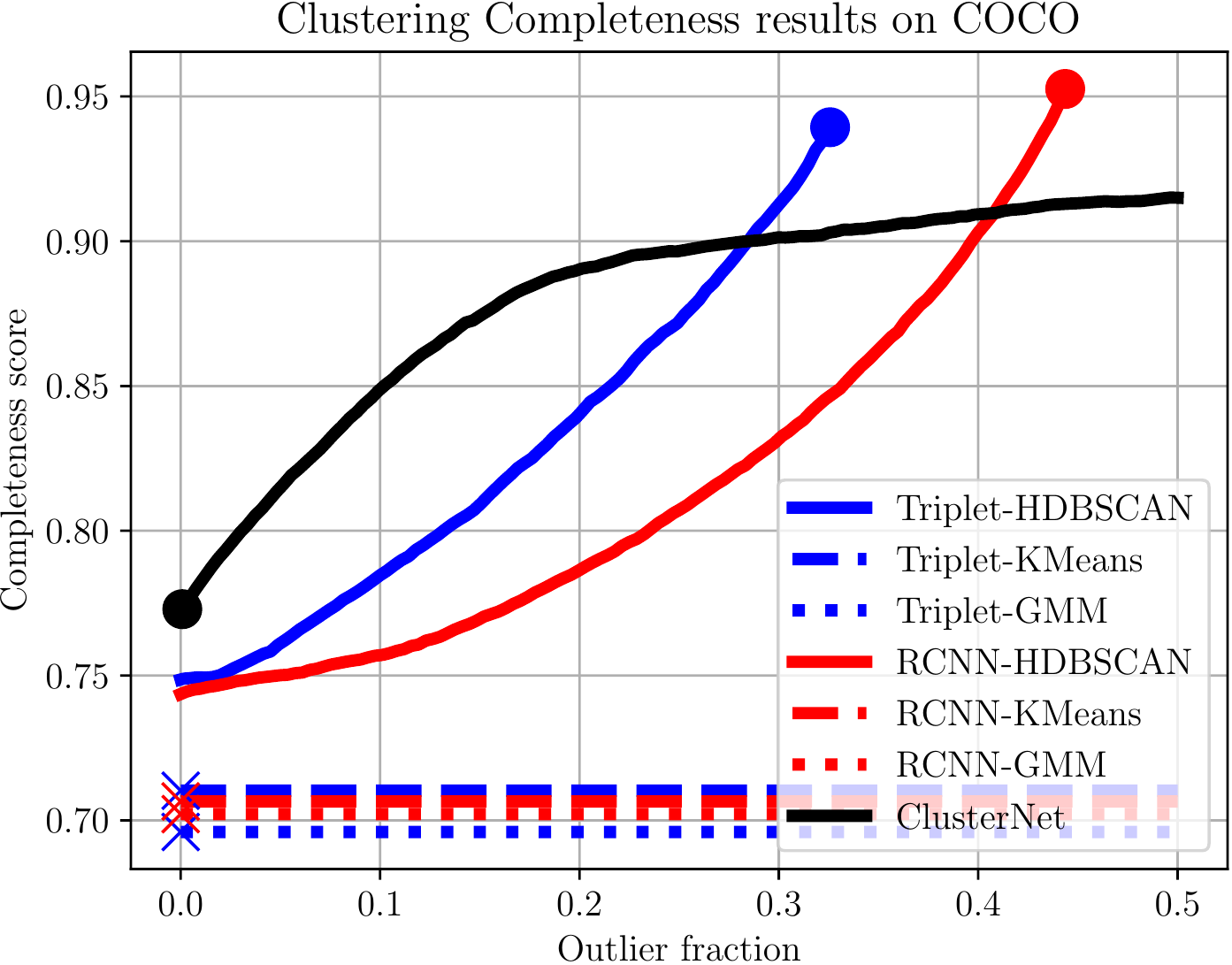}%
   \hspace{10pt}
   \includegraphics[width=0.45\linewidth]{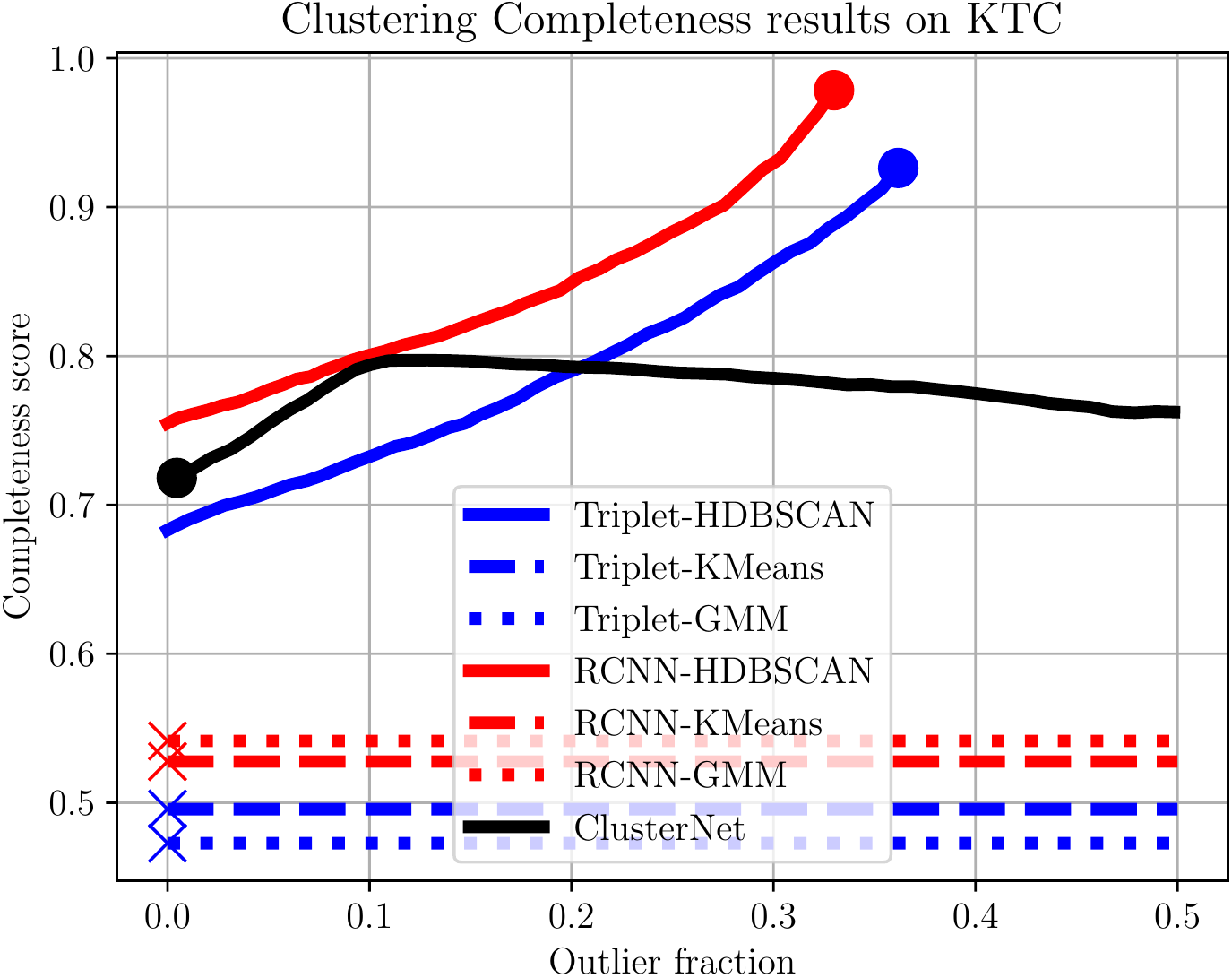}%
\end{center}
   \caption{Additional clustering results on COCO and KTC. Circle and cross markers represent the clustering algorithms' default settings. The methods shown with a circle have been evaluated with different outlier percentages to ensure a valid comparison between methods. Methods with a cross can not be extended for different outlier percentages, but are shown as a line for clarity.}
   \label{fig:add-clustering-eval}
\end{figure*}

We evaluate the performance of our clustering algorithms on both the COCO 'minival' dataset, as well as on our hand-labeled KITTI Track Collection (KTC). In Fig.~\ref{fig:add-clustering-eval}, we present quantitative results of our clustering method in terms of the homogeneity and completeness measures \cite{Rosenberg07}, which extends the AMI scores presented in the main paper.
Intuitively, homogeneity measures the purity of the clusters, \ie the fraction of the cluster members belonging to the dominant class of each cluster. Completeness on the other hand measures the fraction of instances of a dominant class, that are grouped together.

On both datasets, ClusterNet performs reasonably well, especially in terms of completeness, but quickly hits a ceiling when allowing more outliers.
For the k-means and GMM baselines, the RCNN features perform slightly better than the triplet features. For HDBSCAN, however, the triplet features provide a significant advantage except for the completeness score on KTC. 
For any setting of outliers, HDBSCAN outperforms the k-means and GMM baselines in terms of completeness. 

On COCO, we can achieve very high completeness and homogeneity since the feature embedding network is trained on the COCO classes. On KTC, the input data for clustering contains objects of novel categories which were not seen during the embedding network training. In this case, the performance drops by roughly $10\%$ absolute in terms of homogeneity, while the completeness only slightly degrades. 

While RCNN-HDBSCAN has a slight advantage in terms of completeness, for homogeneity, the proposed Triplet-HDBSCAN method clearly outperforms all other considered methods, which is preferable for learning new detectors for the discovered objects.
Overall, our proposed Triplet-HDBSCAN shows the most promising results.

\section{Additional Qualitative Results}

\PARbegin{Tracking.}
We show additional tracking results for the KITTI Raw, Oxford Robotcar, and Schiphol Airport datasets. These results are obtained by applying the first step of our pipeline, \ie the category-agnostic multi-object tracking. Displayed are objects picked after performing classification of tracks and performing inference.

\begin{figure*}[ht]
\includegraphics[width=0.50\linewidth]{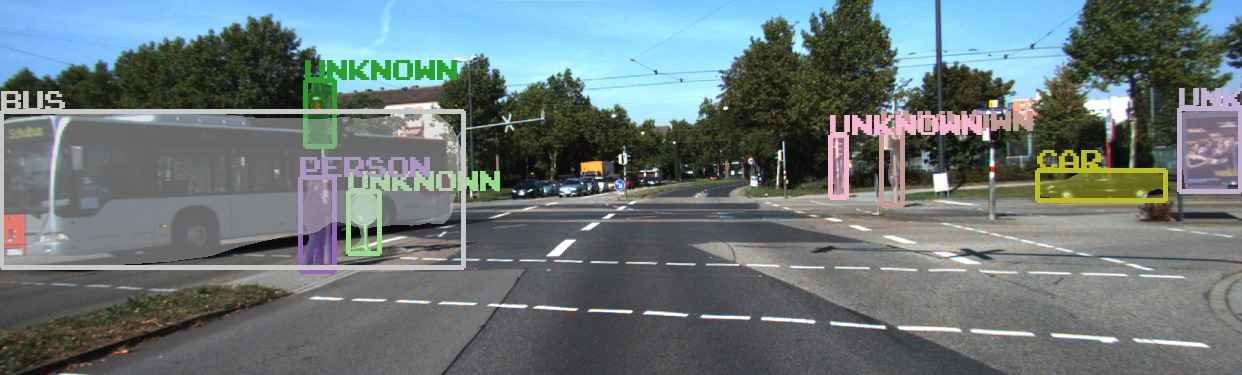}
\includegraphics[width=0.50\linewidth]{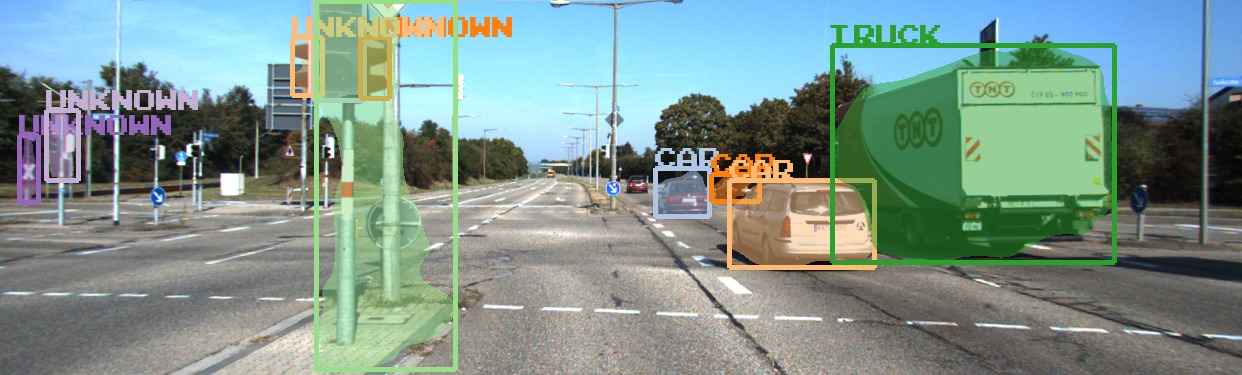}
\includegraphics[width=0.50\linewidth]{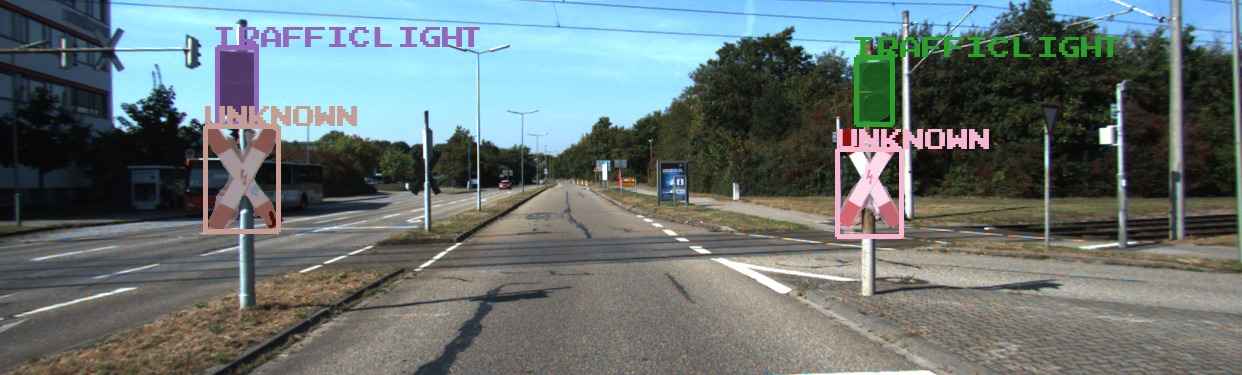}
\includegraphics[width=0.50\linewidth]{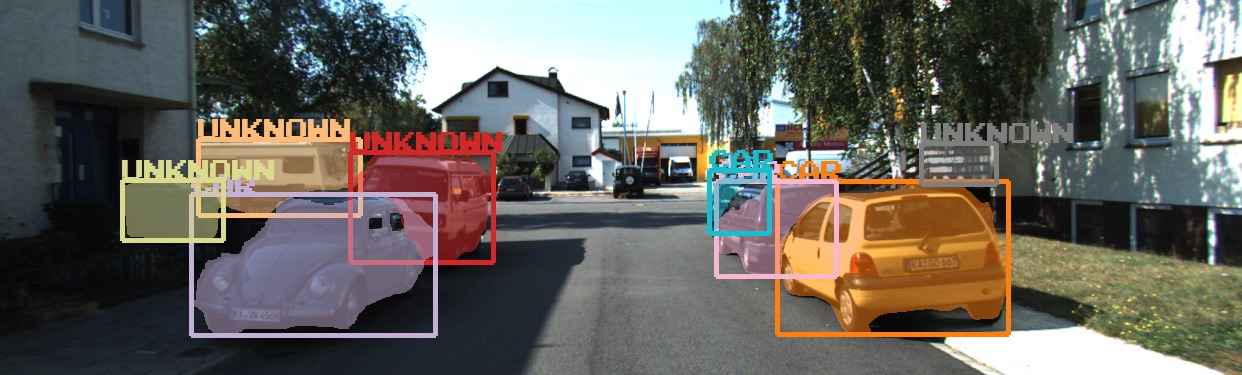}
\includegraphics[width=0.50\linewidth]{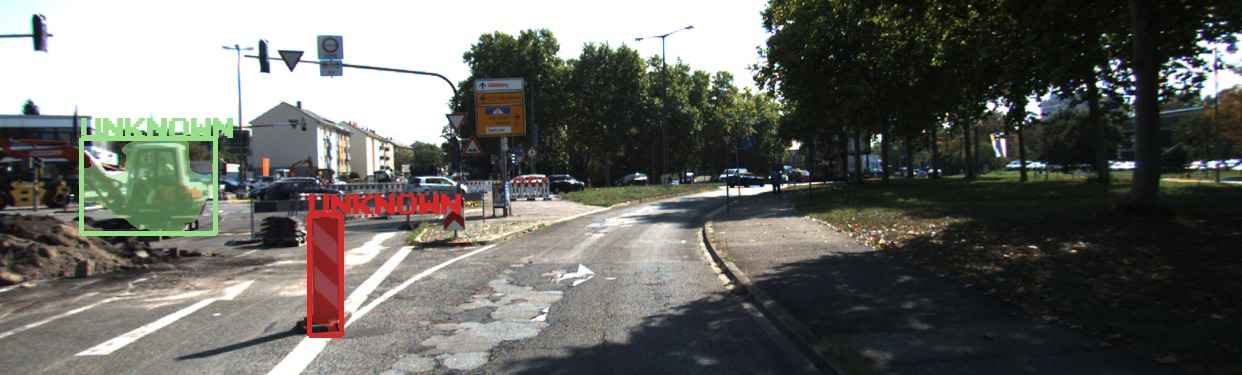}
\includegraphics[width=0.50\linewidth]{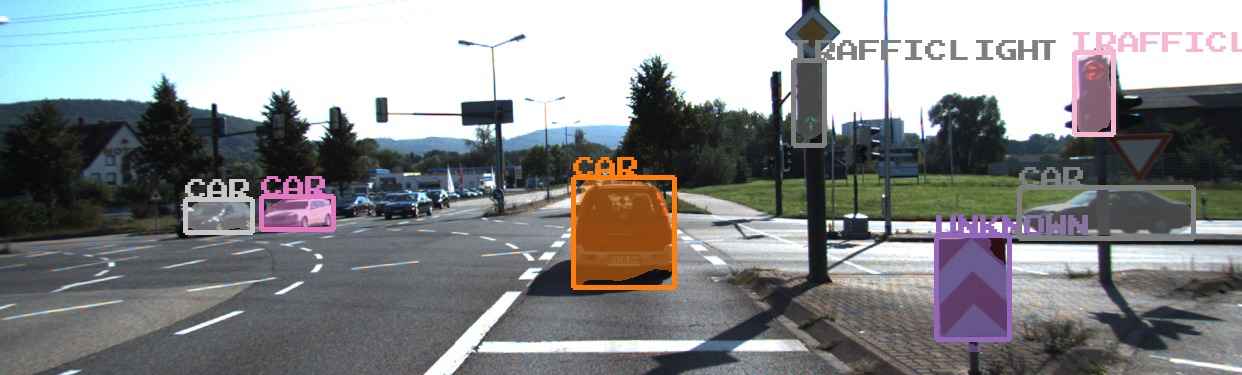}
\includegraphics[width=0.50\linewidth]{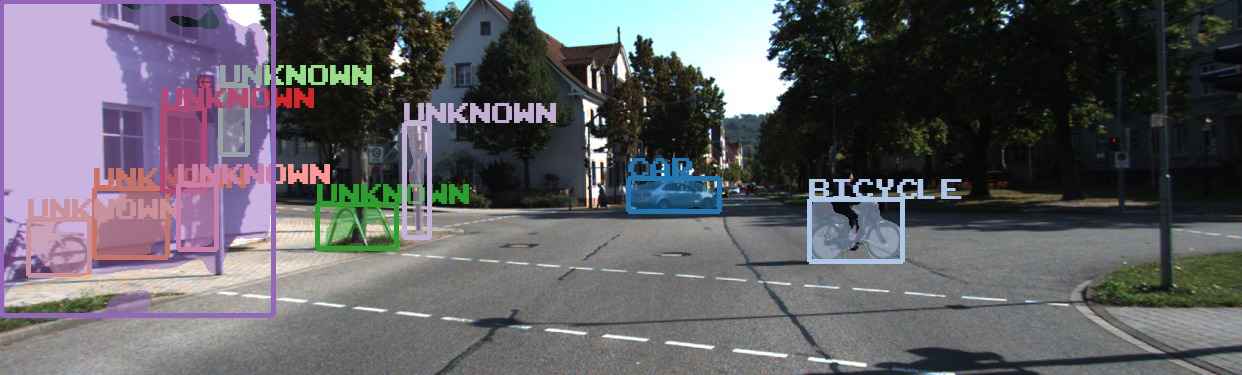}
\includegraphics[width=0.50\linewidth]{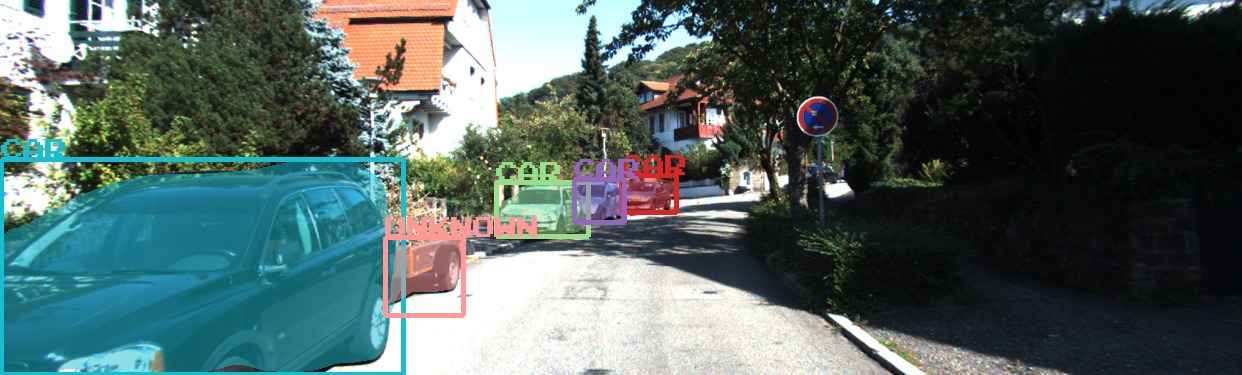}
\includegraphics[width=0.50\linewidth]{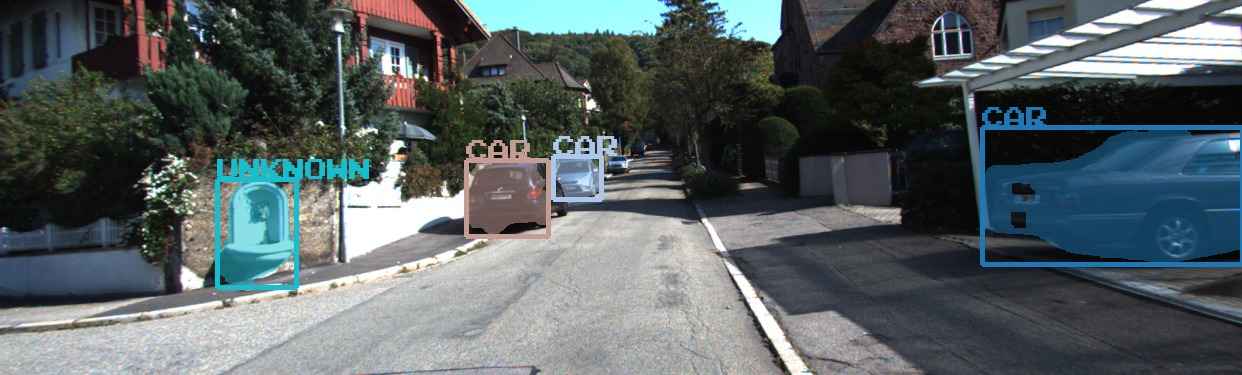}
\includegraphics[width=0.50\linewidth]{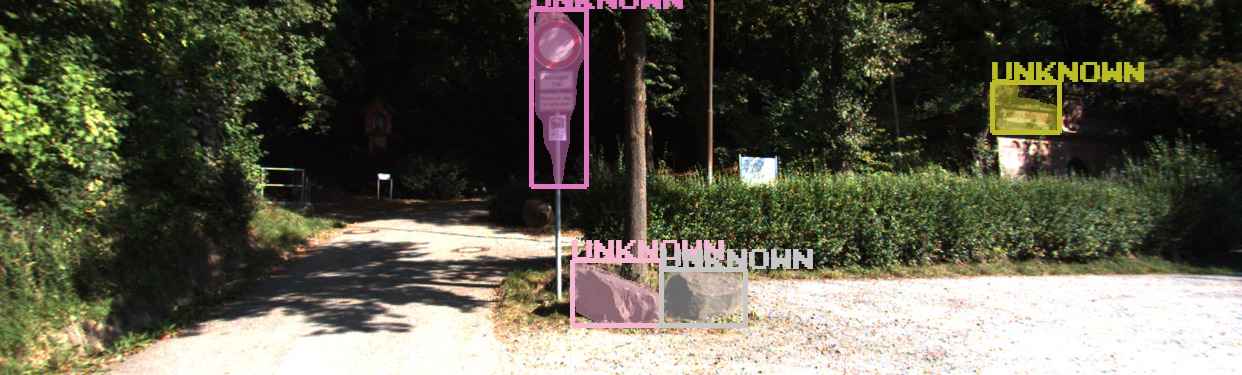}
\includegraphics[width=0.50\linewidth]{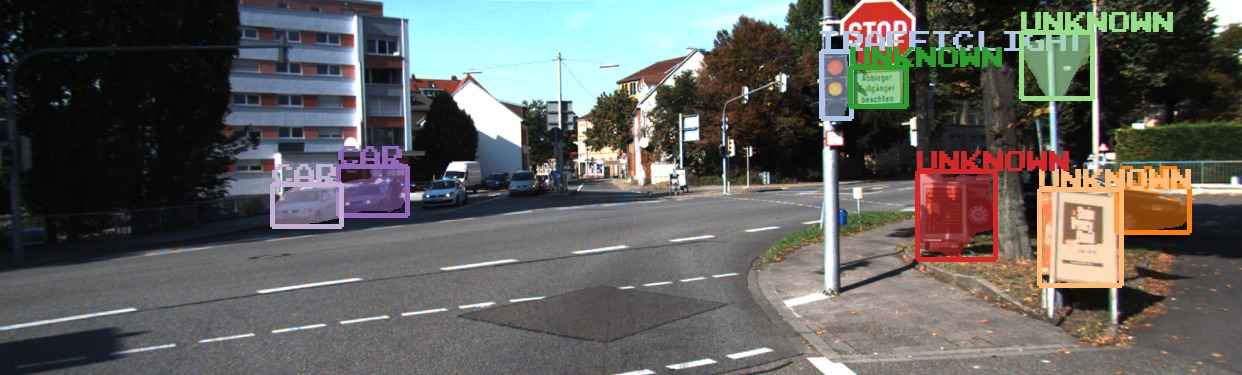}
\includegraphics[width=0.50\linewidth]{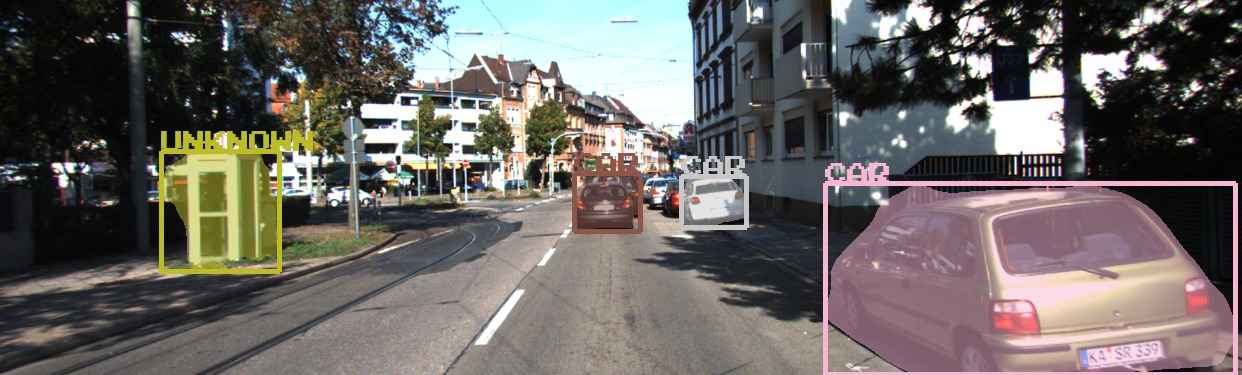}
\includegraphics[width=0.50\linewidth]{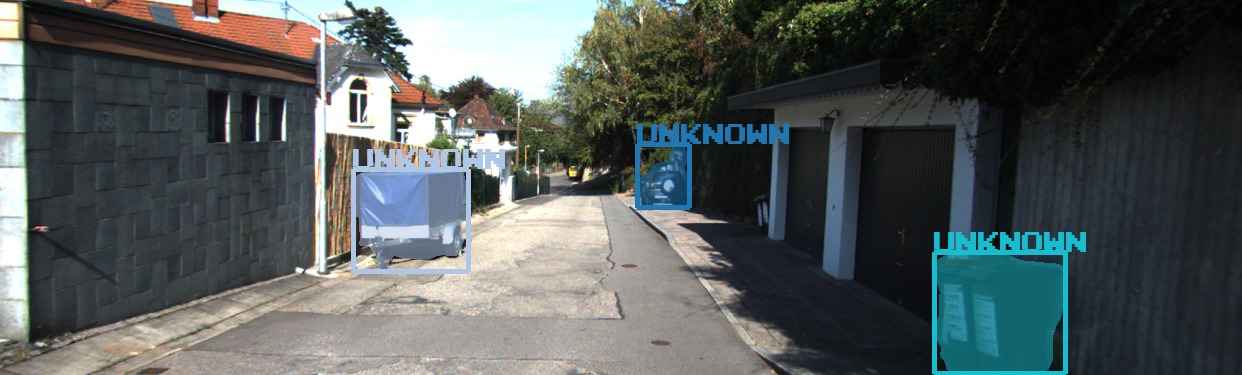}
\includegraphics[width=0.50\linewidth]{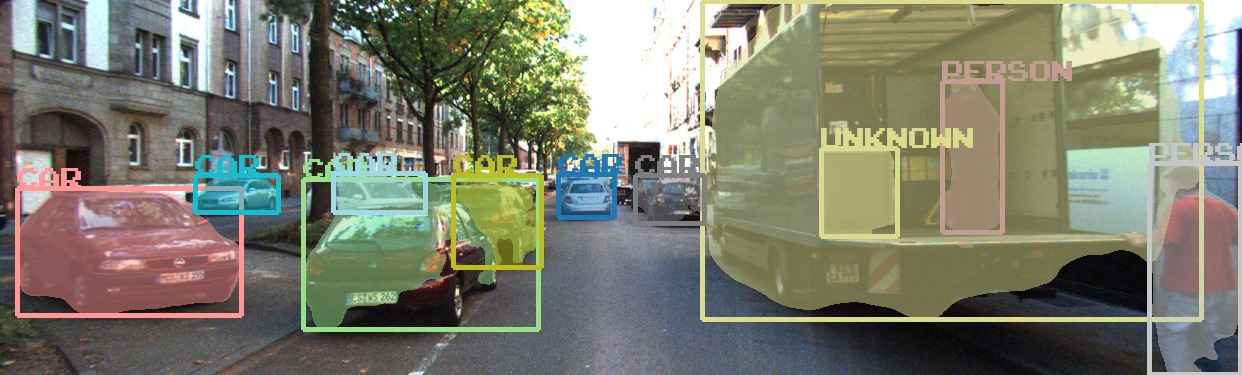}
\includegraphics[width=0.50\linewidth]{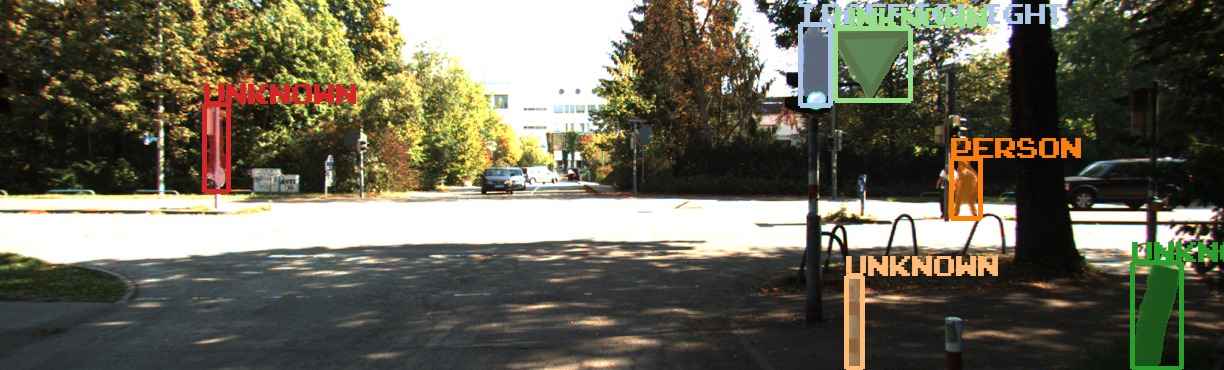}
\includegraphics[width=0.50\linewidth]{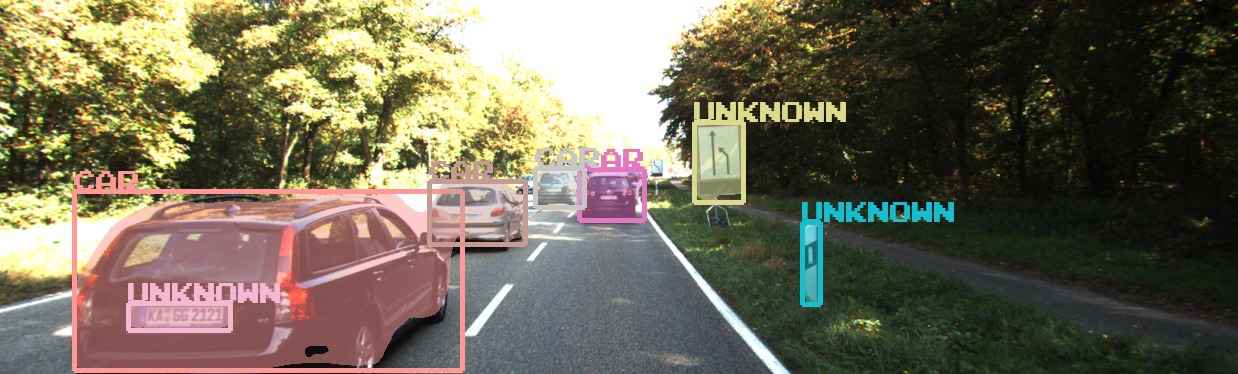}
\caption{Qualitative tracking results on the KITTI Raw \cite{Geiger12CVPR} dataset. Beside tracked objects, recognized by the classifier, we also find new objects such as various traffic traffic signs, car trailers, advertisements, poles, caterpillar machines, post boxes, \etc.}
\label{fig:qualitative_kitti}
\end{figure*}

\begin{figure*}[ht]
\includegraphics[width=0.33\linewidth]{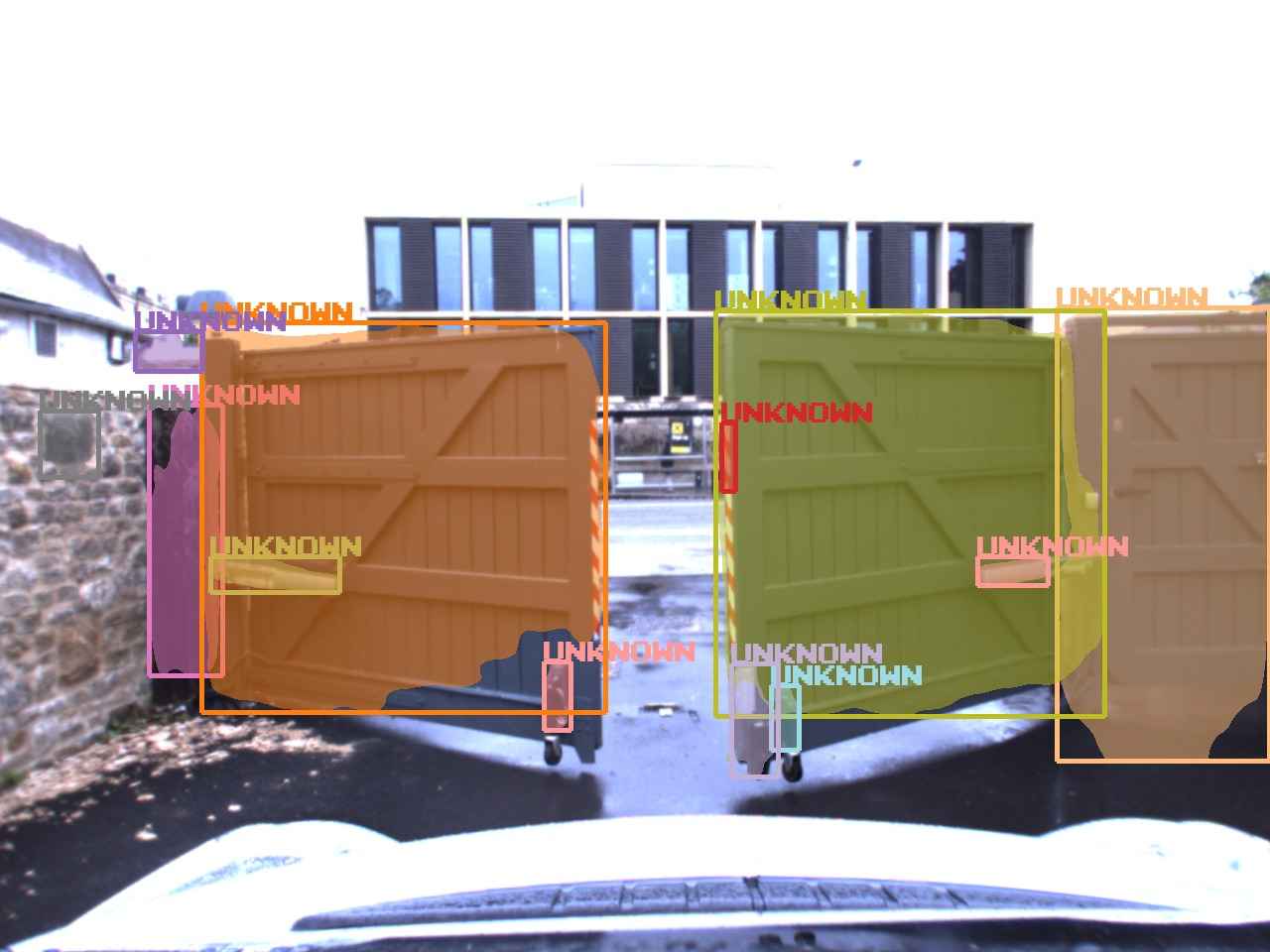}
\includegraphics[width=0.33\linewidth]{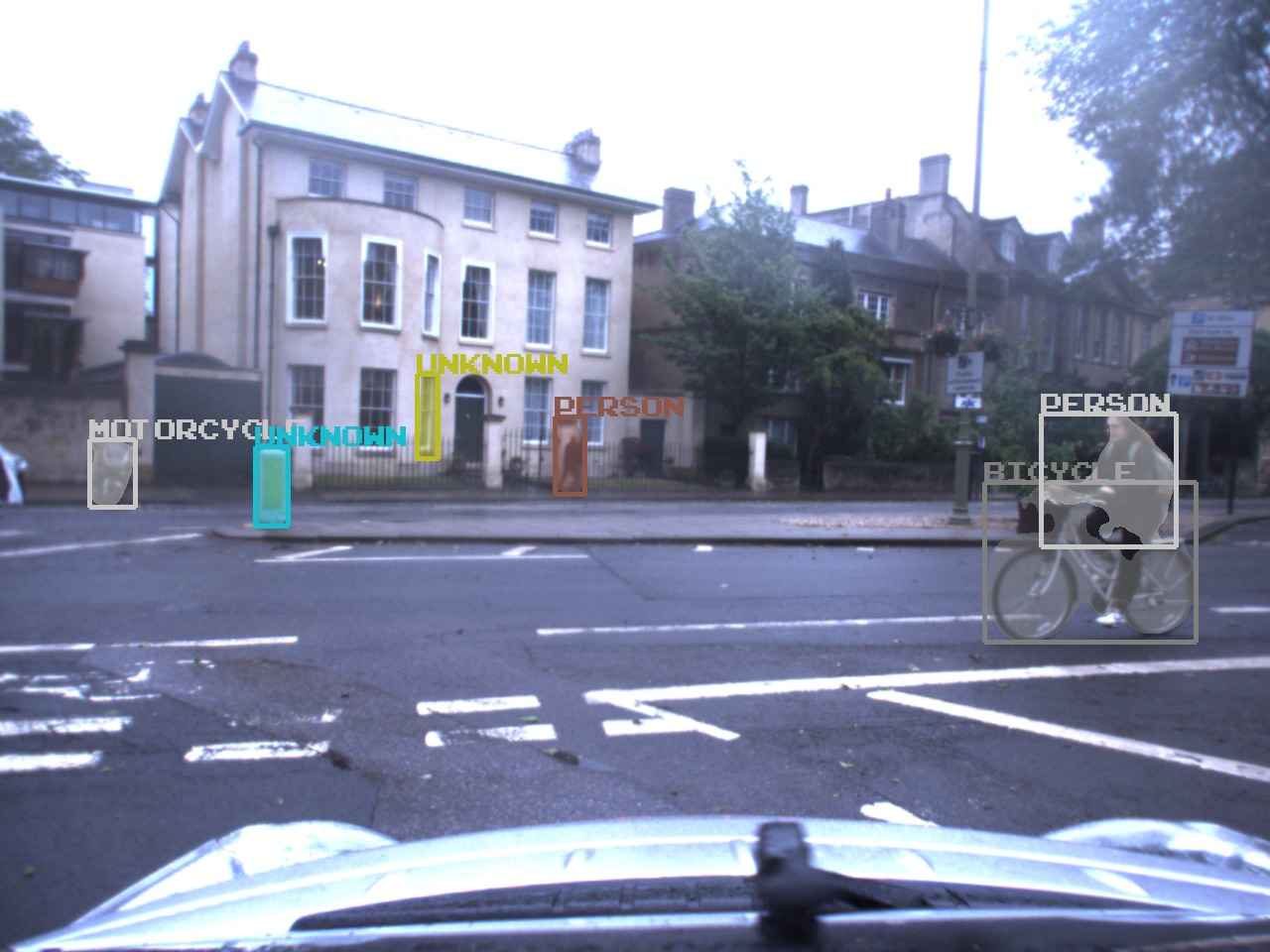}
\includegraphics[width=0.33\linewidth]{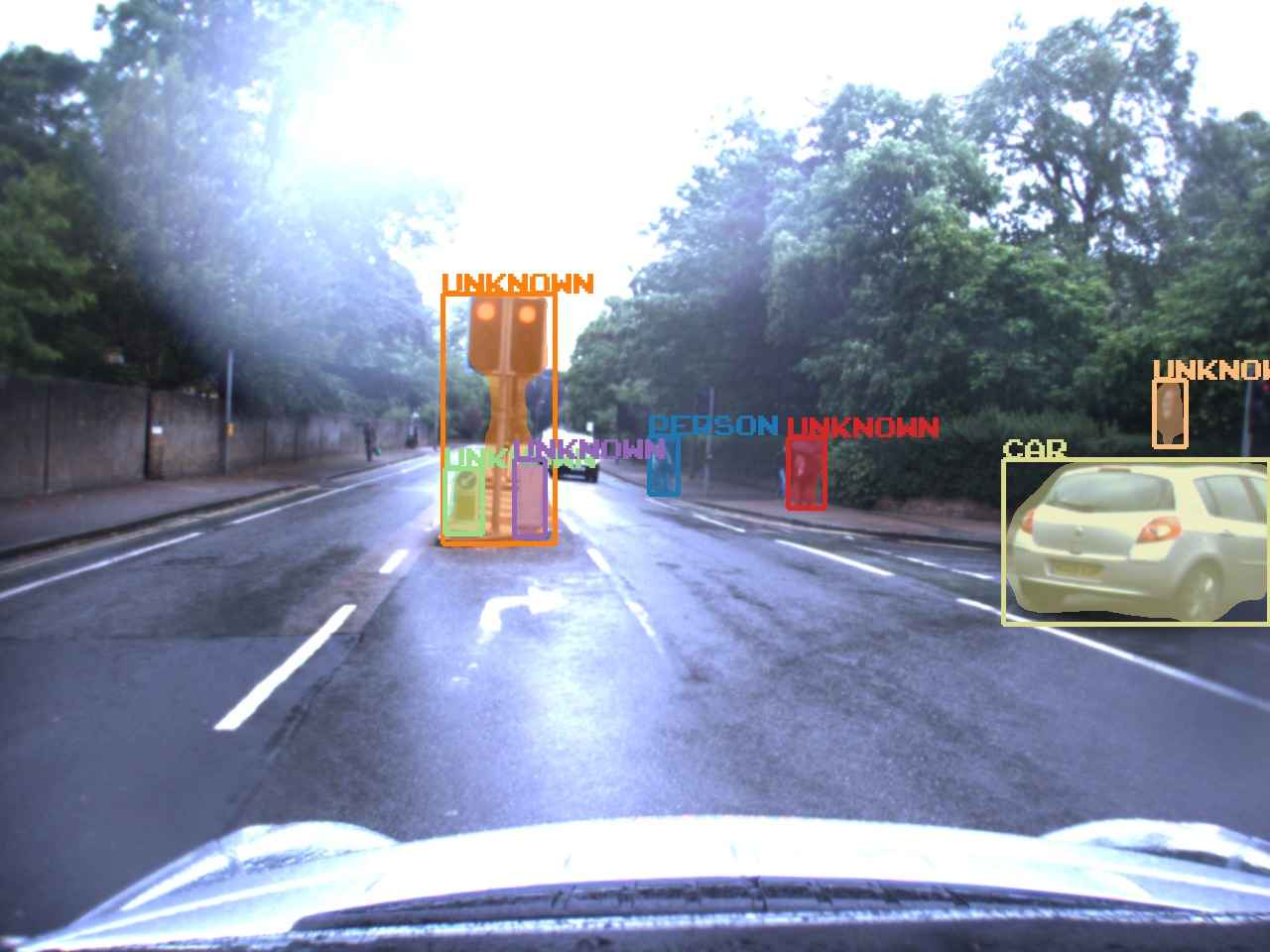}
\includegraphics[width=0.33\linewidth]{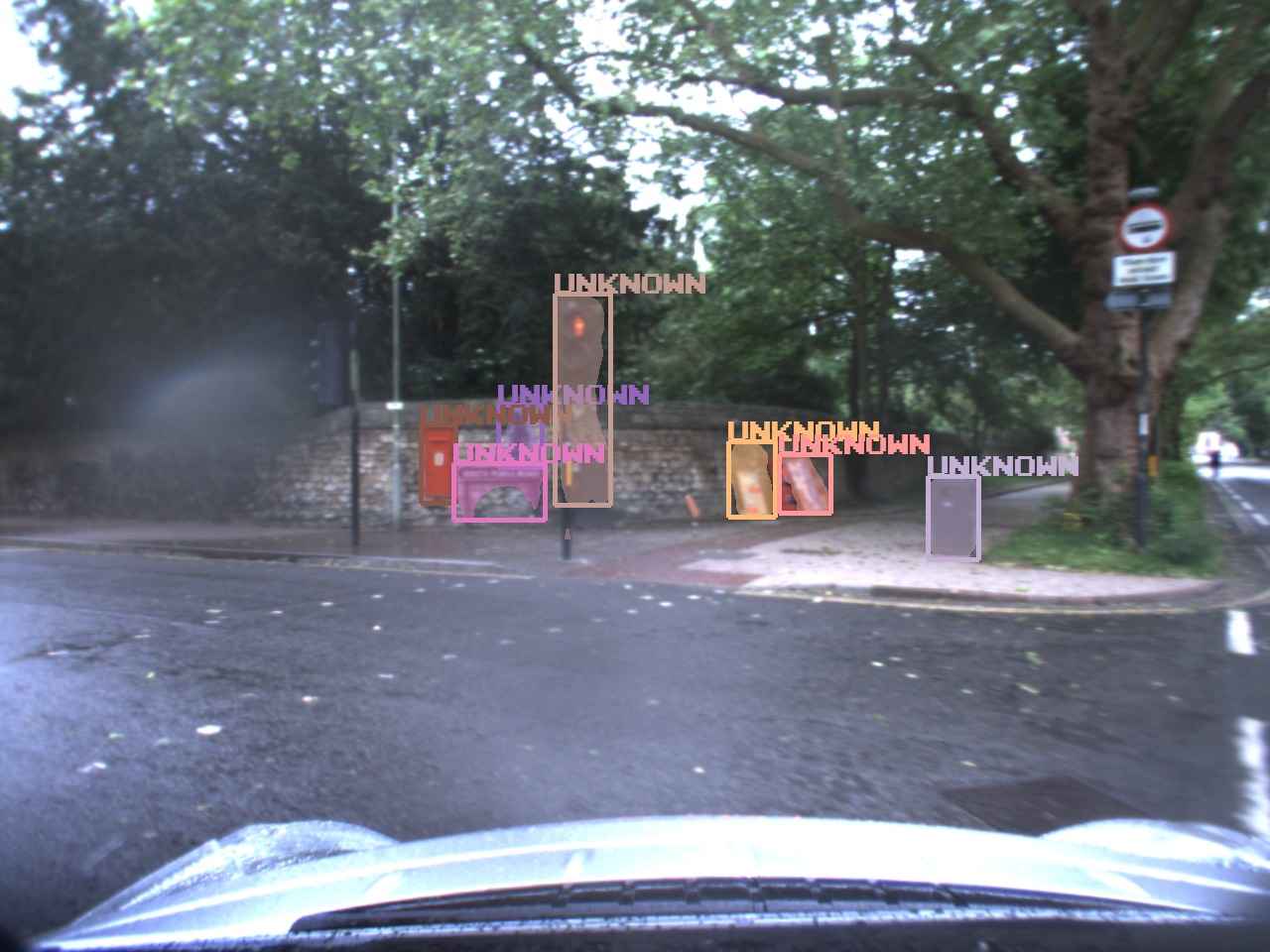}
\includegraphics[width=0.33\linewidth]{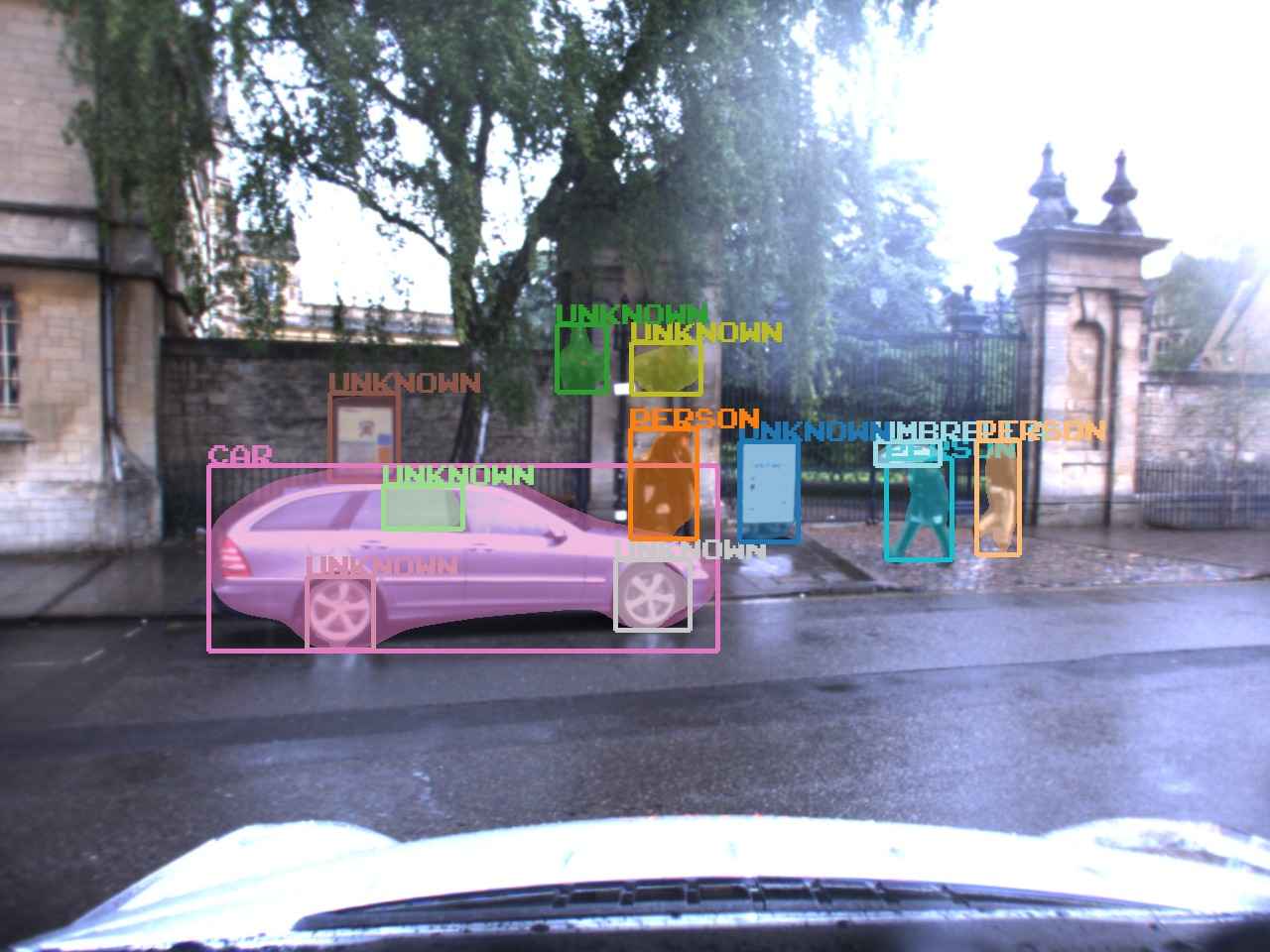}
\includegraphics[width=0.33\linewidth]{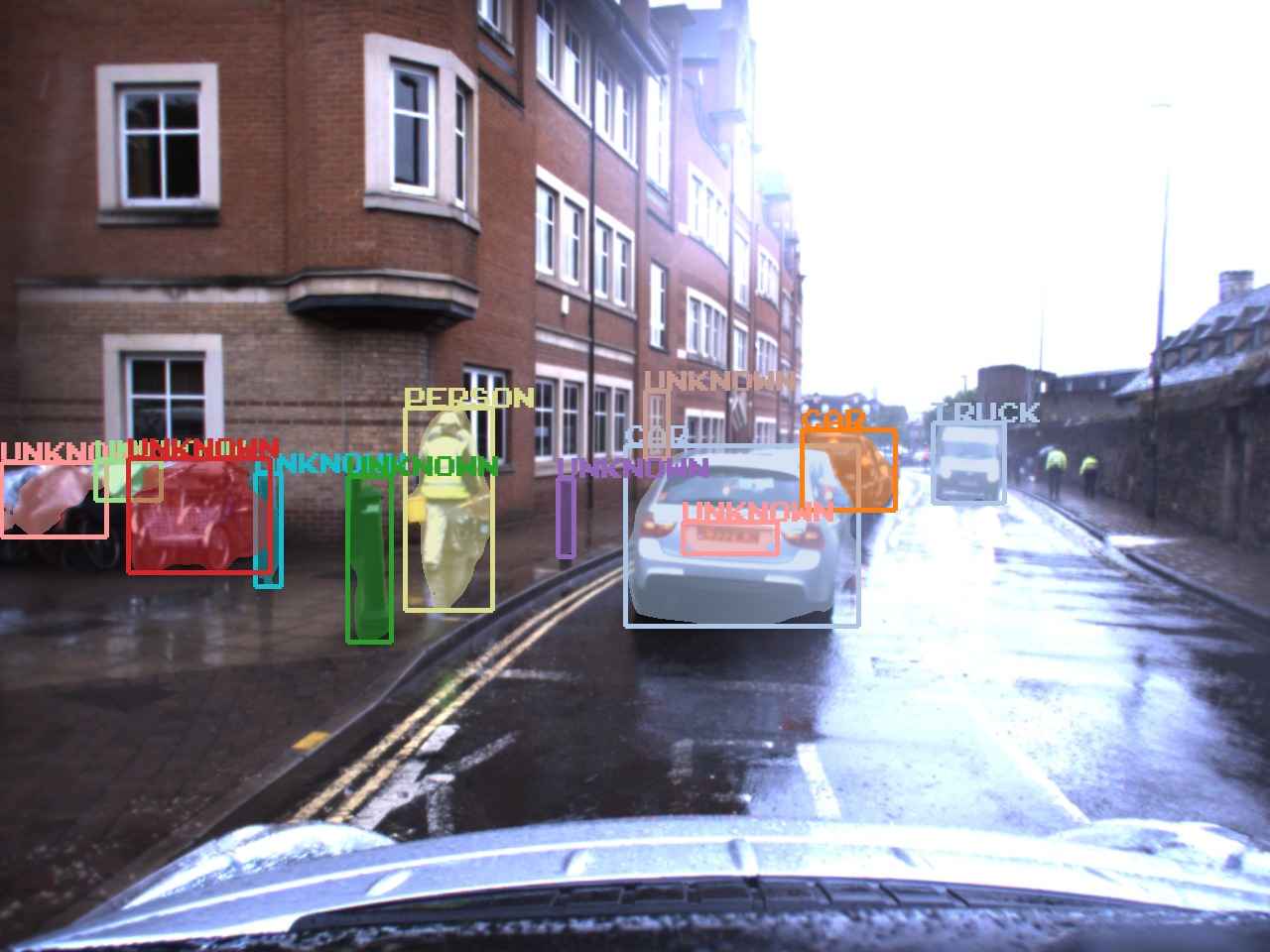}
\includegraphics[width=0.33\linewidth]{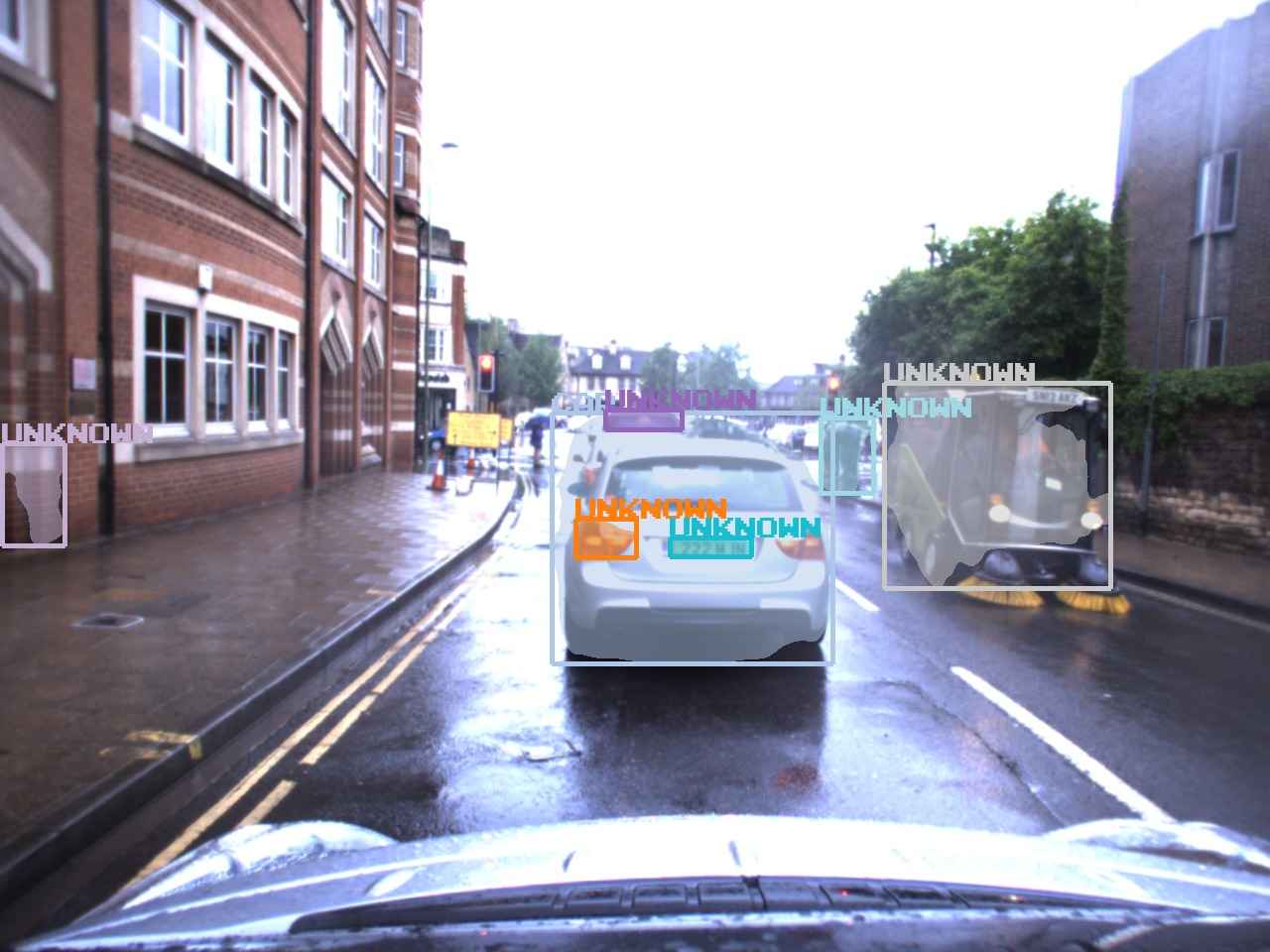}
\includegraphics[width=0.33\linewidth]{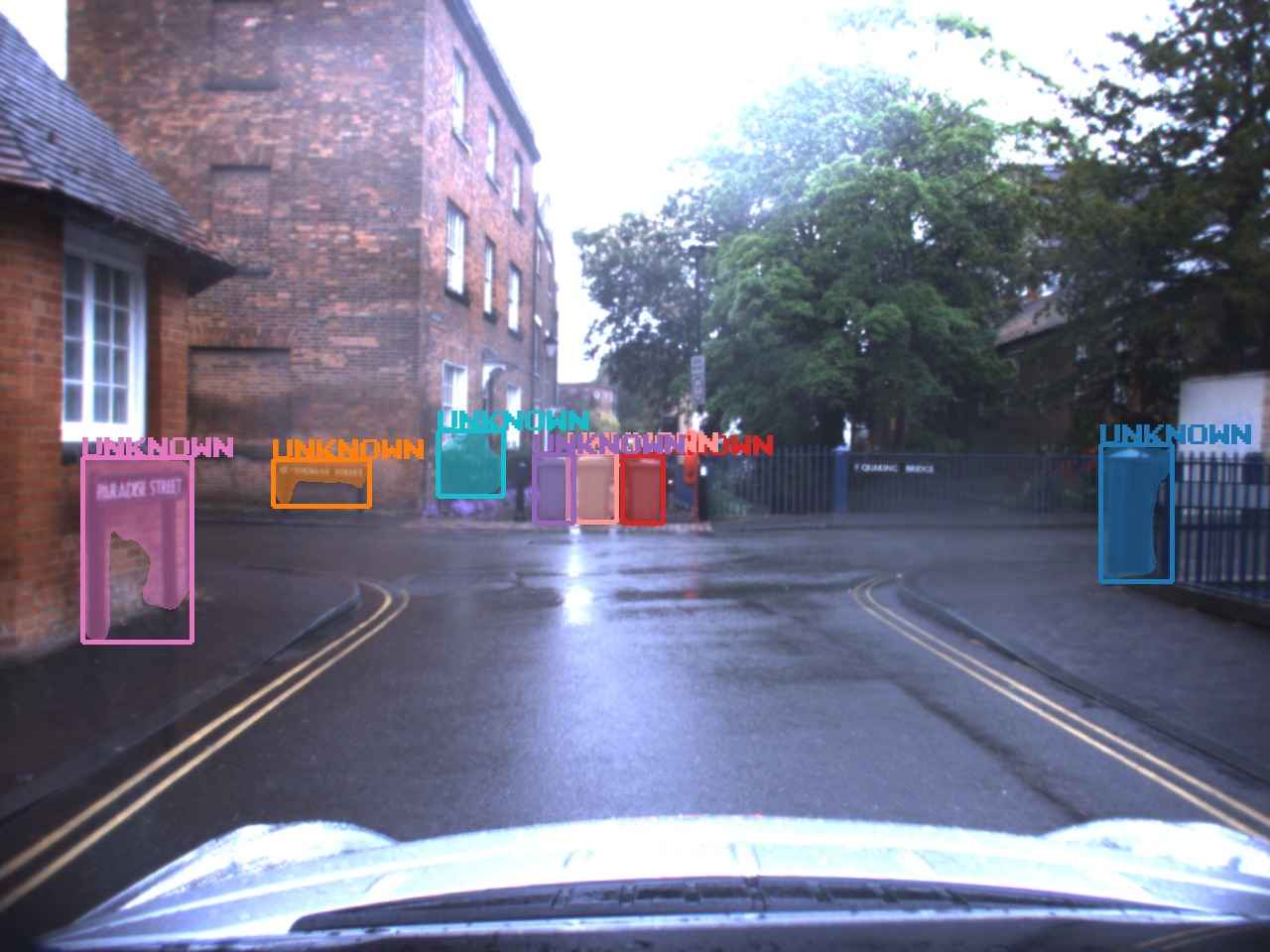}
\includegraphics[width=0.33\linewidth]{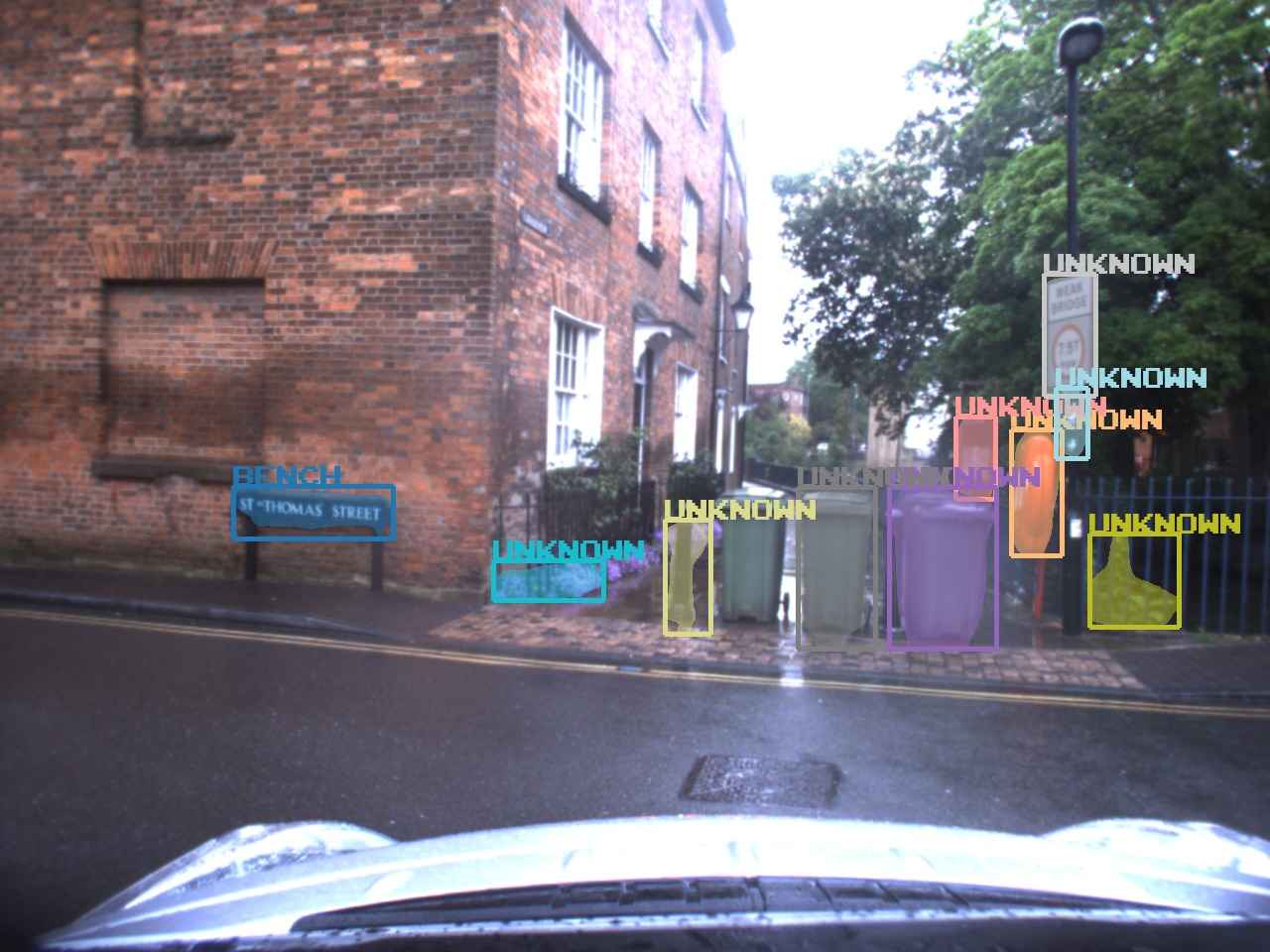}
\includegraphics[width=0.33\linewidth]{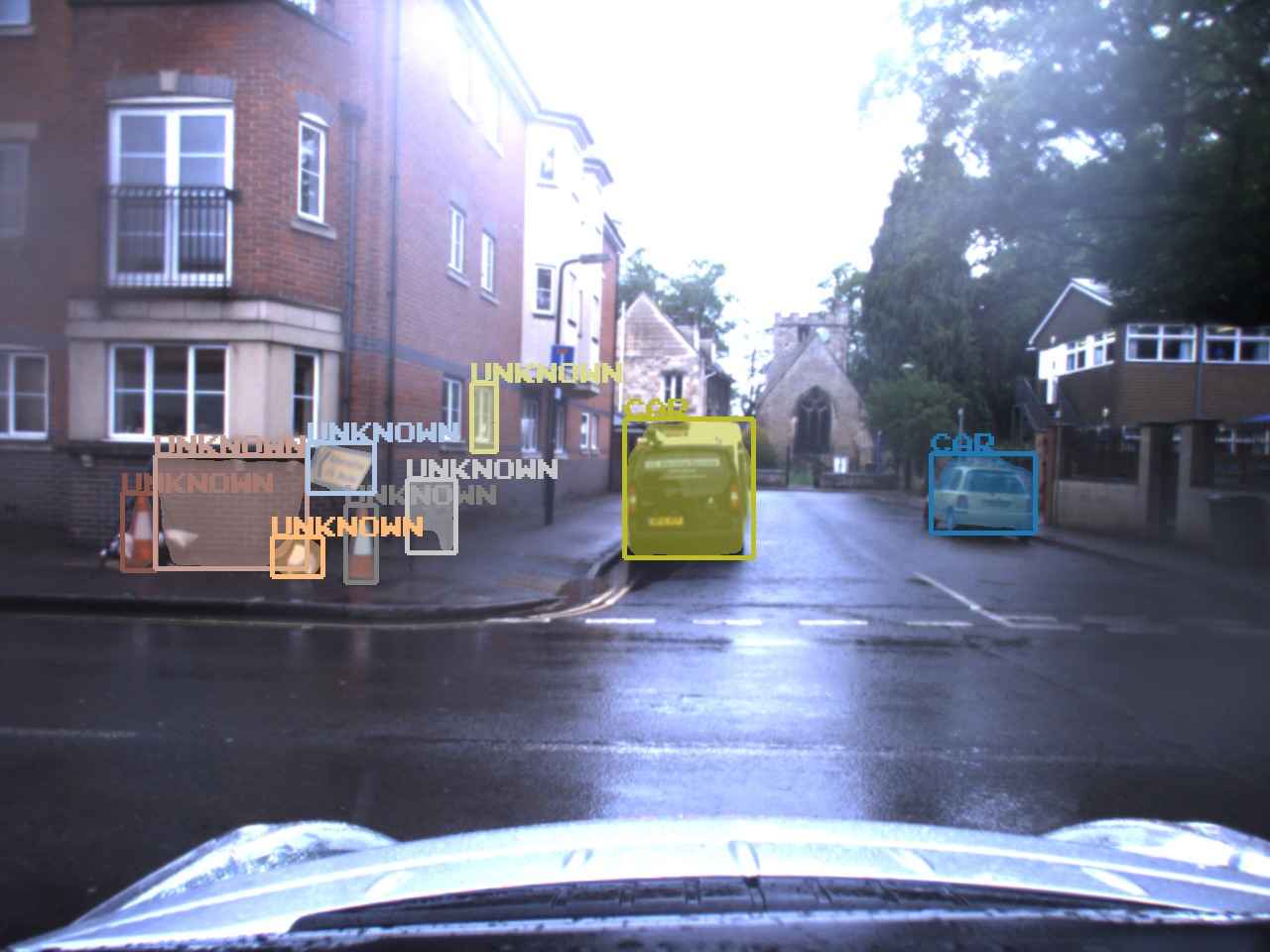}
\includegraphics[width=0.33\linewidth]{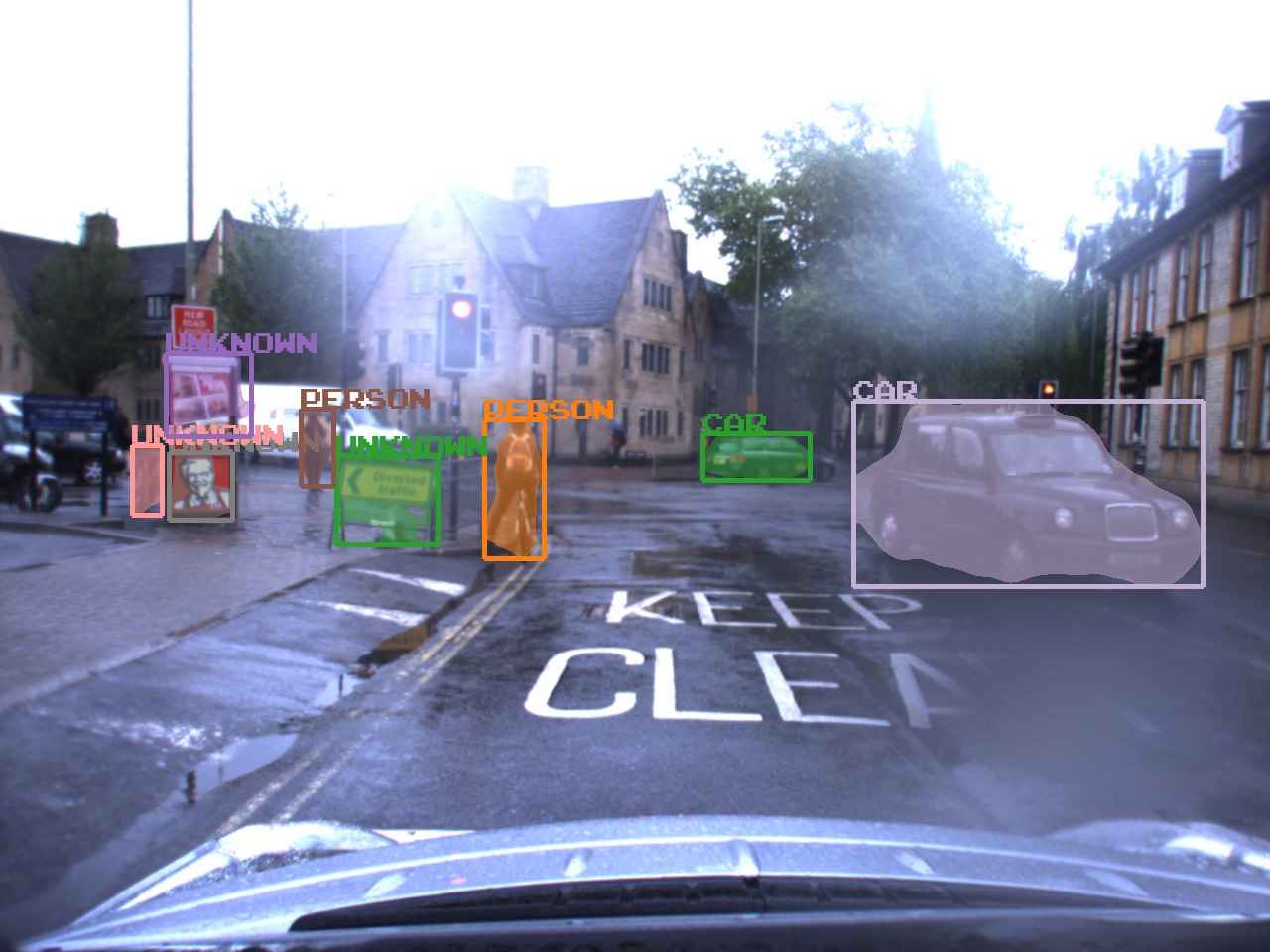}
\includegraphics[width=0.33\linewidth]{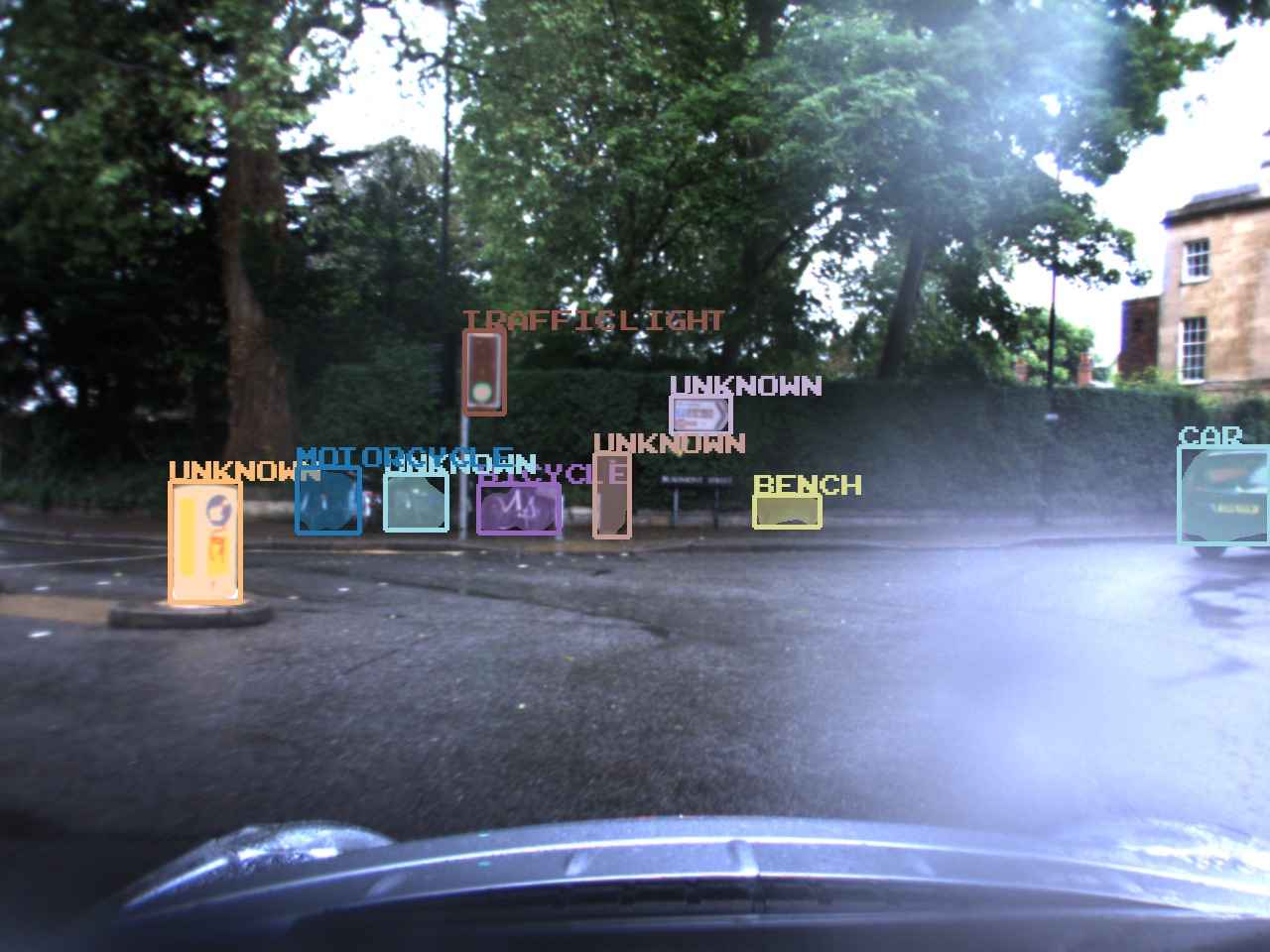}
\includegraphics[width=0.33\linewidth]{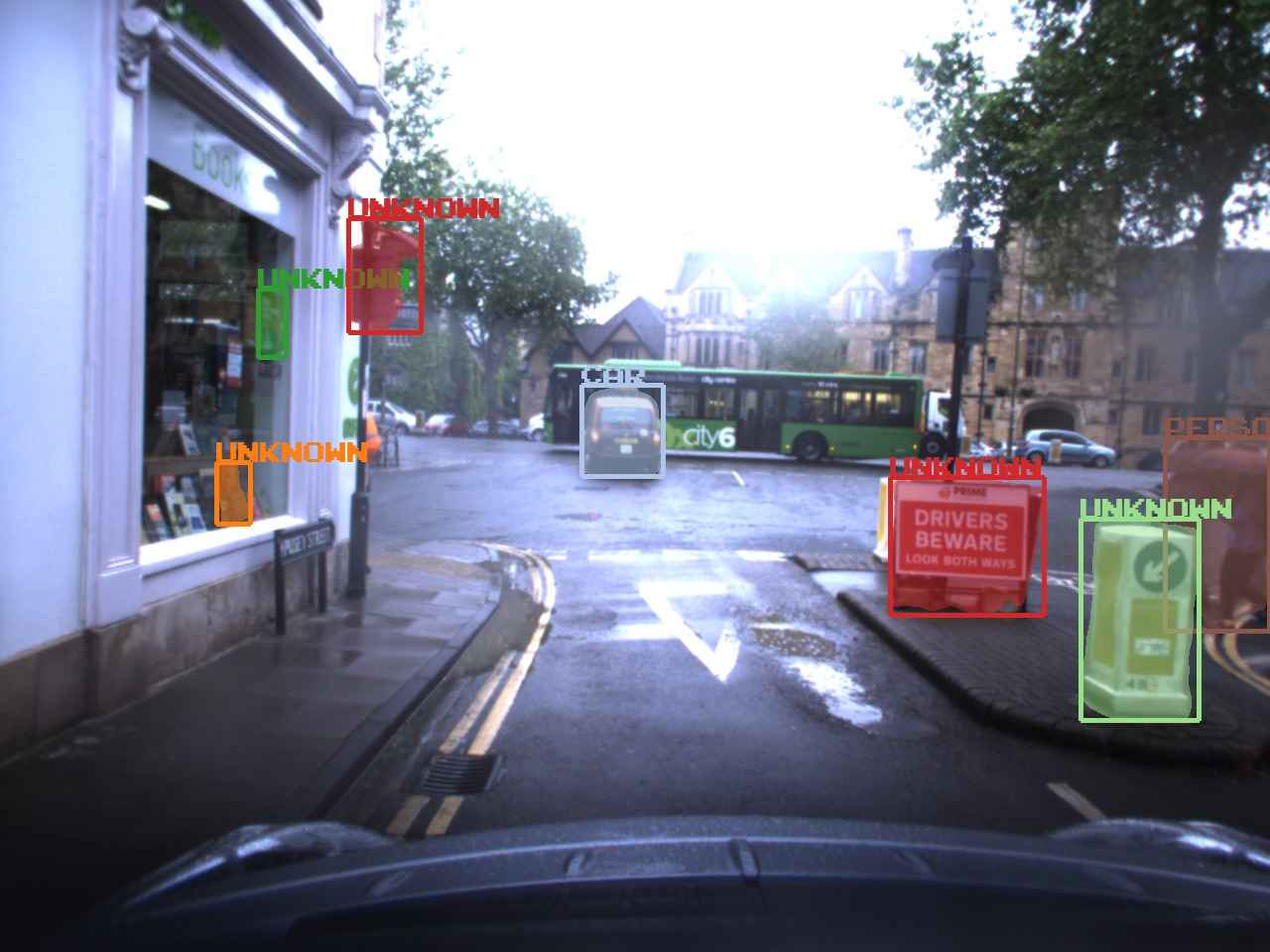}
\includegraphics[width=0.33\linewidth]{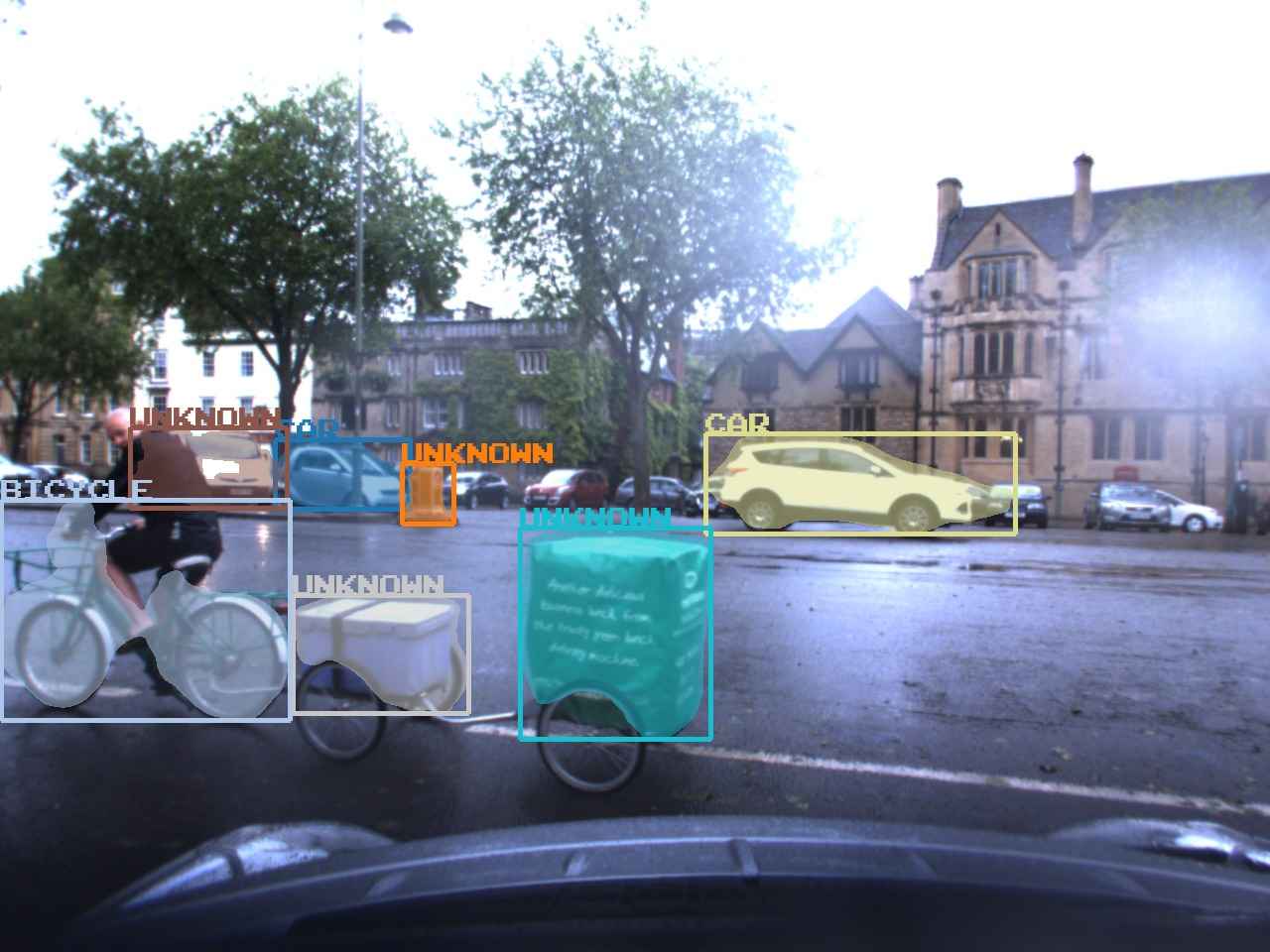}
\includegraphics[width=0.33\linewidth]{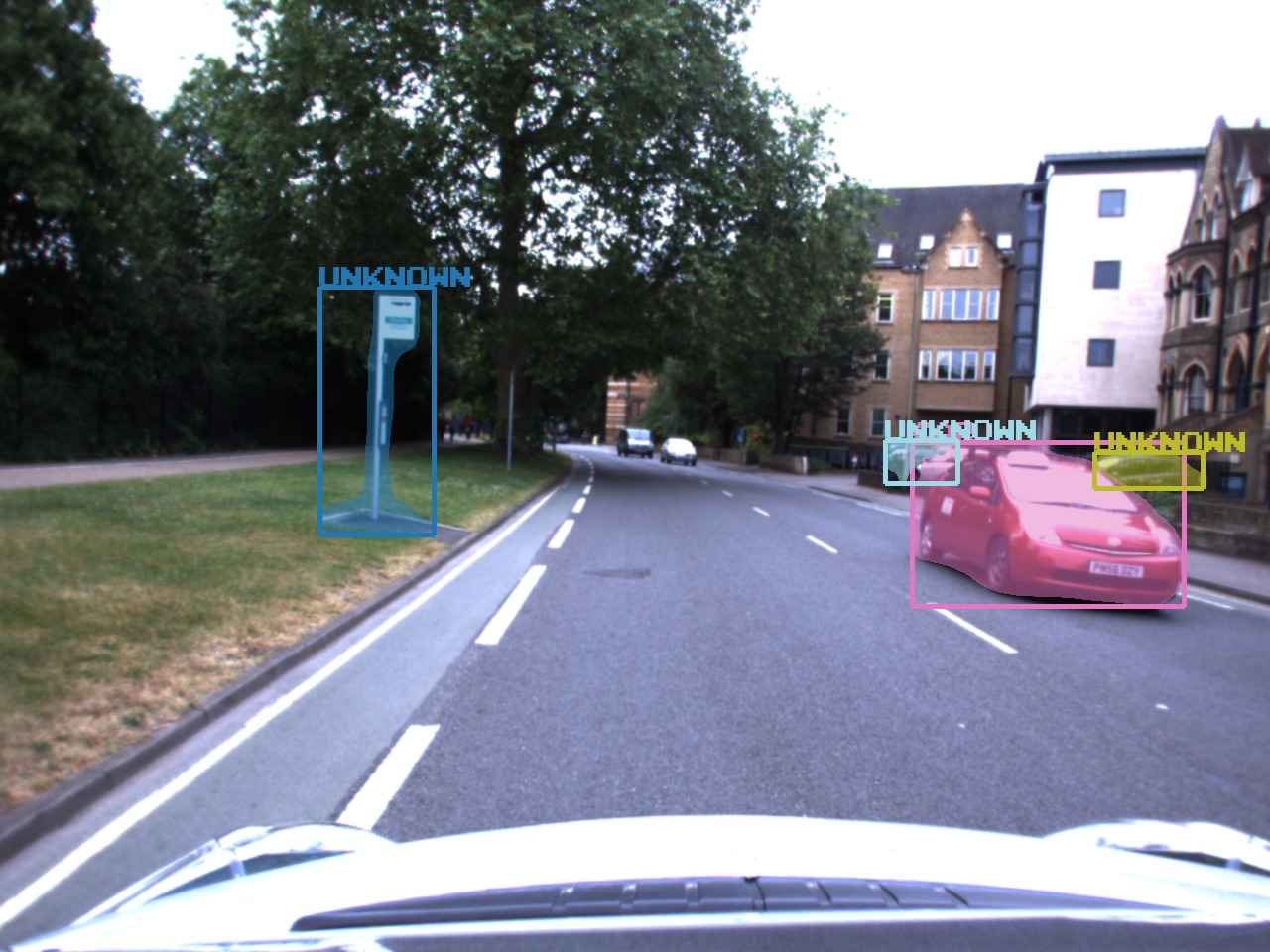}
\caption{Qualitative tracking results on the Oxford Robotcar \cite{Maddern17IJRR} dataset. Beside tracked objects, recognized by the classifier, we also find new objects such as various traffic signs, traffic cones, advertisements, poles, post boxes, street cleaners,  \etc.}
\label{fig:qualitative_oxford}
\end{figure*}

\begin{figure*}[ht]
\includegraphics[width=0.33\linewidth]{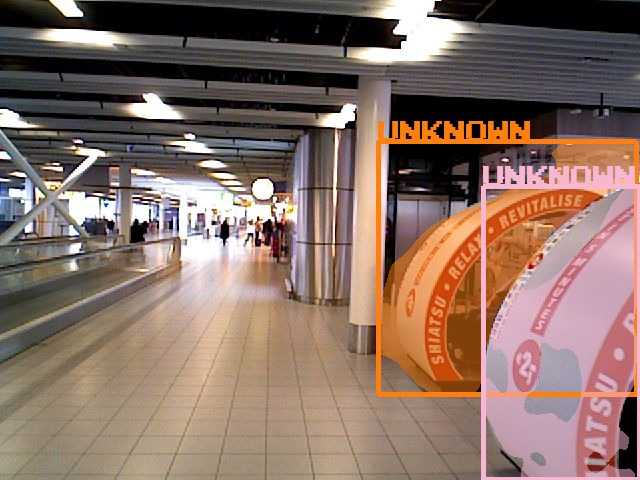}
\includegraphics[width=0.33\linewidth]{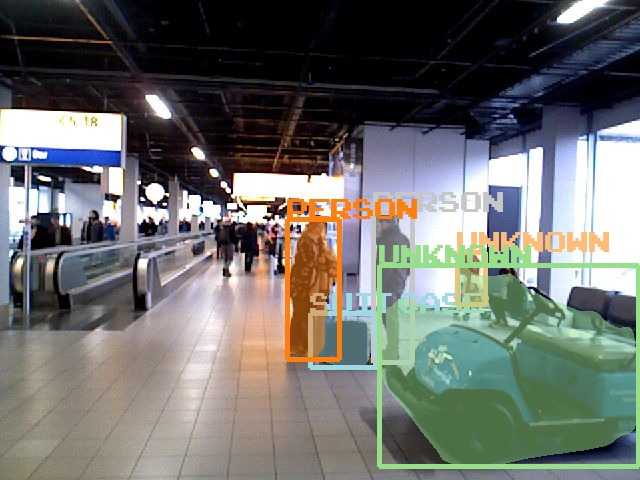}
\includegraphics[width=0.33\linewidth]{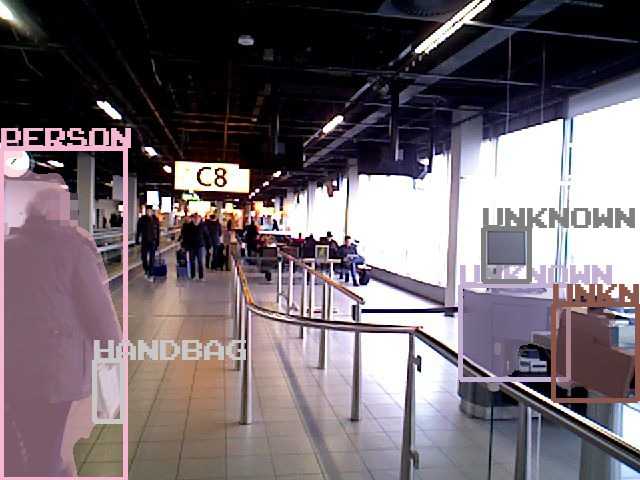}
\includegraphics[width=0.33\linewidth]{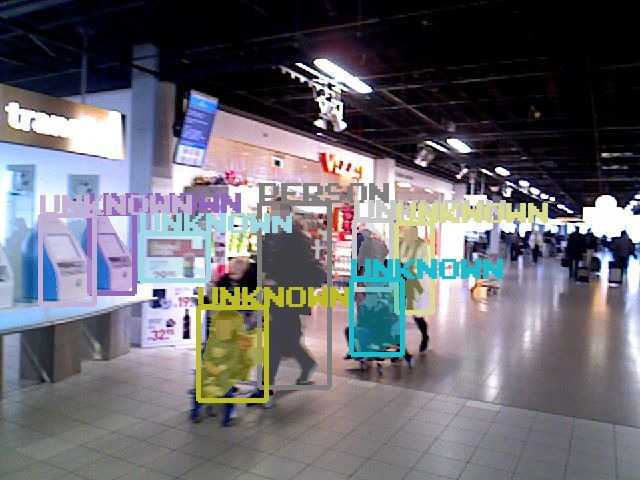}
\includegraphics[width=0.33\linewidth]{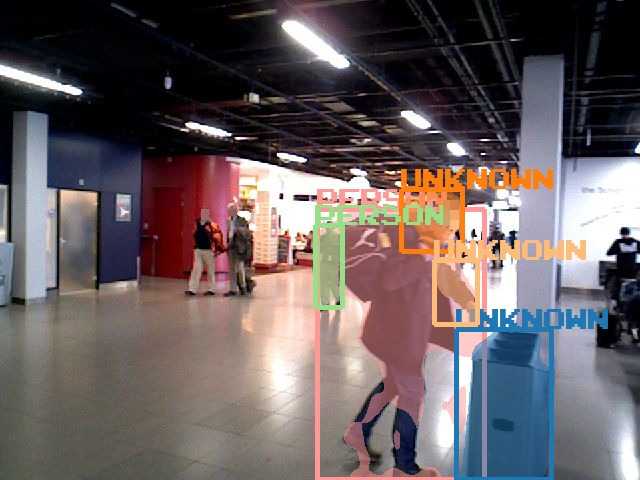}
\includegraphics[width=0.33\linewidth]{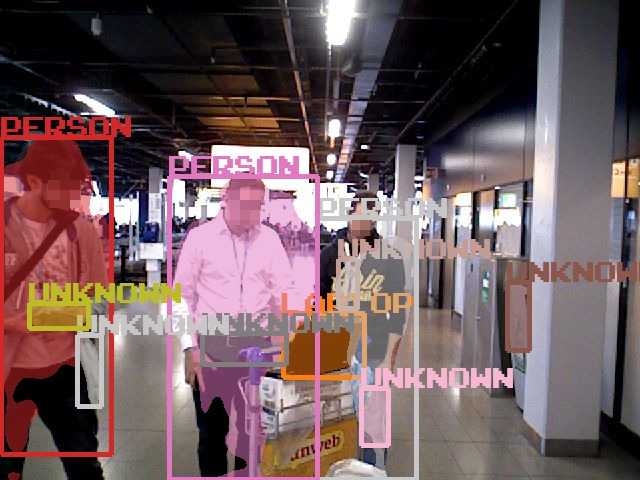}
\includegraphics[width=0.33\linewidth]{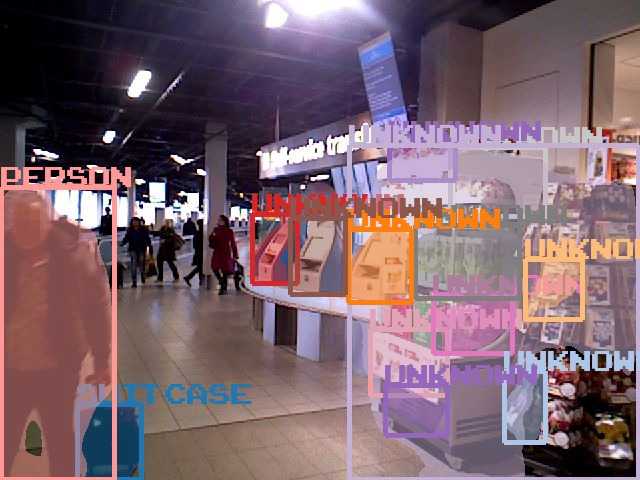}
\includegraphics[width=0.33\linewidth]{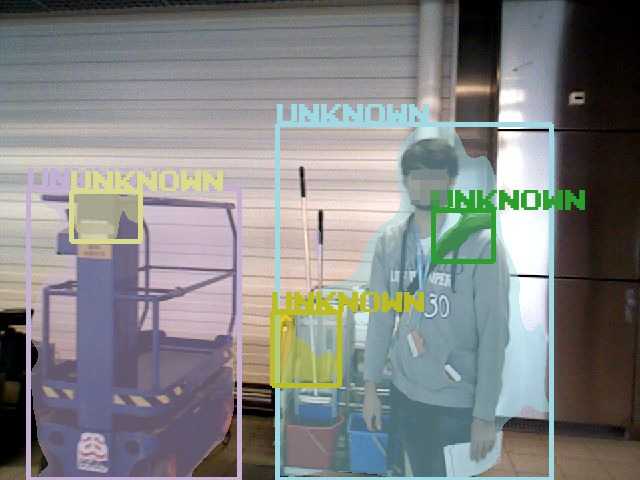}
\includegraphics[width=0.33\linewidth]{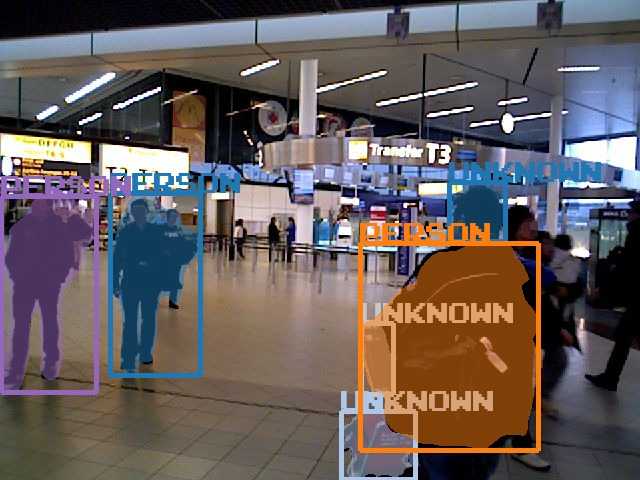}
\includegraphics[width=0.33\linewidth]{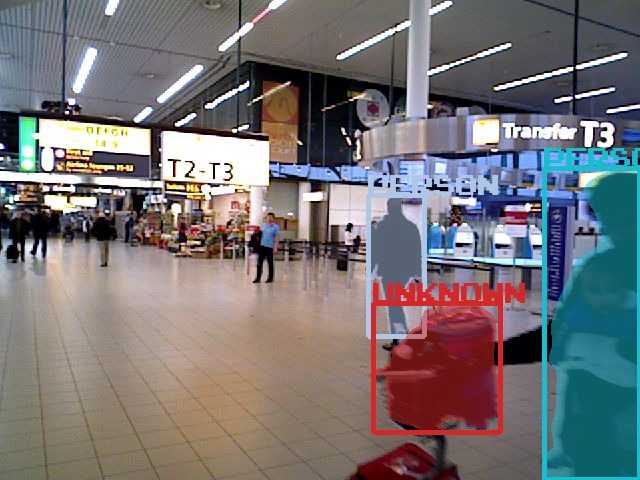}
\includegraphics[width=0.33\linewidth]{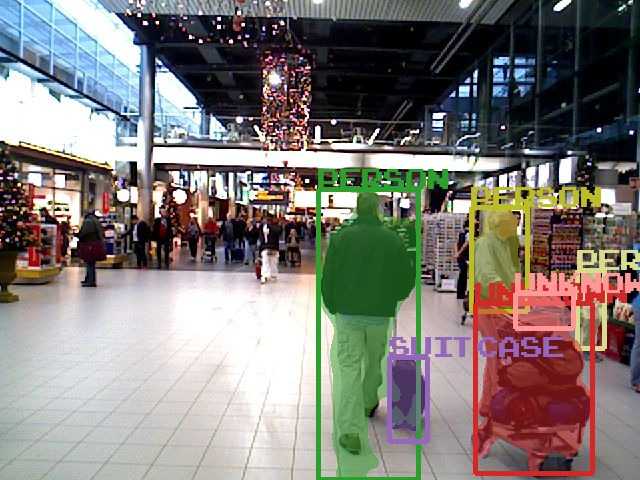}
\includegraphics[width=0.33\linewidth]{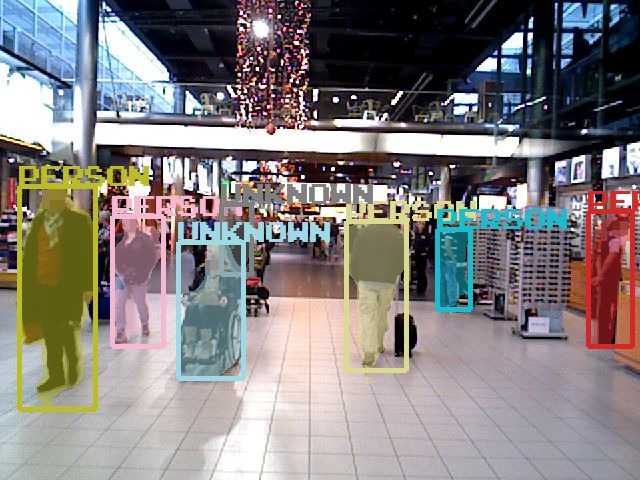}
\includegraphics[width=0.33\linewidth]{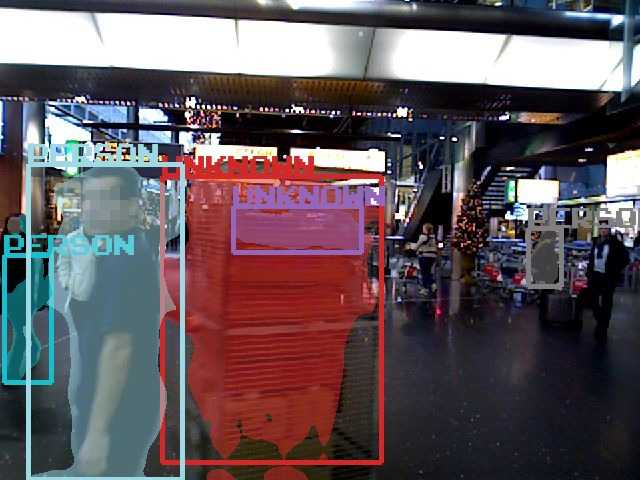}
\includegraphics[width=0.33\linewidth]{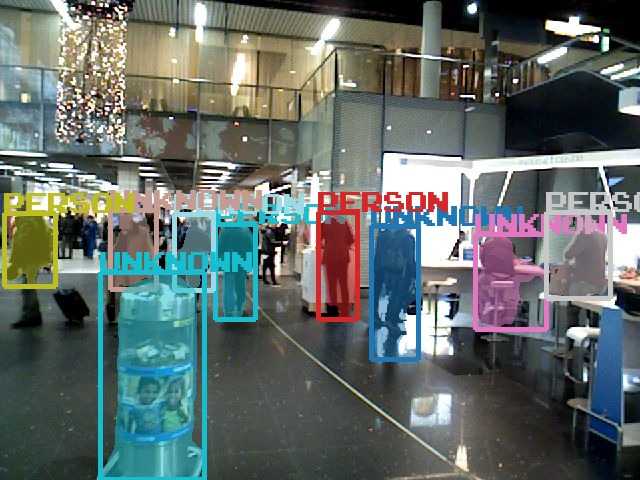}
\includegraphics[width=0.33\linewidth]{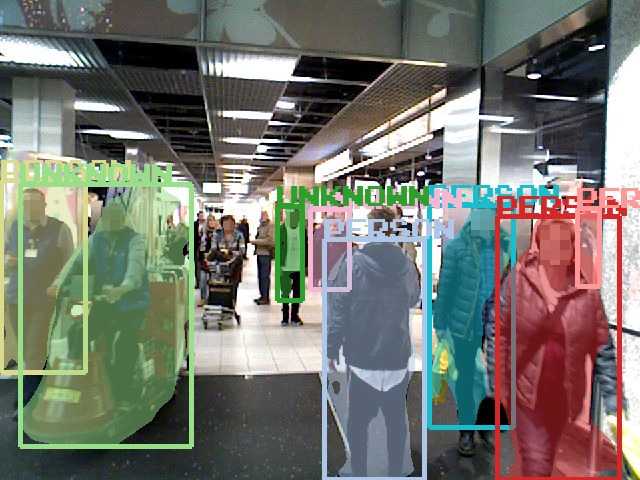}
\caption{Qualitative tracking results on the Schiphol Airport dataset. Beside tracked objects, recognized by the classifier, we also find new objects such as Christmas trees, relaxation booths, screens, garbage bins, wheelchairs, self check-in terminals, various airport mobility vehicles, luggage trolleys, \etc. Some faces have been pixelized to preserve privacy.}
\label{fig:qualitative_schiphol}
\end{figure*}

Figures \ref{fig:qualitative_kitti}, \ref{fig:qualitative_oxford}, and \ref{fig:qualitative_schiphol} show both, tracked ``known`` and ``unknown objects`` (we follow the simple definition - ``known`` objects are from categories labeled in the COCO dataset and all other objects are ``unknown``). As evident, most objects are recognized after the classification step. Among these are common traffic participants, such as \categoryname{bus}, \categoryname{car}, \categoryname{person}, \categoryname{truck}, \categoryname{traffic light}, or \categoryname{bicycle}. In the Schiphol Airport dataset, we additionally observe categories such as \categoryname{handbag}, \categoryname{laptop}, or \categoryname{suitcase} as recognized objects.

Among ``unknown`` objects, we find various traffic signs, car trailers, traffic cones, advertisements, poles, caterpillar machines, post boxes, street cleaners, \etc. In the Schiphol Airport dataset, we also note Christmas trees, relaxation booths, screens, garbage bins, wheelchairs, self check-in terminals, various airport mobility vehicles, luggage trolleys, \etc.

\begin{figure*}
\begin{center}
  \includegraphics[width=1.0\linewidth]{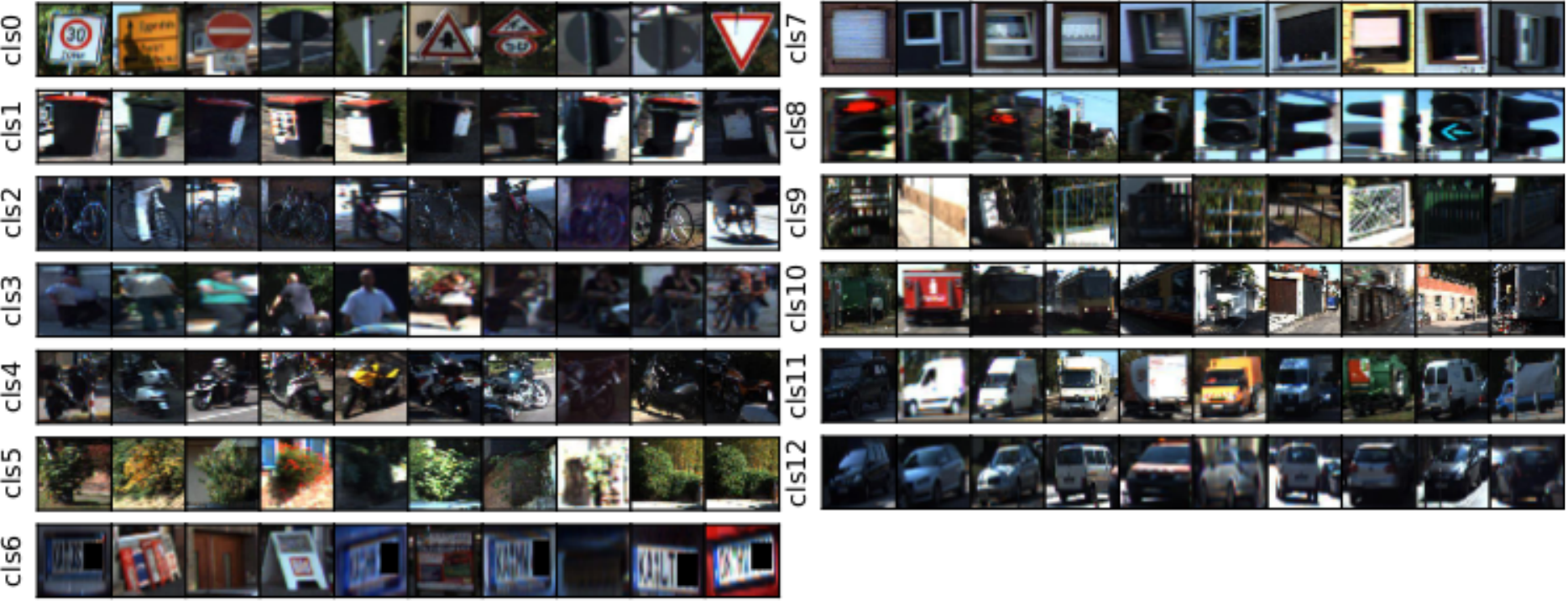}
  \caption{Visualization of the discovered categories on KITTI Raw. Each row shows example crops for one cluster.}
  \label{fig:clustering-KITTI}
  \end{center}
\end{figure*}

\begin{figure*}
\begin{center}
  \includegraphics[width=1.0\linewidth]{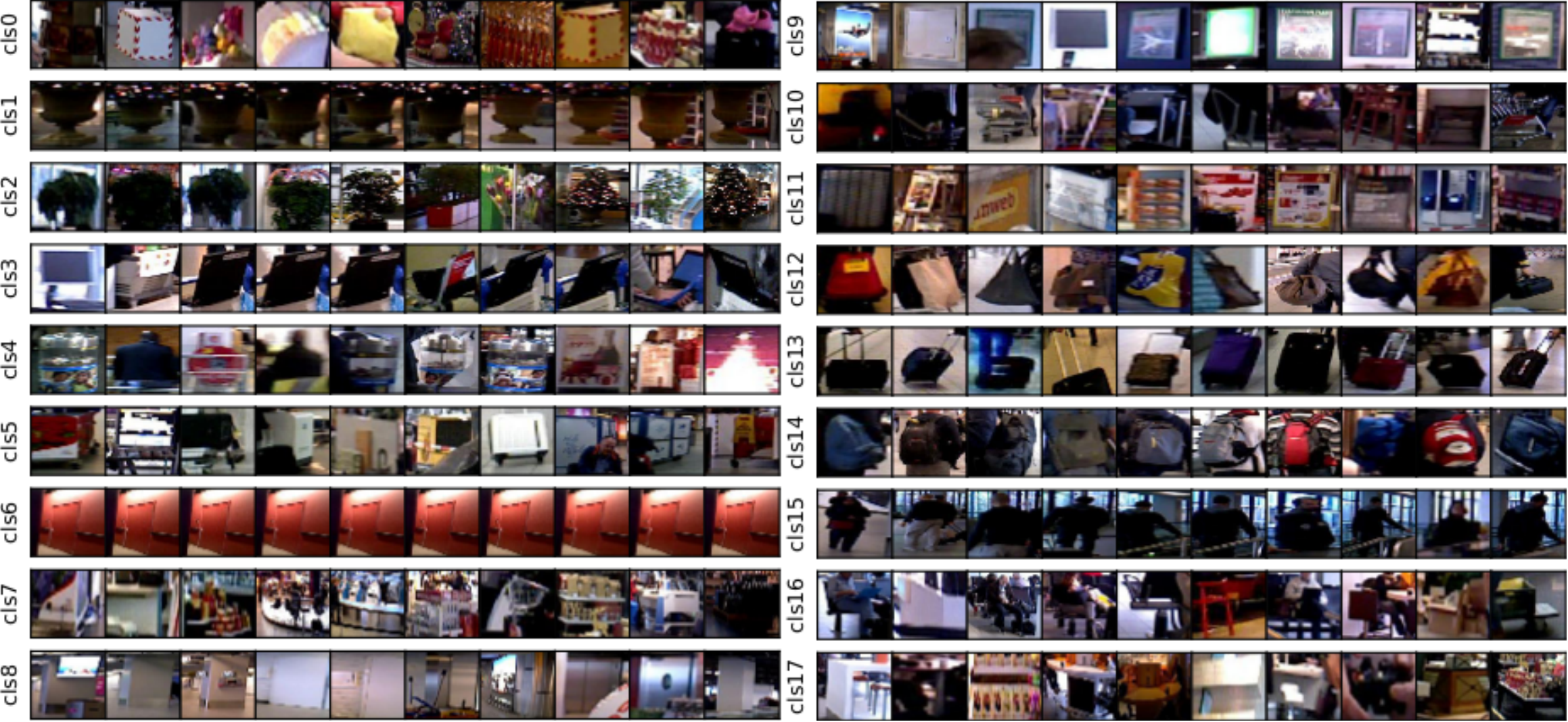}
  \caption{Visualization of the discovered categories on Schiphol. Each row shows example crops for one cluster.}
  \label{fig:clustering-schiphol}
  \end{center}
\end{figure*}

\begin{figure*}
\begin{center}
  \includegraphics[width=1.0\linewidth]{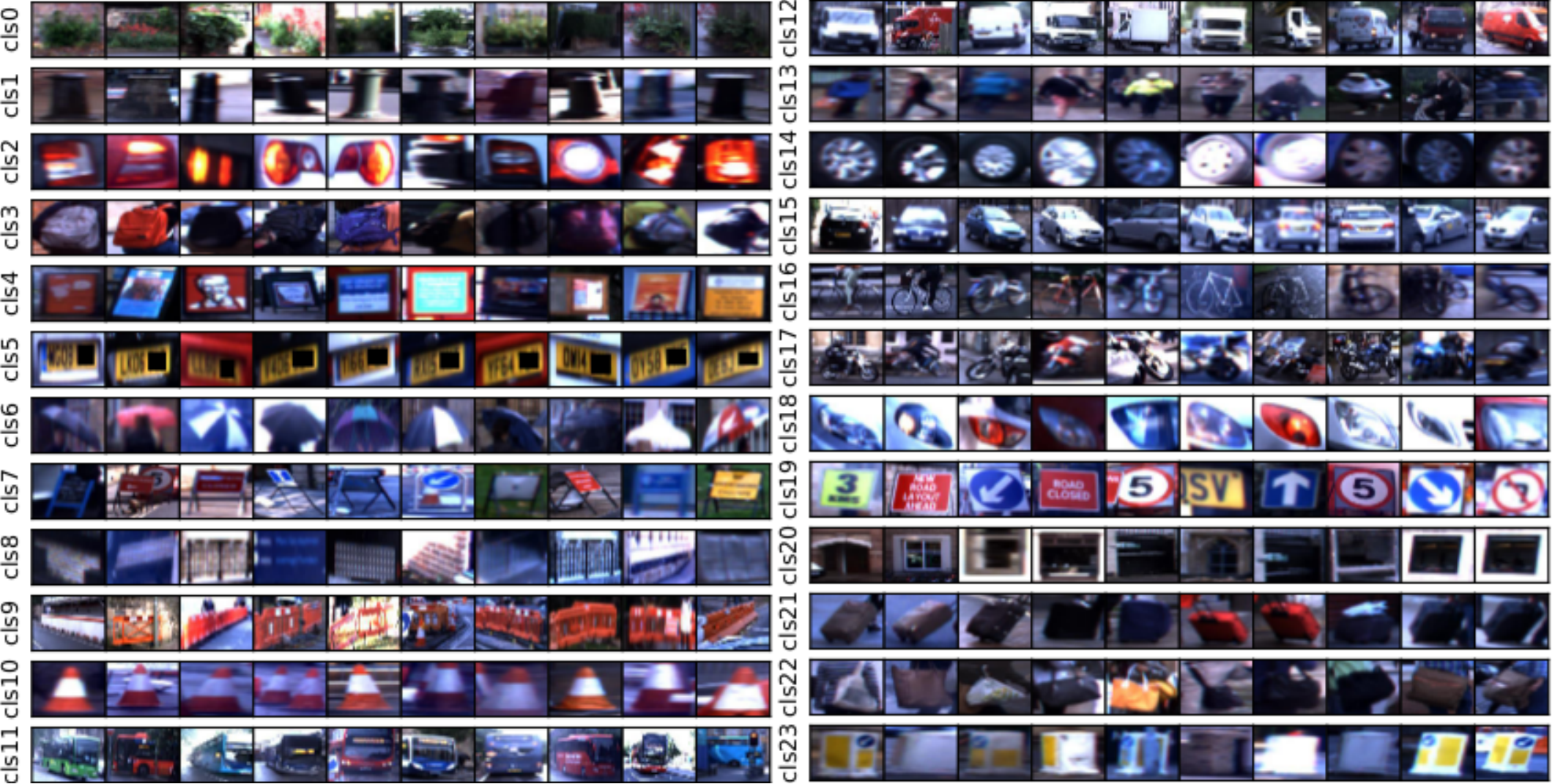}
  \caption{Visualization of the discovered categories on Oxford. Each row shows example crops for one cluster.}
  \label{fig:clustering-oxford}
  \end{center}
\end{figure*}

\PAR{Category Discovery.}
The discovered categories on the three datasets are shown in Figures \ref{fig:clustering-KITTI}, \ref{fig:clustering-schiphol}, and \ref{fig:clustering-oxford}. The categories discovered in the KITTI Raw and Oxford Robotcar datasets are quite similar as both are video datasets recorded in similar driving scenarios. For both datasets our pipeline correctly groups the most common known categories of people, cars, trucks, bicycles, motorbikes, and traffic lights. We also discover a number of unknown categories in both the KITTI Raw and Oxford Robotcar datasets, these are traffic signs, license plates, greenery, windows and fences. Since Oxford is a much larger dataset many more categories of objects were able to be discovered. This includes the already known object categories of buses, suitcases, handbags, umbrellas and backpacks; as well as the unknown categories of indicators, sidewalk posts, advertisement signs, traffic cones and wheels. In the KITTI Raw dataset we also discovered a trash can cluster that was not discovered in the Oxford dataset. 

The Schiphol dataset is very different to the other two datasets as it was captured on the inside of an airport. Within this dataset we manage to discover a number of already known categories such as people, backpacks, suitcases, handbags, vases and laptops; as well as a number of unknown categories such as Christmas trees and screens.

\begin{figure*}[ht]
  \includegraphics[width=0.51\linewidth]{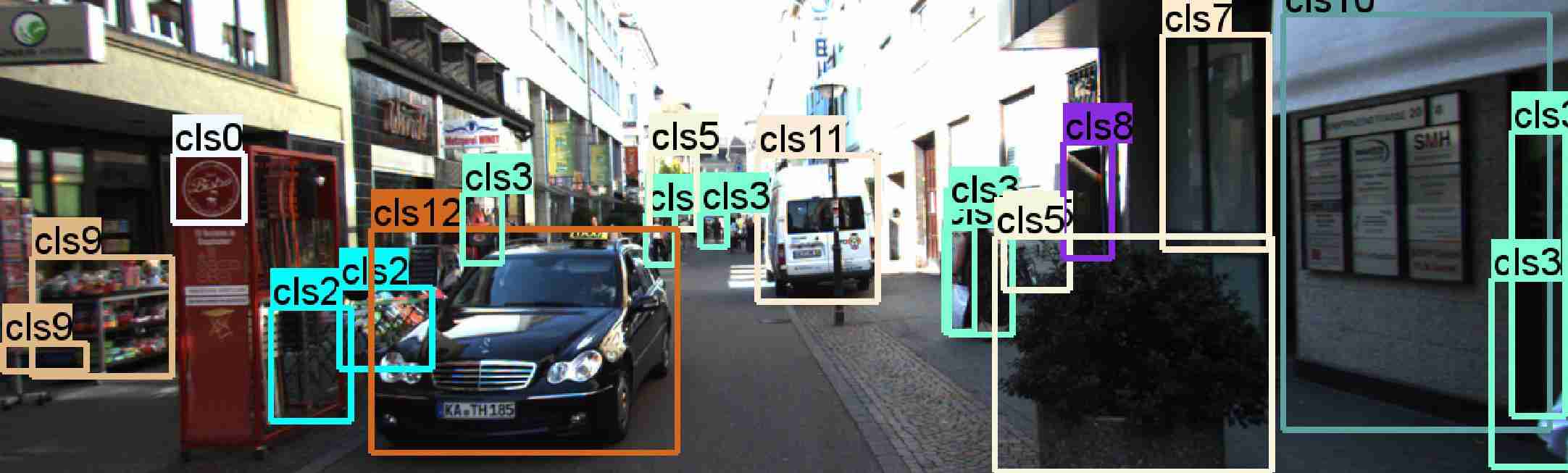}
  \includegraphics[width=0.51\linewidth]{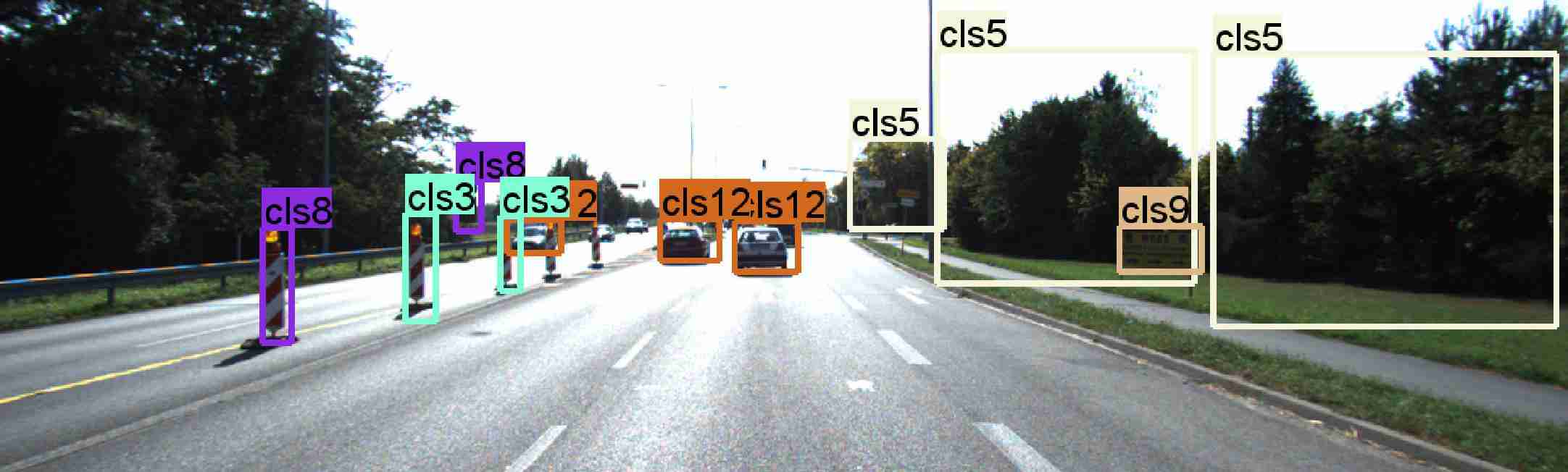}
  \includegraphics[width=0.51\linewidth]{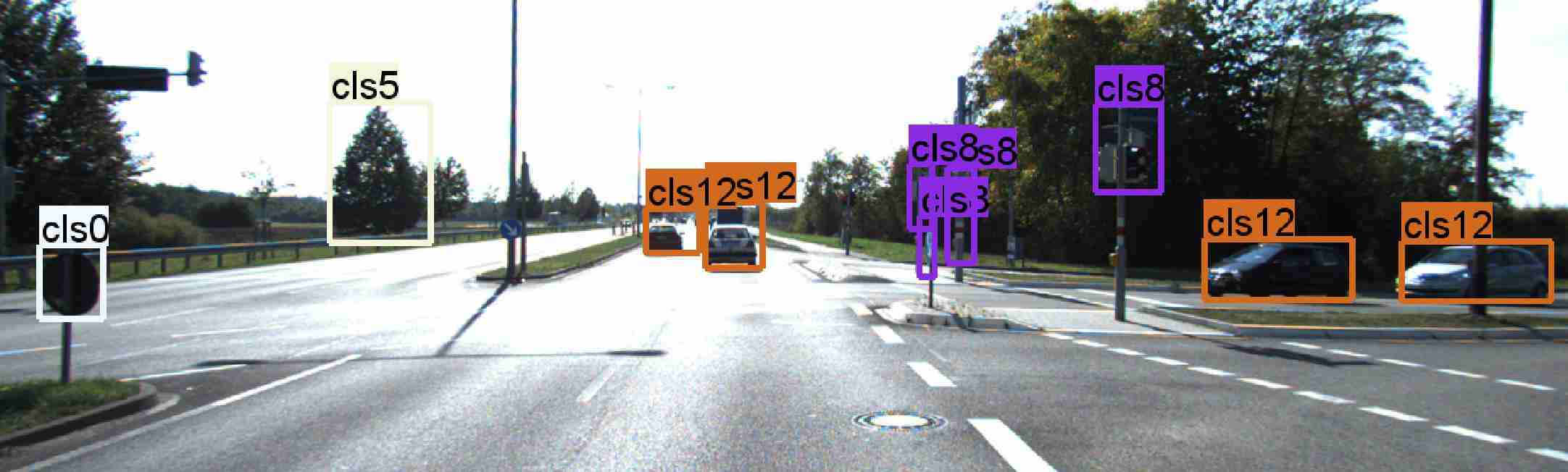}
  \includegraphics[width=0.51\linewidth]{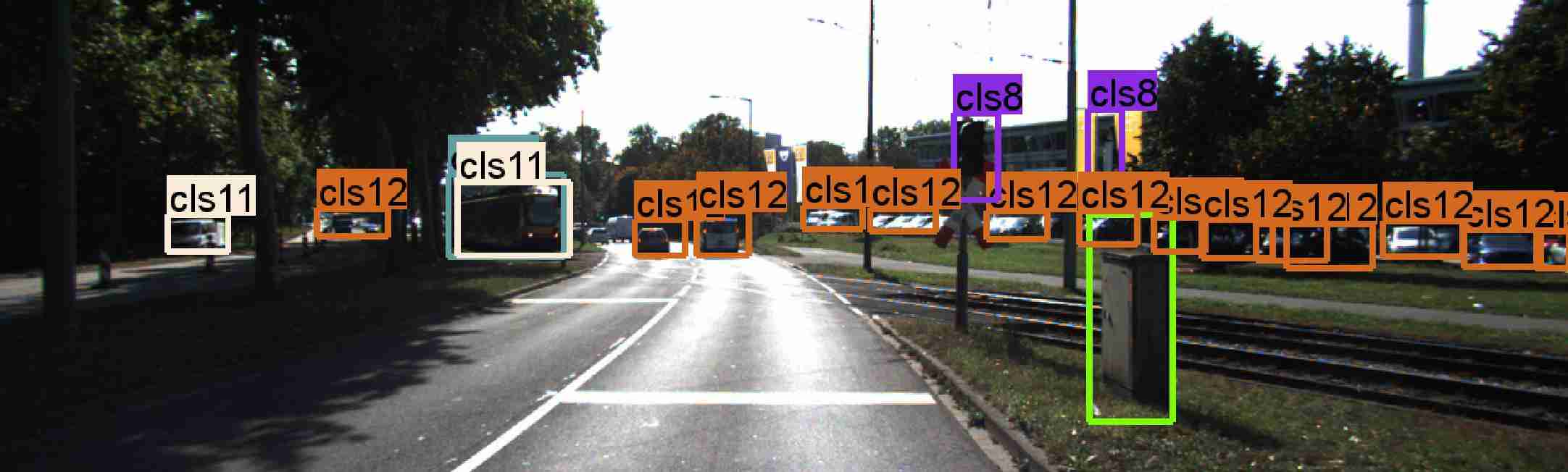}
  \includegraphics[width=0.51\linewidth]{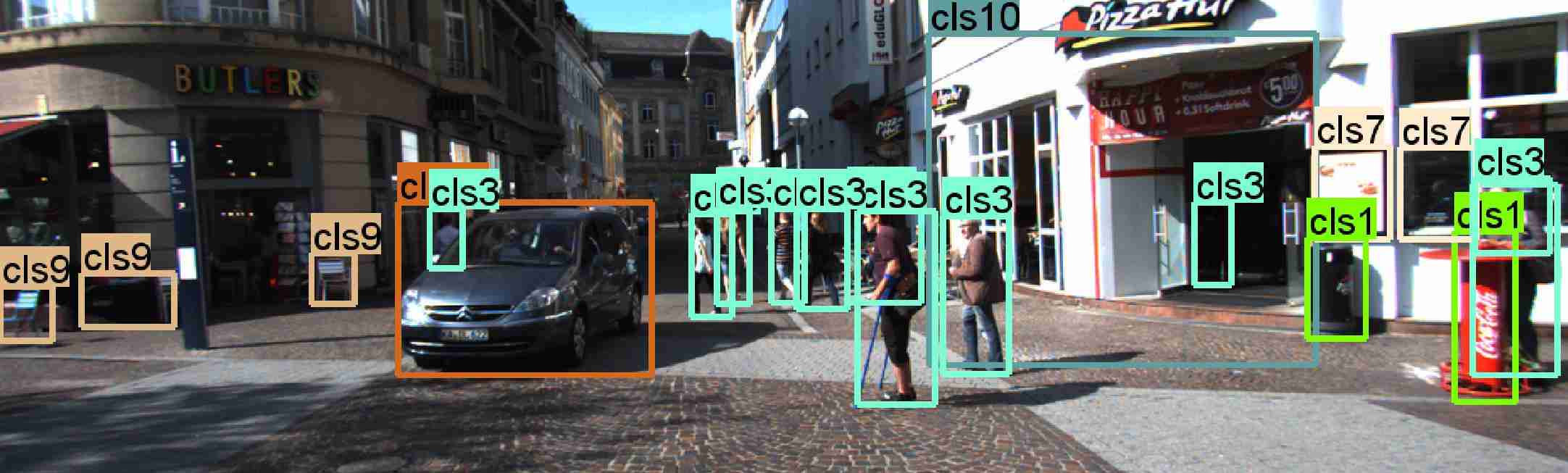}
  \includegraphics[width=0.51\linewidth]{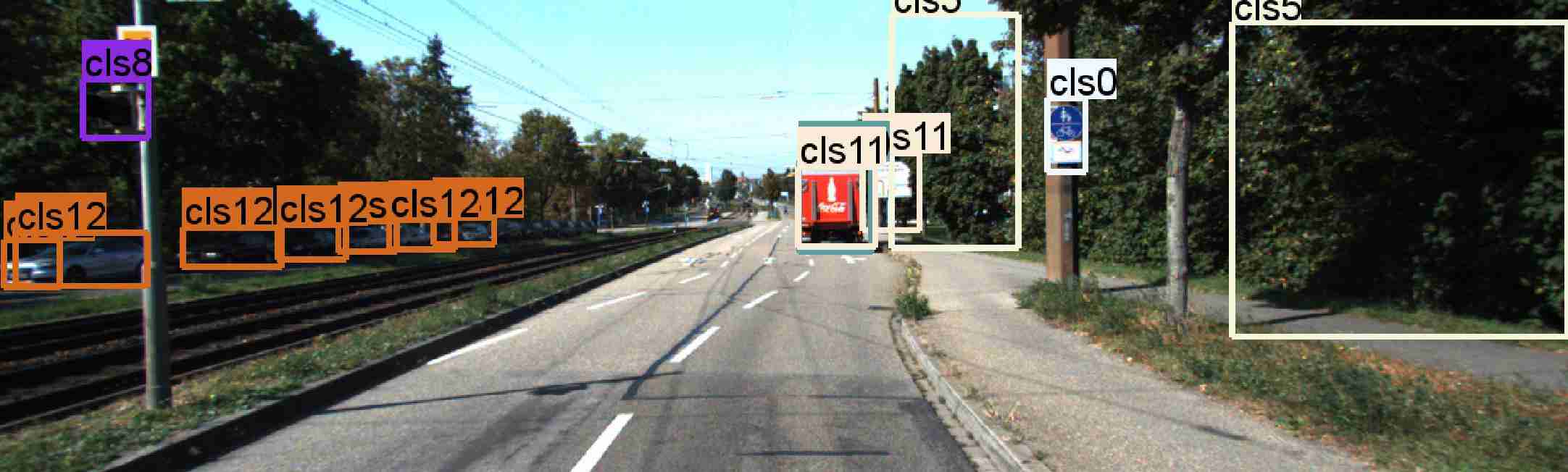}
  \includegraphics[width=0.51\linewidth]{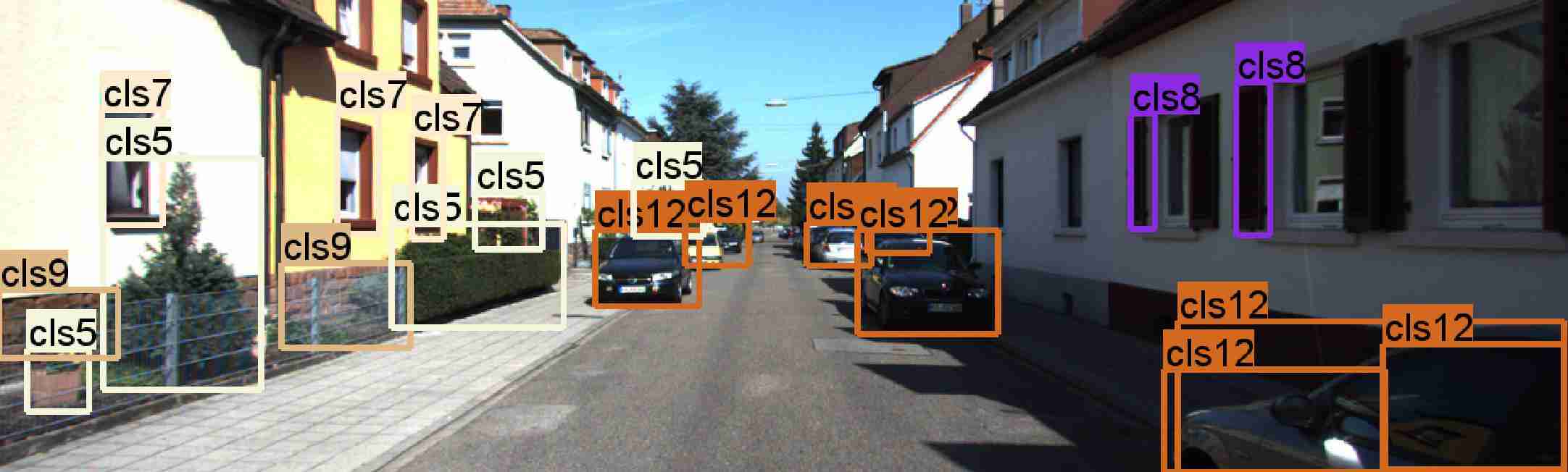}
  \includegraphics[width=0.51\linewidth]{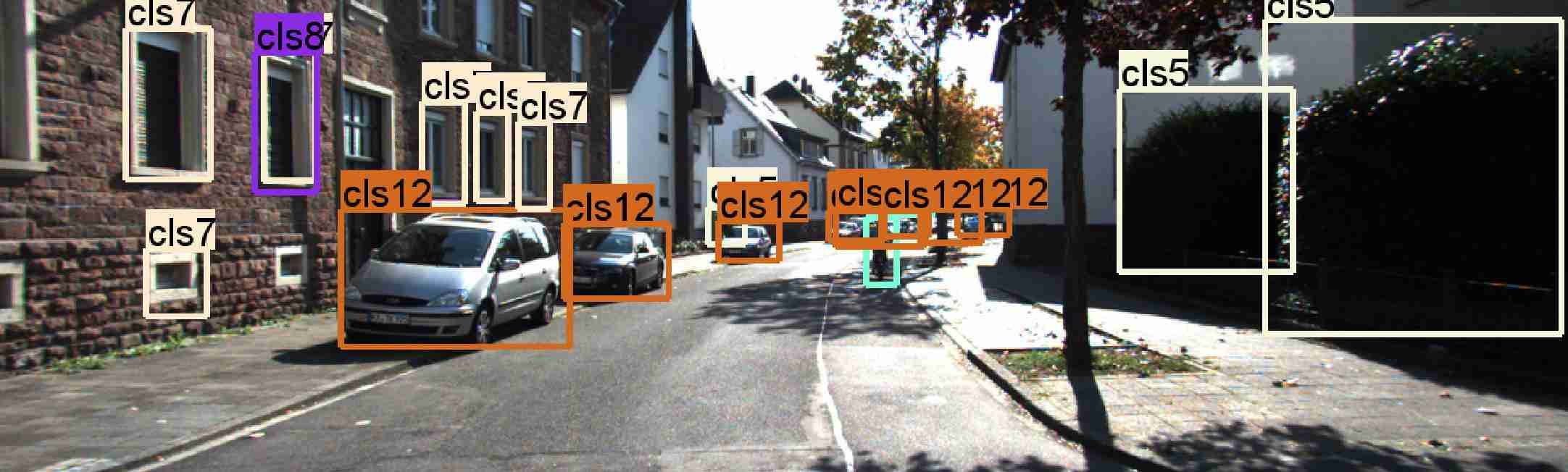}
  \caption{Qualitative results for learning new detectors on automatically discovered categories on KITTI Raw. The clusters are labeled by cls0 to cls12. Here, just the detector is evaluated and no tracking is performed. The detections are noisy since they were only trained on automatically generated and clustered tracks.}
  \label{fig:det-KITTI}
\end{figure*}

\begin{figure*}[ht]
  \includegraphics[width=0.33\linewidth]{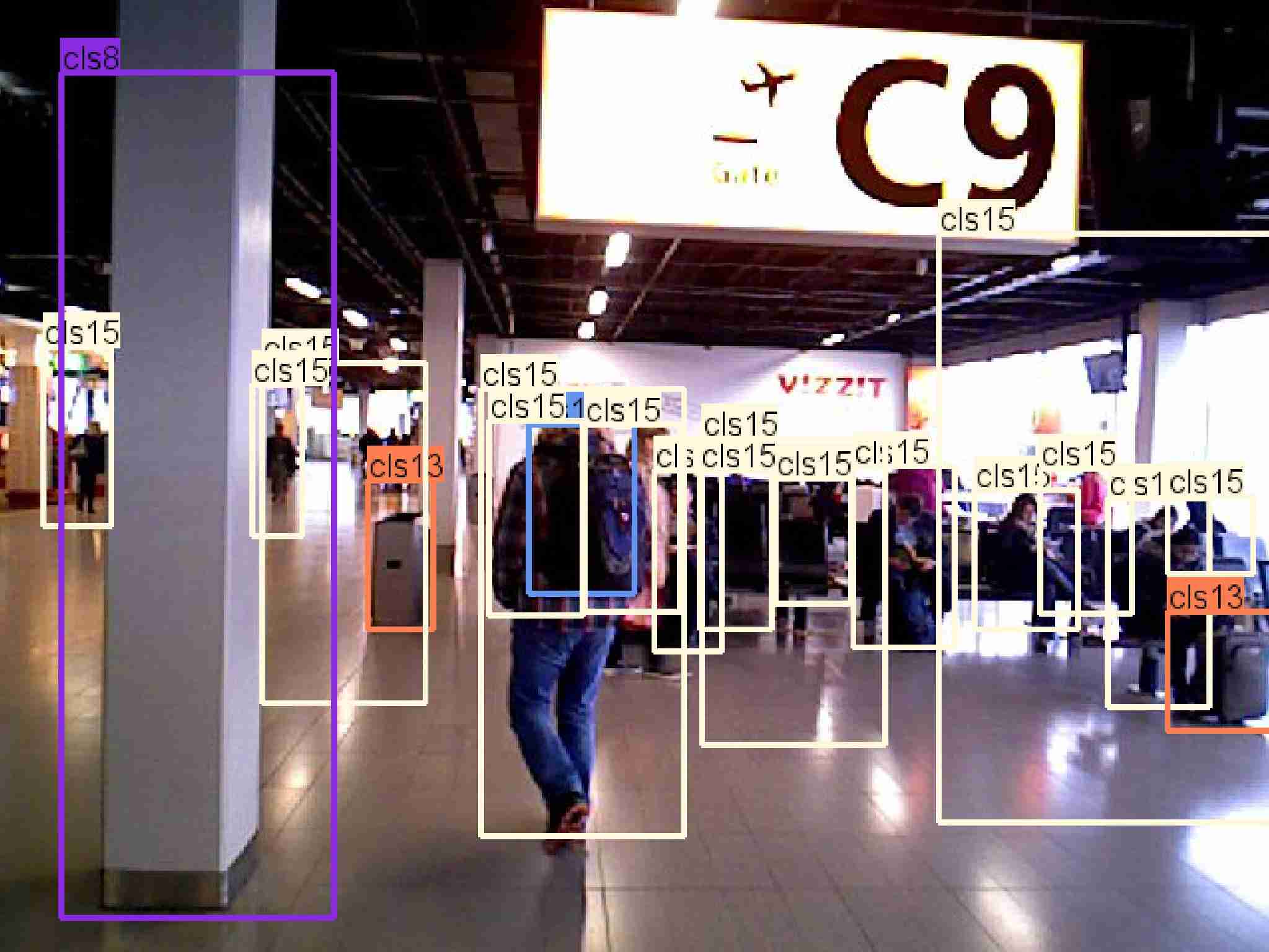}
  \includegraphics[width=0.33\linewidth]{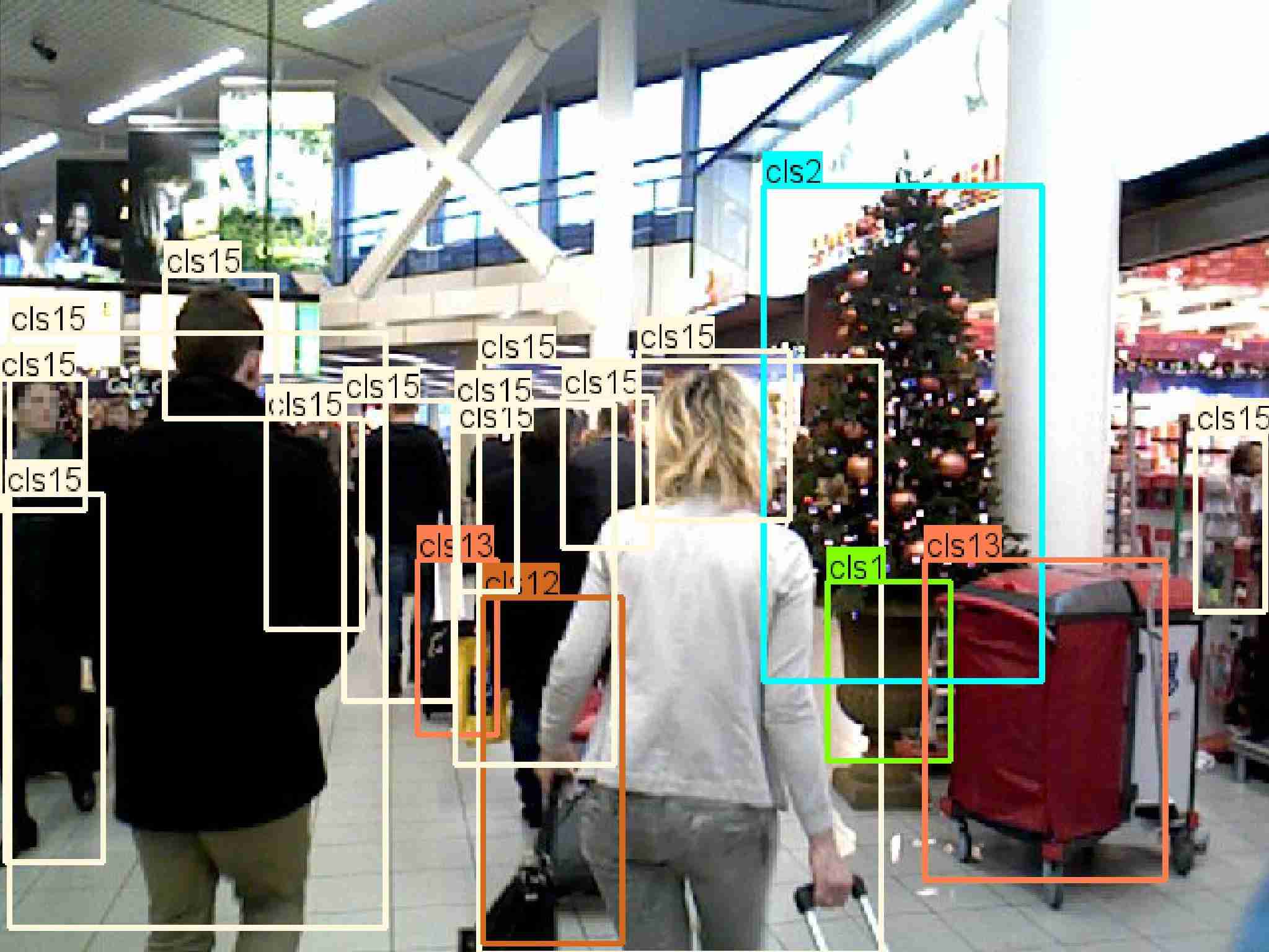}
  \includegraphics[width=0.33\linewidth]{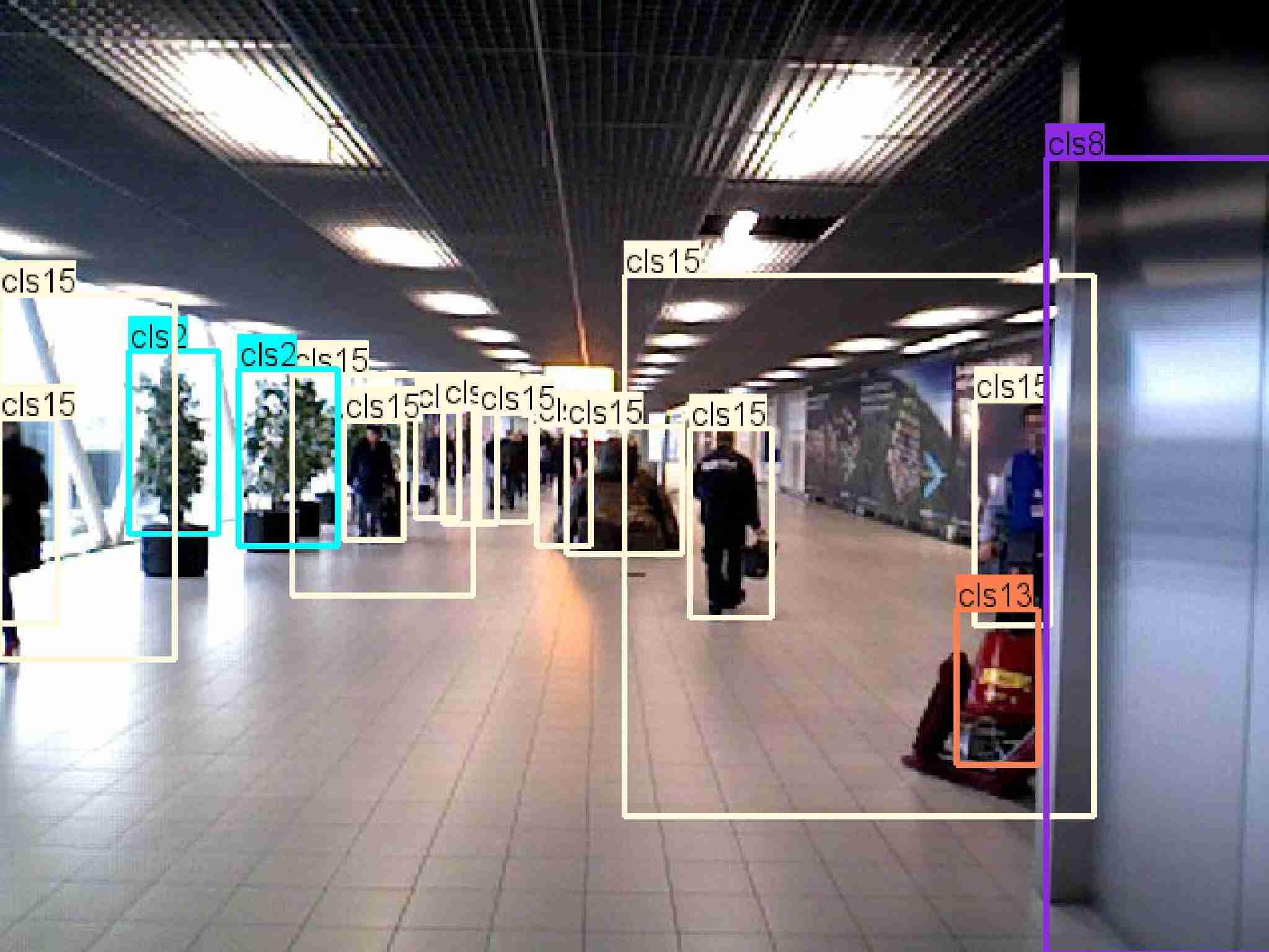}
  \caption{Qualitative results for learning new detectors on automatically discovered categories on Schiphol. The clusters are labeled by cls0 to cls17. Here, just the detector is evaluated and no tracking is performed. The detections are noisy since they were only trained on automatically generated and clustered tracks. Some faces have been pixelized to preserve privacy.}
  \label{fig:det-schiphol}
\end{figure*}

\begin{figure*}[ht]
  \begin{center}
  \newcommand{\mysize}{0.33}
  \newcommand{\myspace}{\hspace{0.3cm}}
  \includegraphics[width=\mysize\linewidth]{figures/detector/oxford/postproc_0000000020_cropped.jpg}
  \includegraphics[width=\mysize\linewidth]{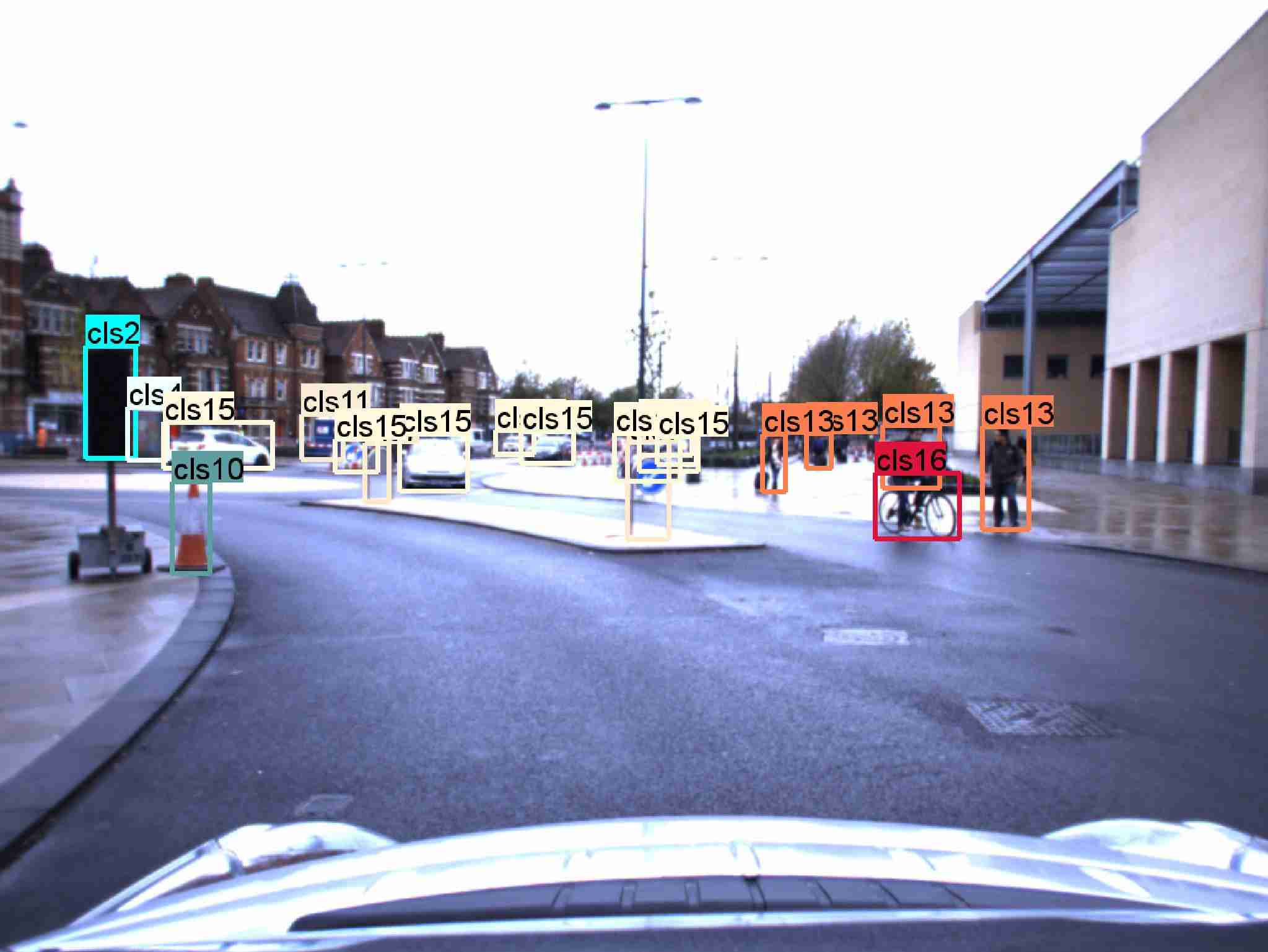}
  \includegraphics[width=\mysize\linewidth]{figures/detector/oxford/postproc_0000000155_cropped.jpg}
  \includegraphics[width=\mysize\linewidth]{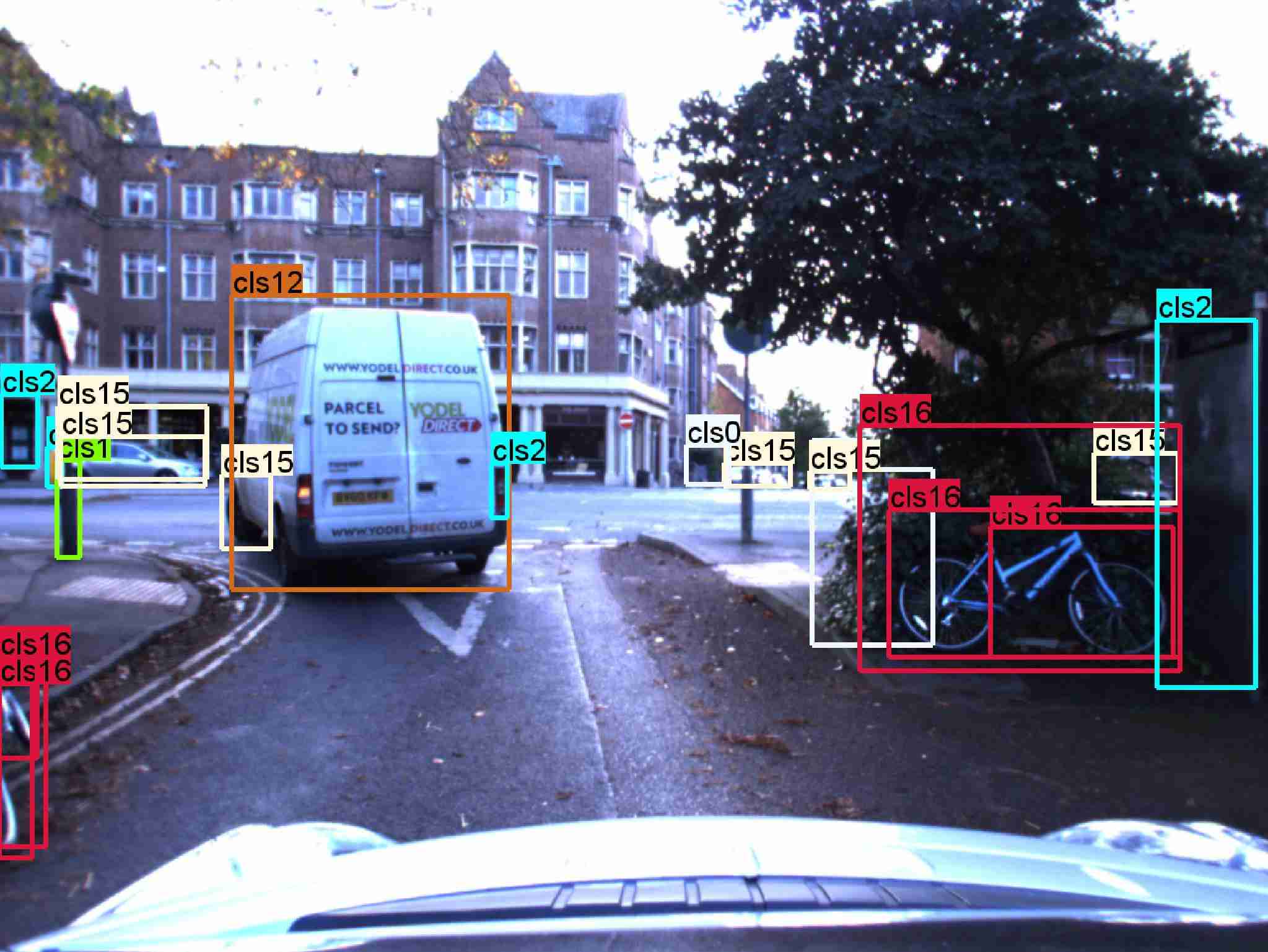}
  \includegraphics[width=\mysize\linewidth]{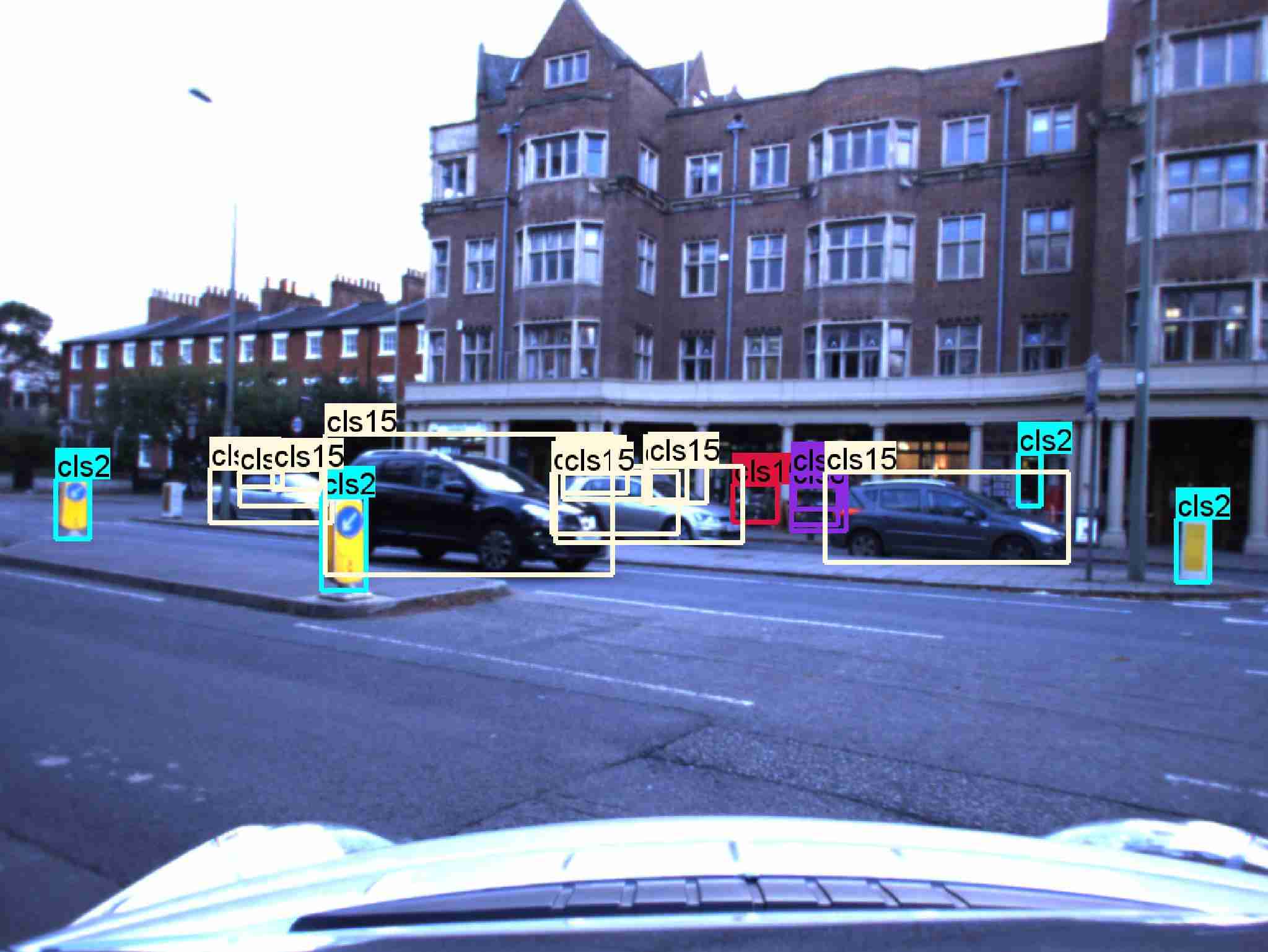}
  \includegraphics[width=\mysize\linewidth]{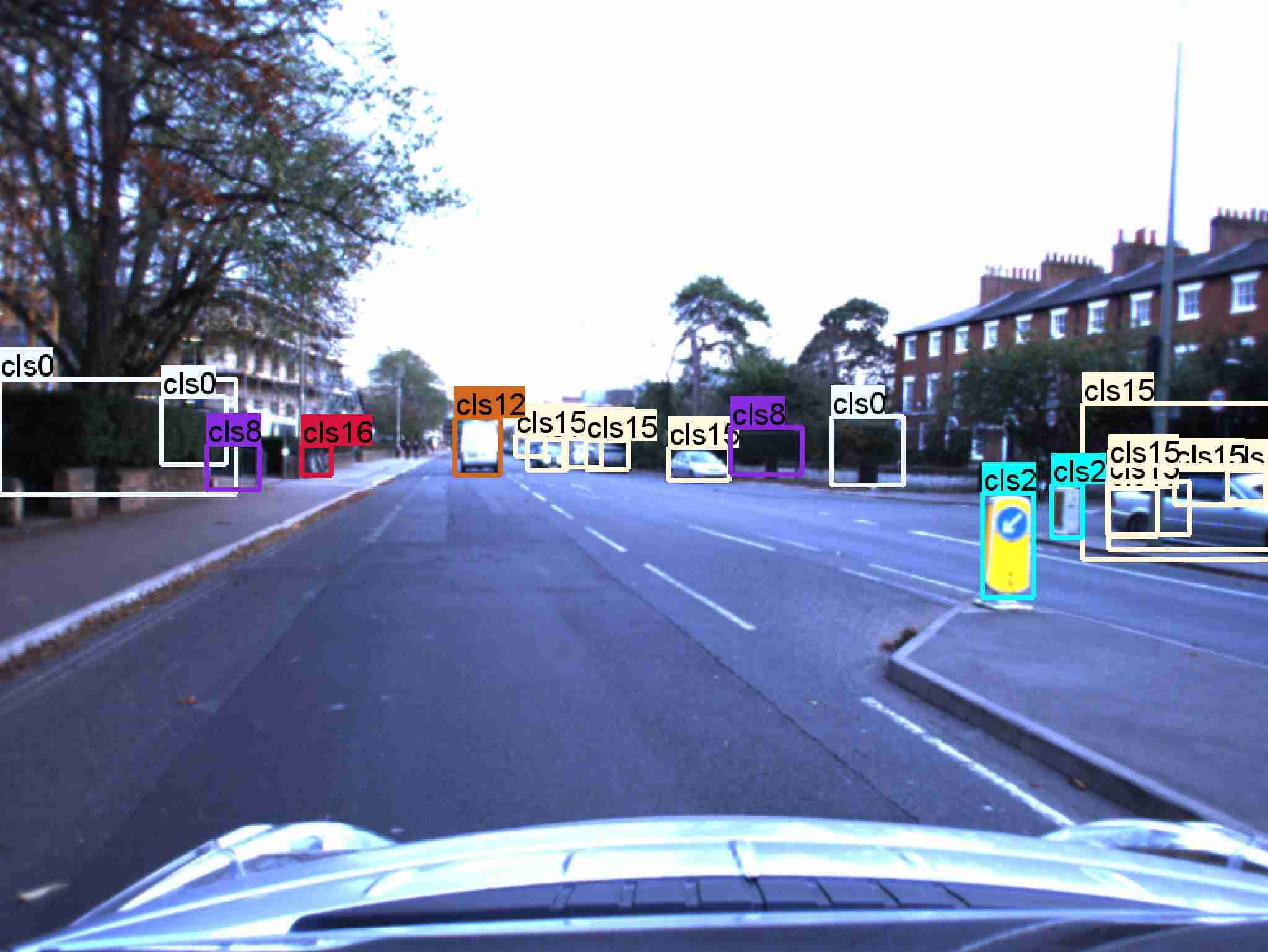}
  \includegraphics[width=\mysize\linewidth]{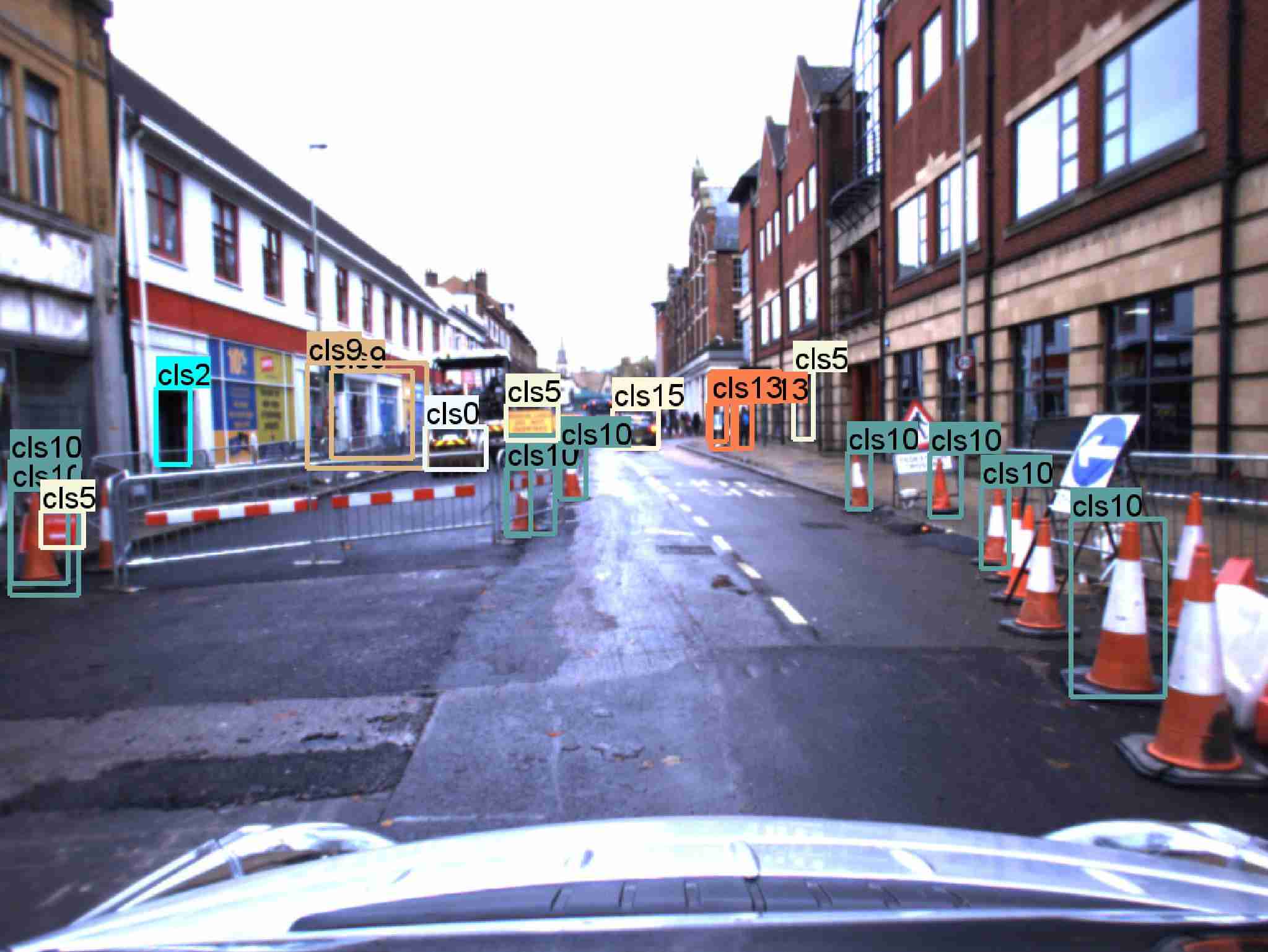}
  \includegraphics[width=\mysize\linewidth]{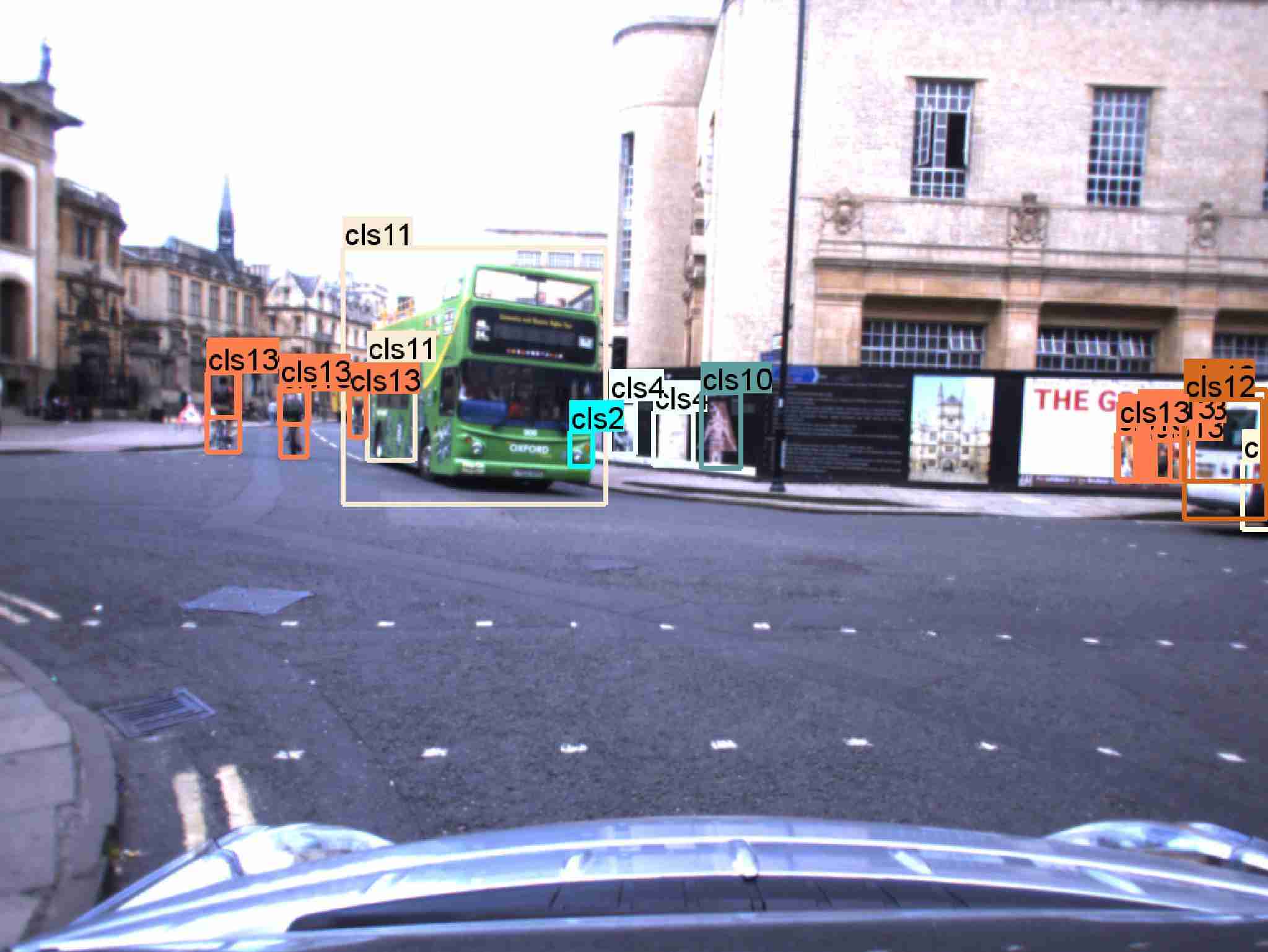}
  \includegraphics[width=\mysize\linewidth]{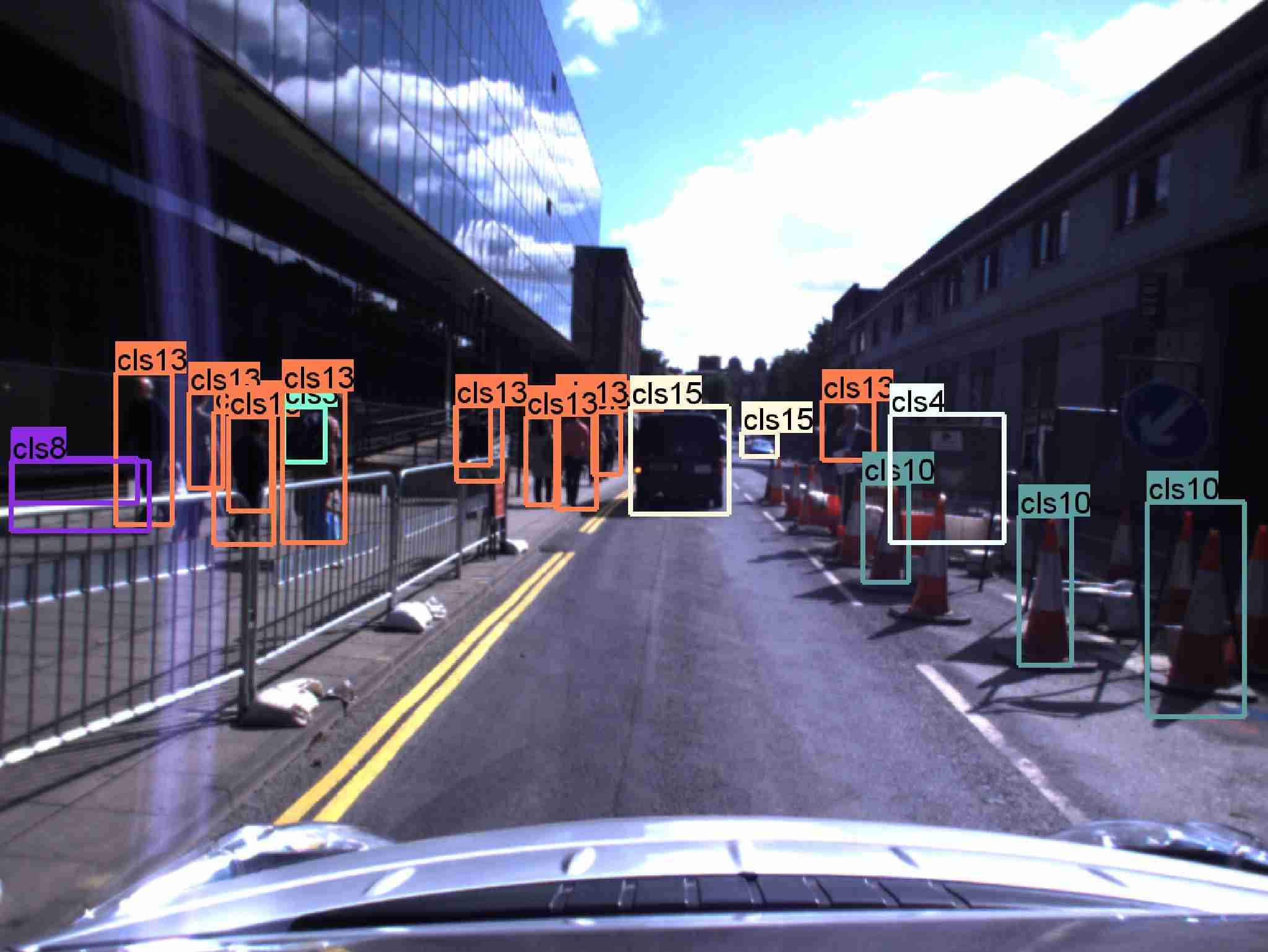}
  \end{center}
  \caption{Qualitative results for learning new detectors on automatically discovered categories on Oxford. We were able to learn detectors for new categories like traffic cones and poles. The clusters are labeled by cls0 to cls24. Here, just the detector is evaluated and no tracking is performed. The detections are noisy since they were only trained on automatically generated and clustered tracks.}
  \label{fig:det-oxford}
\end{figure*}

\PAR{Learning new Detectors.}
Figures \ref{fig:det-KITTI}, \ref{fig:det-schiphol}, and \ref{fig:det-oxford} show qualitative results for new detectors which were learned on the automatically discovered category clusters. The figures show the results of evaluating the learned detectors on the shown frames without using temporal context information. These preliminary results demonstrate that it is indeed possible to learn a detector from automatically obtained clusters. For example, we can detect traffic cones, license plates, car lights, poles, and Christmas trees, which were not annotated in COCO. However, the detectors are not yet as robust as detectors trained on hand-labeled data.


%

%

%% file: KTC_stats.tex
\begin{center}
  \begin{table*}
  \small
    \begin{tabular}{l|c|r|Sc||l|c|r|Sc} 
      \textbf{Name} & \textbf{In COCO?} & \textbf{Frequency} & \textbf{Example} & \textbf{Name} & \textbf{In COCO?} & \textbf{Frequency} & \textbf{Example} \\\hline\hline
      car & yes & \makecell{2,405\\30.04\%} & \cincludegraphics[height=1.0cm]{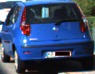} & building & no & \makecell{67\\0.84\%} & \cincludegraphics[height=1.0cm]{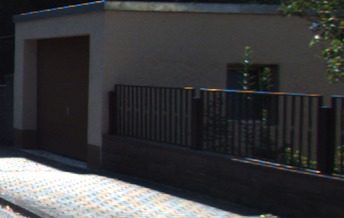} \\\hline
      greenery & (potted plant) & \makecell{1,124\\14.04\%} & \cincludegraphics[height=1.0cm]{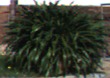} & motorcycle & yes & \makecell{56\\0.70\%} & \cincludegraphics[height=1.0cm]{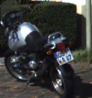} \\\hline
      window & no & \makecell{370\\4.62\%} & \cincludegraphics[height=1.0cm]{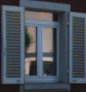} & pole & no & \makecell{47\\0.59\%} & \cincludegraphics[height=1.0cm]{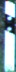} \\\hline
      person & yes & \makecell{272\\3.40\%} & \cincludegraphics[height=1.0cm]{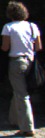} & trailer & no & \makecell{45\\0.56\%} & \cincludegraphics[height=1cm]{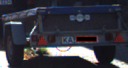} \\\hline
      road sign & (stop sign) & \makecell{192\\2.40\%} & \cincludegraphics[height=1.0cm]{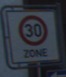} & wheel & no & \makecell{28\\0.35\%} & \cincludegraphics[height=1.0cm]{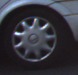} \\\hline
      fence & no & \makecell{185\\2.31\%} & \cincludegraphics[height=1.0cm]{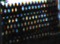} & stair & no & \makecell{20\\0.25\%} & \cincludegraphics[height=0.8cm]{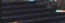} \\\hline
      van & no & \makecell{142\\1.77\%} & \cincludegraphics[height=1.0cm]{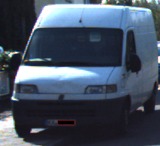} & license plate & no & \makecell{13\\0.16\%} & \cincludegraphics[height=0.6cm]{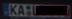} \\\hline
      bicycle & yes & \makecell{134\\1.67\%} & \cincludegraphics[height=1.0cm]{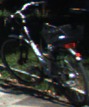} & umbrella & yes & \makecell{13\\0.16\%} & \cincludegraphics[height=1.0cm]{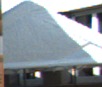} \\\hline
      wall & no & \makecell{128\\1.60\%} & \cincludegraphics[height=1.0cm]{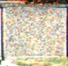} & bench & yes & \makecell{11\\0.14\%} & \cincludegraphics[height=1.0cm]{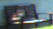} \\\hline
      advertising & no & \makecell{119\\1.49\%} & \cincludegraphics[height=1.0cm]{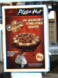} & bus & yes & \makecell{10\\0.12\%} & \cincludegraphics[height=1.0cm]{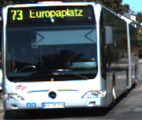} \\\hline
      rubbish bin & no & \makecell{100\\1.25\%} & \cincludegraphics[height=1.0cm]{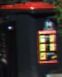} & post box & no & \makecell{6\\0.08\%} & \cincludegraphics[height=1.0cm]{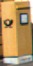} \\\hline
      gate & no & \makecell{97\\1.21\%} & \cincludegraphics[height=1.0cm]{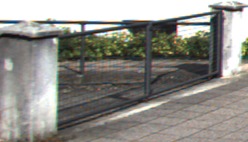} & cone & no & \makecell{3\\0.04\%} & \cincludegraphics[height=1.0cm]{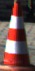} \\\hline
      traffic light & yes & \makecell{87\\1.09\%} & \cincludegraphics[height=1.0cm]{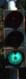} & train & yes & \makecell{3\\0.04\%} & \cincludegraphics[height=1.0cm]{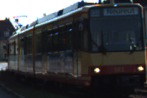} \\\hline
      truck & yes & \makecell{86\\1.07\%} & \cincludegraphics[height=1.0cm]{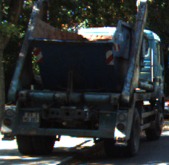} & chair & yes & \makecell{2\\0.02\%} & \cincludegraphics[height=1.0cm]{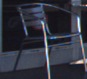} \\\hline
      transformer & no & \makecell{84\\1.05\%} & \cincludegraphics[height=1.0cm]{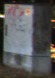} & animal & (dog, horse, ...) & \makecell{1\\0.01\%} & \cincludegraphics[height=1.0cm]{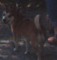} \\\hline
      stone pillar & no & \makecell{76\\0.95\%} & \cincludegraphics[height=1.0cm]{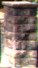} &  &  &  &  \\\hline
      cyclist & yes & \makecell{72\\0.90\%} & \cincludegraphics[height=1.0cm]{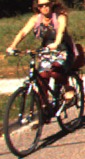} & unknown & - & \makecell{1,190\\14.87\%} & \cincludegraphics[height=1.0cm]{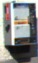} \\\hline
      door & no & \makecell{72\\0.90\%} & \cincludegraphics[height=1.0cm]{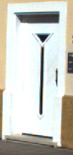} & tracking error & - & \makecell{745\\9.31\%} & \cincludegraphics[height=1.0cm]{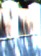}
    \end{tabular}
    \label{Tab:allClasses}
    \vspace{5pt}
    \caption{Overview of all the object classes, extracted from KITTI Raw. The frequency gives the relative amount of tracks of that class inside the dataset. License plates have been made unreadable. Note that 18 (+3 super classes) are not part of the COCO dataset.}
\end{table*}
\end{center}